%% file: main.tex
\documentclass[a4paper,12pt,oneside]{book}

\usepackage[headings]{fullpage}
\usepackage{setspace}
\usepackage{enumitem}

  \usepackage[top=4.0cm,left=4.0cm,bottom=2.5cm,right=2.5cm]{geometry}
\usepackage{graphicx}
\usepackage{float}
\usepackage{array}
\usepackage{subfig}
\usepackage{booktabs}
\usepackage{multirow}
\usepackage{appendix}

\usepackage{hyperref}
\usepackage[printonlyused]{acronym}
\usepackage{algorithmic}
\usepackage{algorithm}
\usepackage{ifpdf}
\usepackage{arabtex}
\usepackage{utf8}

 \usepackage{comment}
 \usepackage[sort,compress]{cite}

\usepackage[utf8]{inputenc}
\usepackage{array}
 
 \usepackage{times}
 \usepackage{amsmath}
\usepackage{lmodern}

\newlength\longest
\newcolumntype{M}[1]{>{\centering\arraybackslash}m{#1}}

\usepackage{gb4e}
\noautomath

\newcommand{\submissionDay}{}
\newcommand{\submissionMonth}{May}
\newcommand{\submissionYear}{2021}
\newcommand{\submissionDate}{\submissionDay~\submissionMonth,~\submissionYear}
\newcommand{\typeOfThesis}{ Ph.D. Thesis}

\newcommand{\titleOfThesisOne}{ Thesis Title}

\newcommand{\authorOfThesis}{Caroline Nabil Sabty}

\ifpdf
\fi

\begin{document}
\pagestyle{plain}
\pagenumbering{Roman}

\input{GUC_TitlePage}
\input{dedication}

\input{acknowledgments}

\input{abstract}
\tableofcontents
\clearpage 

\pagestyle{headings}
\pagenumbering{arabic}

\setlength\parskip{.5\baselineskip plus .2\baselineskip
	minus .4\baselineskip}


\input{IntroFinal}

\input{Background}
 \input{RelatedWorkNew}

\input{NERMSA}

 \input{NERCS}

\input{Pre-trained_Embeddings}

 \input{Data-Augmentation}
\input{LID}

\input{conclusion}

\input{appendix}
 
\bibliographystyle{acm}
\bibliography{main}
\addcontentsline{toc}{chapter}{References}

\end{document}

%% file: GUC_TitlePage.tex
\newcommand{\titlePage}{

\thispagestyle{empty}
\begin{center}
	\textbf{Media Engineering and Technology Faculty}\\[1mm]
	\textbf{German University in Cairo}\\[1mm]
	\includegraphics[width=2.5cm]{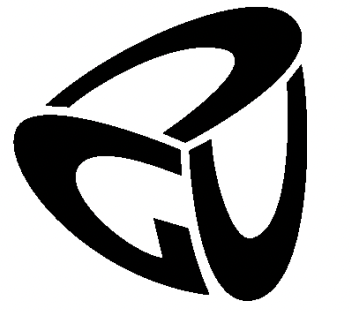}
	
	\vspace{2cm}
	\doublespacing
	{\Huge \textbf{Computational Approaches to Arabic-English Code-Switching
}}\\
	\singlespacing
	\vspace{2cm}
	{\large \textbf{\typeOfThesis}}\\
	
	\vfill
	\parbox{1cm}{
  		\begin{large}
    			\begin{tabbing}
       			Author: \hspace{2cm}  
        			\=Caroline Nabil Samy Sabty\\[2mm]
      			Supervisors: 
        			\>Prof. Dr. 
Slim Abdennadher\\[2mm]
		
      			Submission Date: 
        			\>\submissionDate\\
    			\end{tabbing}
  		\end{large}
	}\\
\end{center}
\clearpage
}
\titlePage
\thispagestyle{empty}
This is to certify that:
\begin{itemize}
\item[(i)] the thesis comprises only my original work toward ...
\item[(ii)] due acknowlegement has been made in the text to all other material used
\end{itemize}

\vspace{2cm}
\begin{flushright}
\rule[0mm]{6cm}{0.2mm}\\
\authorOfThesis\\
\submissionDay~\submissionMonth,~\submissionYear\\
\end{flushright}

%% file: dedication.tex
 \label{chap:Dedication}
\clearpage

\thispagestyle{empty}
\null\vfill
{
\settowidth\longest{\huge\itshape  my angel and beloved sister Nathalie;}
\centering
\parbox{\longest}{%
  \centering{\huge\itshape%
 I would like to dedicate my thesis to my angel and beloved sister Nathalie\par\bigskip
  }   
  \Large\MakeUppercase{}\par%
}}

\vfill\vfill

\clearpage


%% file: acknowledgments.tex
\chapter*{Acknowledgments}
\addcontentsline{toc}{chapter}{Acknowledgments}
\label{chap:ack}
First and foremost, praise and thanks to God for His continuous showers of blessings, unconditional love and guidance throughout my journey. Looking back at every step along the way I can see clearly His perfect plan for me and it was all done because of His help and blessing.

I would like to thank my supervisor and mentor, Prof.
Dr. Slim Abdennadher, my role mode and great supporter throughout my journey. He was the source of inspiration and encouragement to pursue post-graduate studies. He was always there for me with his constant guidance and supervision throughout my years of studies and career. He believed in me and motivated me during hard times. I was fortunate I got to learn so much from him. I learned how to be very passionate about teaching no matter how many times the information is being repeated. I learned the excitement of doing research and how to keep trying till reaching the goal. Words will never be enough to express my gratitude to him for being where I am today. 

I am eternally grateful to my sister Nathalie. She was the perfect example for persistence and having the right mindset through out the most challenging situations when getting something done. She has never given up no matter how burdensome something might get. Everyday during my work, I was channeling her mindset to encourage myself, to advance more and to eventually reach that finish line. Even though I wanted her to be here to witness this moment more than anything else, I can safely say that I felt her presence with me in every page, guiding me, looking over my shoulder and never leaving my side. I am eternally grateful to you, Nathalie.

I want to express my utter gratitude to my parents (Nabil and Mervat) who have been nothing but supportive and devoted to me and to my work. I am indebted for their endless love and prayers. Despite all the hardships we faced, they never failed to have my back, constantly pushed me forward and encouraged me to put my work above anything. So thank you wholeheartedly for always raising me up the way you do.

I am genuinely grateful for my loving, caring husband and backbone (Nader) for always believing in me, for all the sacrifices he did and for all his support and assistance to make sure my work flew undisturbed. I am a lucky and blessed wife to have such a man by my side during this journey. I also want to thank my kids (Gamal and Nathalie) for showering me with their love through it all and also for their acceptance of the times where i was mostly focused on my work.

I want to thank my mother-in-law (Nadia) for her continuous support and help. I learned a lot from her, she is a true meaning of a self-giving person. A special thank you for my father-in-law (Gamal) who was a true legend. Marie, Rami, and Marguerite, thank you for your continuous encouragement.

I cannot thank enough my friends and sisters Nada and Marlein, who were always there for me. Things would have been much harder without them, I am grateful to have them in my life. One of the things I really cherish is sharing the journey of my PhD. with my very special and loving friends, Injy and Alia. Injy has helped me find my way in the field and the topic and Alia used to listen to me and help me overcome a lot of challenges. 

Nermine, Mirna, Carole and Steve my oldest lifetime friends, regardless
of where you live and how often we meet, I never feel that you are away, thank you for always being there for me. Carine, Amir, Caroline and Mina thank you for your continuous encouragement.

A special thank you to Dr. Mohamed Elmahdy for his guidance at the beginning of my research journey and for helping me to find my way and passion to my topic. 
To all my instructors, friends and colleagues especially Dr. Mervat, thank you for the motivation and inspiration.


%% file: abstract.tex
\chapter*{Abstract}
 \addcontentsline{toc}{chapter}{Abstract}
\label{chap:abstract}
Natural Language Processing (NLP) is a vital computational method for addressing language processing, analysis, and generation. NLP tasks form the core of many daily applications, from automatic text correction to speech recognition. While significant research has focused on NLP tasks for the English language, less attention has been given to Modern Standard Arabic and Dialectal Arabic. Globalization has also contributed to the rise of Code-Switching (CS), where speakers mix languages within conversations and even within individual words (intra-word CS). This is especially common in Arab countries, where people often switch between dialects or between dialects and a foreign language they master. CS between Arabic and English is frequent in Egypt, especially on social media. Consequently, a significant amount of code-switched content can be found online. Such code-switched data needs to be investigated and analyzed for several NLP tasks to tackle the challenges of this multilingual phenomenon and Arabic language challenges. No work has been done before for several integral NLP tasks on Arabic-English CS data. In this work, we focus on the Named Entity Recognition (NER) task and other tasks that help propose a solution for the NER task on CS data, e.g., Language Identification. This work addresses this gap by proposing and applying state-of-the-art techniques for Modern Standard Arabic and Arabic-English NER. We have created the first annotated CS Arabic-English corpus for the NER task. Also, we apply two enhancement techniques to improve the NER tagger on CS data using CS contextual embeddings and data augmentation techniques. All methods showed improvements in the performance of the NER taggers on CS data. Finally, we propose several intra-word language identification approaches to determine the language type of a mixed text and identify whether it is a named entity or not.

%% file: IntroFinal.tex
\chapter{Introduction}
\label{chap:intro}

Humans invented natural language to communicate with each other. Natural Language Processing (NLP) is a research field dealing with how computers manipulate natural language inputs and outputs \cite{schutze2008introduction}. 
Nowadays, many trending applications are being used daily, such as Siri\footnote{https://www.apple.com/siri/} and Alexa\footnote{https://www.alexa.com/} that rely on NLP tasks; ranging from simple ones like automatic text correction to more complex ones like information extraction and speech recognition. 
Much work has been conducted on various NLP tasks for some significant languages (e.g. English). However, relatively less work is available for other languages used in computing (e.g. Arabic).


Arabic is one of the languages most spoken globally, ranking the sixth, as it is spoken by around 274 million people \cite{eberhand2020ethnologue}. It exists in different forms, such as Modern Standard Arabic (MSA) and Dialectal Arabic (DA). MSA is mostly used in a formal context. DA is used more in everyday life communications (written or spoken) between Arabic speakers. There are several dialectal types of Arabic in different countries, such as, for example, Egyptian, Levantine, Gulf, and Iraqi dialects of Arabic. In addition to the main two forms of Arabic, native speakers switch between their dialect and other dialects or foreign languages in the same conversation, which is known as code-switching (CS) \cite{elfardy2013code}. 
CS is defined as the embedding of linguistic units such as phrases, words, and morphemes of one language into an utterance of another language \cite{myers1997duelling}.
As a result of globalization and better quality of education, a significant percentage of the population in Arab countries have become bilingual/multilingual. This has raised the frequency of CS among Arabs in their daily communications. 
For instance, in Algeria, people tend to code-switch Arabic with French; meanwhile, in Egypt, people tend to code-switch it with English, as shown in the following example of code-switching within the same sentence:
\begin{center}
\setcode{utf8}
 .London \< في > Saturday \< الاتحاد يوم> vs \< كأس السوبر السعودي الهلال  >
 \\(The Saudi Super Cup Al Hilal vs Al Aitihad on Saturday in London.)
\end{center}
As user-generated content increases, it is easy to observe that users mix between different languages in the same word, a practice known as intra-word CS. For example, speakers can say \textit{quizat} which is composed from the English word \textit{quiz} and the suffix \textit{at} referring to plural in Arabic. Multilingual social media users are generating a huge amount of data containing a high frequency of CS. However, such mixed words/sentences are usually overlooked in NLP tasks; as NLP tasks are designed to process texts written in a single language. The rise of social media networks and other informal communication platforms has led to an increased need to apply NLP tasks on CS, and not only DA and MSA.


Though posing several orthographic and morphological challenges, along with the challenges posed by the different dialectal variations, Arabic NLP has achieved many successes and developments \cite{darwish2020panoramic}. There are still considerable gaps for several essential NLP tasks, such as Language Identification (LID) and Named Entity Recognition (NER) tasks. Less work has been done for these tasks on MSA and DA than on other languages, and no work has been done for Arabic-English CS data. 
NER refers to recognizing spans of text that refer to real-world entities and classifying them into different types or categories (e.g., person, location, organization). 
NER has proved to be highly significant to various tasks in NLP, such as information retrieval and question answering tasks. Also, there are several use-cases for the NER task such, as text classification, customer support, content recommendation and semantic annotaion. Applying NER task on Arabic-English CS data faces more challenges than the already existing ones of automatically processing MSA and DA. Some of these added challenges are having two different languages in the same sentence as well as the limited or lack of CS data for the NER tasks. 

In this work, we tackled the challenge to apply the NER task on CS data by presenting several NER taggers and complementing the taggers with other NLP tasks that help in enhancing the performance on such data. Also, we implement CS Intra-word Language Identification (LID) approaches. Language Identification (LID) refers to the task of determining the language type of a text.
Intra-word LID involves segmenting mixed words and tagging each part with its corresponding language identification and states whether a word is a named entity or not. It could be used as a pre-processing task for other NLP tasks.

In this work, state-of-the-art techniques were used to implement the Arabic Name Entity Recognition task and Language Identification of intra-word. The main contributions included: 
\begin{enumerate}[label=\alph*)]
\item Collecting and annotating the required Arabic-English CS corpus for NER task
\item Developing different taggers for MSA and CS NER
\item Applying enhancement techniques to improve the performance of the CS NER taggers
\item Creating Arabic-English CS corpus for intra-word LID
\item Developing different approaches for LID of intra-word CS
\end{enumerate}

\begin{figure}[ht!]
    \centering
 \includegraphics[width=13cm]{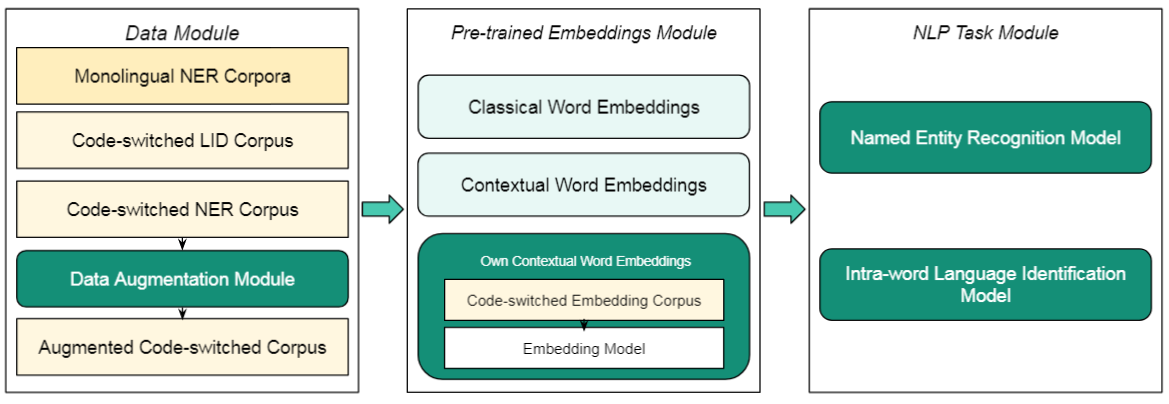}
    \caption{Proposed pipeline approach for applying NLP tasks on CS data}
    \label{fig:sum}
\end{figure}\par


We propose a pipeline approach to apply NLP tasks on CS data as shown in Figure \ref{fig:sum}. To apply the tasks of NER and LID on the CS text, we first collected and annotated data. 
To the best of our knowledge the first two corpora for NER and LID of intra-word code-switching for Arabic-English language pairs, were presented and were composed of 6,525 and 2,507 sentences, respectively.

To apply the NER task, three tagging techniques based on supervised algorithms were presented. We started with Conditional Random Field along with word embeddings for MSA. It was hypothesized that integrating word embedding features to the conventional lexical and contextual features could improve Arabic NER performance. Since most CRF implementations support categorical features only, continuous word embedding vectors were clustered. The best word embedding setting of combining fine and coarse cluster IDs resulted in a 76.4\% F1-score on ANERCorp \cite{benajiba2007anersys} with a relative improvement from the baseline of 11.7\%. Thus, the system achieved the best performance by the following features: Current word, Stemming, Lexical, Contextual, POS tagging, fine and coarse word embedding cluster IDs. 
Afterwards, with the successful use of deep learning models in solving a wide range of NLP tasks, including NER and reaching state-of-the-art performance~\cite{lample2016neural,survey2020}, several deep learning variations were investigated in order to reach the final models with the best performance on MSA and CS data. The usage of different types of classical and contextual word embeddings was also investigated.
The performance of the NER tagger on MSA increased by 7.31\% as compared to the CRF model and the results equalled 83.71\% F1-score. 
Regarding the NER tagger on CS data, we improved significantly the performance of the system by 25.69\% absolute F1-score from its baseline to reach 77.69\% F1-score using the BiLSTM-CRF model and classical and contextual word embeddings. 

 Two enhancement techniques were developed in order to further improve the performance of the CS NER taggers, CS contextual embeddings and data augmentation. While many NLP models such as NER perform well using contextual word embeddings, they face more challenges when dealing with CS data. Recently, bilingual word embeddings drew attention to embedding in the same space words from two languages. Most of the standard bilingual word embedding techniques are intended work on monolingual texts, not on a mix of two languages. Thus, these techniques are not the ideal option to learn embeddings for code-switched tasks \cite{pratapa2018word}. We propose a solution to train for the first time bilingual contextual embedding models used state-of-the-art types on a CS Arabic-English corpus composed of 144 million tokens that we collected, they are represented in the pre-trained embeddings module in Figure \ref{fig:sum}. We also propose a new contextual word embedding model called KERMIT based on the previous work of \cite{devlin2018bert,clark2020electra} capable of mapping both Arabic and English words inside one vector space efficiently in terms of data usage. The highest results achieved by combining Contextual string embedding with Arabic FastText embedding enhanced the NER model by 0.51\%.


Also, an extensive training data-set was needed to improve the performance of the NER system. As stated before, there was a problem of lack of data with the resources suitable for the NER task, especially on CS data. Due to the scarcity of data, several approaches were devised to overcome this issue. One way to produce more data is using data augmentation techniques. We apply three different data augmentation techniques on our CS corpus to automatically create new labeled training data from available ones represented in the data module in Figure \ref{fig:sum}. 
Investigating data augmentation within the NLP field is challenging and less tackled. This was the first time data augmentation techniques have been applied on NER on any CS data due to the complexity of the different languages and the diversity of the various NLP tasks.
Our proposed methods show more of an increase of 1.51\% in the F1-score than the NER model without data augmentation. 


 
 Moreover, we construct the LID model using Segmental Recurrent Neural Networks (SegRNN). We investigate the usage of different word embeddings with SegRNN. Our highest LID system for tagging the entire data-set is obtained using SegRNN alone, achieving an F1-score of 94.84\% and recognizing mixed words with F1-score equal to 81.15\%. Besides, the model of the SegRNN with FastText embeddings achieve the highest results equal to 81.45\% F1-score for tagging the mixed words.

The rest of this thesis is outlined as follows. Chapter 2 discusses the linguistic background about the Arabic language and then different machine learning approaches used with sequential text to apply NER and LID tasks. Chapter 3 presents some related work for code-switching data, Named Entity Recognition, and other related tasks. Chapter 4 presents the different NER taggers on MSA data along with their evaluations and results. Chapter 5 illustrates the data collection and annotation and the NER models of CS data with their evaluation and results. In Chapter 6 and 7, the two enhancement techniques for the NER tagger on CS data of CS contextual embeddings and data augmentation are presented. The corpus of the LID task and the implemented models with their evaluations and results are presented in Chapter 8. Finally, Chapter 10 concludes the thesis and presents recommendations for future work.

%% file: Background.tex
\chapter{Background}\label{chap:back}
This chapter first presents the linguistic background of the Arabic language, including its varieties and challenges in some NLP tasks. 
The traditional machine learning and deep learning approaches are discussed as used with sequential text to apply tasks like NER and LID.

\section{Linguistic Background}
Languages reflect the human mind. It is a way of communication where humans can express themselves. One of the oldest languages is Arabic.  
It is the official language of 24 countries dominantly lying in the middle east, and north Africa \cite{31}.
Arabic is also one of the top nine languages used on the web \cite{egger2016common}. It is a rich morphological language.
 
\subsection{Varieties of Arabic Language}
The Arabic language has three main varieties: Classical Arabic (CA), Modern Standard Arabic (MSA), and Dialectal Arabic (DA). CA was used in the Quran and early Islamic literature. MSA is the formal language in almost all Arab countries. It is used in schools and universities, in the media, and in formal writing such as Arabic newspapers and letters. It is one of the six official languages of the United Nations used in their meetings and documents. DA (Colloquial) is the language used in informal daily communication \cite{shaalan2014survey,darwish2020panoramic}. Within each Arab country and its regions, there are different dialects such as Egyptian, Lebanese and Tunisian. The Arabic dialects themselves differ, sometimes significantly, depending on many factors (e.g., geographical location, social and economic status). 
There is a huge gap between the written form of Arabic MSA and the different Arabic dialects as spoken in different areas due to their significant number. Recently, DA is also being used as the main written language on social media. Nevertheless, the main focus of most NLP research was on MSA. 

In addition to the three main varieties of Arabic, native Arabic speakers typically mix MSA and dialectal Arabic. Due to the presence of many multilingual speakers in Arabic countries, people often mix multiple languages in the same context, known as Code-Switching (CS) \cite{bullock2009cambridge}. 
CS is defined as the embedding of linguistic units such as phrases, words, and morphemes of one language into an utterance of another language \cite{myers1997duelling}. This linguistic behavior (or practice) occurs in both forms of the language, whether spoken or written. The primary language appears the most inside an utterance, while the secondary language used inside the utterance is the language of embedded words or phrases. There are three main types of Code-Switching: Inter-sentential, Intra-sentential and Intra-word.

\begin{itemize}
\item Inter-sentential CS refers to switching between different languages from one sentence to another \cite{bokamba1989there}. For example:\newline 
\setcode{utf8}
That's a great idea! \< ممكن نخرج بكرا. >
\\(We can go out tomorrow. That's a great idea!)
\item Intra-sentential CS refers to using multiple languages within the same sentence \cite{bokamba1989there}.
 For example:\newline 
\setcode{utf8}
 .\< و اخرج > lab  \<  و  >  project 
 \<عندي >
 weekend\<
 بعد ال >
 \\(After the weekend, I have a project and a lab and I will go out.)

\item Intra-word CS refers to mixing between different languages in the same word \cite{Mager2019}. For example:
\newline
The word \textit{quizat} which is composed from the English word \textit{quiz} and the suffix \textit{at} indicating a plural word in Arabic.

\end{itemize}
Researchers use several definitions for CS, such as \cite{muysken2000bilingual} who decided to use the term Code-Mixing (CM) instead of intra-sentential CS. However, other researchers do not distinguish between the different types of CS such as inter or intra-sentential and refer to both types as Code-Switching \cite{Cetino1997}.
The phenomenon of code-switching has been increasingly reported in linguistic studies in the past years as more people tend to code-switch.
Furthermore, this phenomenon has become popular in Arab countries, where people code-switch between different dialects or their own dialect and foreign languages. For instance, it is common to mix Arabic and French in Tunisia and Egyptian Arabic and English in Egypt. CS behavior is common in online interaction, especially among social media users, generating vast CS data \cite{barman2014code}. For example, on Twitter, the multilingual users using code-switching are more active than the monolingual ones and therefore, they produce more code-switched text \cite{hale2014global}. 


\subsection{Challenges of Arabic Language}
Although Arabic is one of the languages that are used the most, it presents several challenges for NLP tasks. 
One of the significant problems that face NER and LID, in general, is the lack of labeled data, a task for which large annotated corpora are needed.  A huge scarcity in available resources exists, especially for dialectal Arabic and Code-Switching. Moreover, Arabic poses several orthographic and morphological challenges, in addition to having dialectal variations \cite{darwish2020panoramic}. 

\subsubsection{Arabic Orthography}
\setcode{utf8}
The Arabic script is one of the main linguistic properties that are the most challenging to the automatic processing of Arabic \cite{farghaly2009arabic}. The alphabet of the Arabic language is composed of 28 letters that are all consonants, three long vowels, and three short vowels. The long vowels are (\< ا >) pronounced (Alef), (\<و >) pronounced as (Waw), and (\<ي >) pronounced as (Ya'a). The short vowels, present in the pronunciation of words, distinguish words from each other, but no unique letters represent them in writing. They are sometimes represented by special marks called diacritics above or below a letter. The majority of Arabic text, especially on social media, does not carry the diacritic marks and this leads to orthographic ambiguity \cite{darwish2020panoramic}. The diacritics give various meanings to the same lexical form. It is easy for a native Arabic speaker to read such words and distinguish between them based on the context surrounding the word. Nevertheless, it is very challenging for a computational system. For instance, the word \<يحيي > without diacritics, it could be considered a named entity of a type person name \textit{Yahya}, or not a named entity and could be a verb (gives life back) or a verb (greets) \cite{zayed2015named}. 

One of the characteristics of the Arabic letters is that they have various shapes according to their position in the word. For example, the letter \<م > \textit{m} has three forms, it is written at the beginning as \<مــ > , at the middle as \<ــمــ >  or at the end as \< ــم > \cite{shaalan2019challenges}. 
The lack of consistency in Arabic orthography poses particular challenges for different NLP tasks such as NER. One of the reasons English NER is easier than the Arabic NER is that most of the names begin with capital letters. These show that a word or its succession is a named entity but that is not an Arabic option. For example, in English, the name \textit{Sara} starts with a capital letter, but in Arabic the same name \<ساره > does not contain any special marks or indications that the word refers to a person name. Moreover, it is common in Arabic, similar to other languages, to face ambiguity between named entities. For example,  \<أحمد أبد > (Ahmed Abad) refers to both a person name and a location name. This is a conflicting situation as the same NE could be tagged as two different NE types \cite{shaalan2014survey}. It is also expected that nouns and adjectives that are not named entities could be mixed up with Arabic nouns that are named entities. For example, the word \<أمل > could mean hope or refers to a name of a person, which might cause ambiguity while recognizing the named entities \cite{shaalan2019challenges}. 


There is a high degree of spelling inconsistency while writing MSA and DA, especially on social media. Furthermore, a non-Arabic word could be transcribed into Arabic, and it will be called Arabizi. There are no specific transcription schemes for such words
as Arabic script has a high level of ambiguity. For instance, the NE "Washington" referring to a city could be transcribed to any of the following Arabic words:  
\<وشنطن , واشنغطن , واشنطن , واشنجطن  > \cite{shaalan2019challenges}. The Western European languages have fewer speech sounds than Arabic, making Arabic complicated by having many NE variants that might be incorrect.






\subsubsection{Arabic Morphology}
The performance of information retrieval and other tasks, including NER, is affected by the representations of the words and their extracted morphosyntactic features. In morphologically complex languages, such as Arabic, it is challenging to extract such features due to its extremely inflectional properties \cite{el2012accurate}. In other languages like English, clitics are treated as separate words. However, in Arabic, they are agglutinated to the words as features for several indications (e.g., number, gender, and mood) \cite{darwish2020panoramic}. The syntactic relationship between words in the sentence is represented by the inflectional endings. Thus, the relationships between words are one of the morphological challenges in Arabic. There is a set of clitics for the Arabic language attached to named entities or words in general, such as prepositions, conjunctions, or both. For example, the word \<وبمصرنا > (and by our Egypt) contains the named entity \<مصر >  (Egypt) of type Location attached to it prefixes and suffixes \cite{shaalan2014survey}. 

Moreover, usually Arabic verbs have a root of three or four characters. There is a template for the derivation of the verbs in Arabic: Verb/Lemma = Root + Pattern; the patterns could make the verb in the past, present/future tense \cite{shaalan2019challenges}. For example, the verb \<يكتب > (future/present form from \textit{write}) is composed of the root \<كتب > and the letter/pattern \<ي >. In addition, it is possible to add 0 or more prefixes and suffixes to form a Word/New Verb: Prefix (es) + Verb/Lemma + Suffix (es) \cite{farghaly2009arabic}. For instance, the word \<سيكتبه > (he will write it) is composed of the verb \<يكتب >, the prefix \<س > and the suffix \<ه >. 
An example that is also not common in other languages is to have one Arabic word that translates to an entire sentence. For instance, the Arabic word \<وسيدرسونها> that is translated to a sentence composed of five English words, \textit{and they will study it} \cite{darwish2020panoramic}. 
 

Thus, in many ANLP tasks, some operations should be applied to process the text and reduce the words to an acceptable abstract form, such as stemming, root extraction, and lemmatization. The stemming is applied to remove the prefixes and suffixes of the words. The root extraction identifies the root/verb of a word composed of three or four letters. The lemmatization is applied to relate a given word to its actual lexical or grammatical morpheme. This process is done by converting the verb to its perfective, third person, singular form and converting the noun or adjective to its singular indefinite form. 
For example, the word \<نحتاجهم > (we need them) after applying a stemming technique will be \<نحتاج > (we need), after root/verb extraction will be \<حوج > (need) and after lemmetization will be \<احتاج > (needed) \cite{el2012accurate}
.
\section{Traditional Machine Learning Approaches}
Machine learning (ML) builds systems that can apply statistical learning techniques to learn from data, identify patterns and make predictions automatically. ML methods focus on extracting meaningful data from narrative text, which is a distinctive sub-field of NLP \cite{manning1999foundations}. Concerning the NER task, the generation of statistical models for named entity predictions is achieved using ML algorithms that determine named entity types from annotated texts \cite{shaalan2014survey}. Machine learning uses learning algorithms that need large training and testing data along with a set of features from these data. There are three categories of machine learning methods, supervised, semi-supervised, and unsupervised. The difference between the three methods is mainly the usage of prior knowledge. Supervised methods learn to predict by training on annotated examples. They are usually used for classification or regression tasks.

Usually a corpus is divided into two or three parts. The first one is the train, used for training the parameters of the model. The second one is the test, used in the final evaluation of the model. The third one is the validation used to reach the optimal values of the model hyper-parameters. Generally, the ML process in the NER task starts by converting the text into structured data, text featurization. One of the best known and currently used text featurization methods is word embeddings discussed in this Chapter as well.

One of the simplest supervised approaches is the Na\"{i}ve Bayes; it is usually used as a benchmark or baseline in several NLP tasks.
Among the traditional supervised machine learning approaches that have proven to be very successful in different NLP tasks, especially in NER, is Conditional Random Fields (CRF) \cite{benajiba2008arabic2}.

\subsection{Na\"{i}ve Bayes Classifier}

Simple Bayesian Classifier is one of the most effective classifications used in machine learning. These probabilistic approaches put relevant assumptions about the problem data, and build a model based on these assumptions. While training the machine learning model the parameters of the probabilistic model are learned. The new input classifications are applied following the Bayes rules. Na\"{i}ve Bayes is one of the classifiers based on Bayes theorem of probability used to predict the class of unknown data-sets. The model assumes that there is no relation between the different features of the input and ignores the correlations. The label or class \emph{Y} could be related to only one feature node. This assumption is not accurate in most of the tasks; however, it makes the task simpler especially when dealing with a large number of attributes \cite{mccallum1998comparison}. 
 



\subsection{Conditional Random Fields}


 CRFs are one of the main statistical machine learning techniques. They are used as sequence classifiers to segment and label sequence data based on probabilistic models \cite{lafferty2001conditional}. The probabilistic sequence classifiers compute a probability distribution over the possible labels to choose the best label sequence of units (e.g., words, sentences, letters).
CRF could be considered as an enhancement or generalization of Hidden Markov Models (HMM) and Maximum Entropy (ME) \cite{lafferty2001conditional}. 
The HMM considers the observed events such as the words available in the input and the causal factors in the probabilistic model that are hidden, such as part-of-speech tags \cite{miller1999hidden}. The labeling decision is only dependent on the current input with its corresponding observed object. In general, for the NER task, the goal is to predict label \emph{Y} for input \emph{X}. The HMM is a generative probabilistic model which specifies the joint distribution: $p(X,Y)$. Nevertheless, the CRF is a discriminative undirected graph model, and specifies the conditional distribution of the \emph{y} labels given \emph{x} inputs: $p(Y|X)$ \cite{galliani2017gray}.

The ME method calculates estimated probabilities relying mainly on the imposed constraints while making a few other assumptions. The constraints are deduced from the training data, and they express some relationships between features and output \cite{berger1996maximum}. This method selects the probability distribution with the highest entropy that satisfies the previous property \cite{chieu2003named}. While the ME model considers the dependencies between neighboring states, CRF calculates in a single model all the state transitions globally to avoid the Label Bias problem of ME \cite{wallach2004conditional}.

 \begin{figure}[h]
 \centerline{\includegraphics[width=6cm]{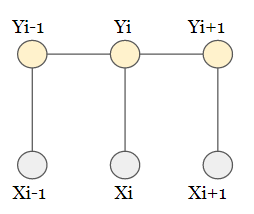}}
 \caption{Graphical structures of CRF for sequences}
 \label{fig:crf}
 \end{figure}

One of the main advantages of CRF is the ability to consider contextual information before assigning a label to a word. The input feature states of the CRF model are the sequence of features (e.g., capital, the previous word, the following word) given in the same order as the input sentence, and the features correspond to the tokens. The sentence-level information is included in the model by selecting the output sequence that maximizes the possibility of the whole sequence \cite{lafferty2001conditional}. N-gram algorithm and other available NLP techniques consider words as atomic units that do not have anything in common, which results in simplicity and capability to train a large amount of data. However, these techniques mostly recognize words available in the training data \cite{mikolov2013efficient}. CRF models can predict many interdependent variables, which is needed in the NER task. Thus, CRF would overcome some of the main challenges of NER. The first such challenge is that some entities are rare and do not appear in the training set and should be identified based on context. The second one is the ambiguity problem of named entities, which could be solved by considering the context of the entity (previous or following word) \cite{patil2020named}.

\section{Deep Learning Approaches}
Recently, in several NLP tasks, the state-of-the-art performance was achieved using deep learning techniques, a subset of machine learning. The main advantage of deep learning models is automatically extracting complex features from input data instead of the manual handcrafted feature engineering used in other ML methods. It also refers to Artificial Neural Network (ANN) with multiple complex neural network layers.

\begin{figure}[h!]
  \centering
  \subfloat[Neural Unit]{\includegraphics[width=140pt]{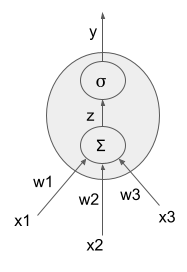}\label{fig:ANN1}} 
  \hfill
  \subfloat[Feed Forward Neural Network]{\includegraphics[width=180pt]{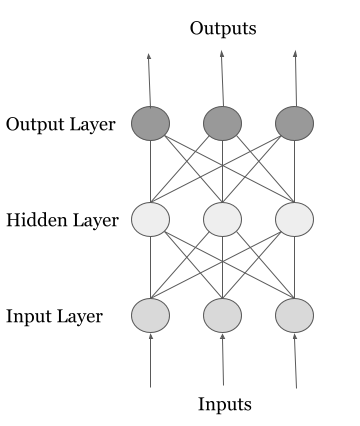}\label{fig:ANN2}} 
\caption{ANN Basic Architecture }
\label{fig:ANN}
\end{figure}

ANN is inspired by how the human brain processes information. The central computational units of a Neural Network (NN) are artificial neurons where directed edges interconnect them. The connections between the neurons are represented by the weights, which determine the impact of one neuron on another \cite{kozyrev2012classification}. ANN is designed to recognize patterns and detect trends represented as numerical values in vectors, into which all input data should be translated. The neural unit, as shown in Figure \ref{fig:ANN1} takes as input vector \emph{x} having weight \emph{w} expressing the importance of this input to the output and \emph{z} is the weighted sum of inputs. The output of the unit is \emph{y} which is based on the activation function; in this example, it is the Sigmoid function \cite{abiodun2018state}. The activation function determines whether and to what extent the input should progress and affect the output of the network. The weights \emph{w} are adjusted based on an error signal/feedback during learning to find the desired output.
The selection of the type of activation function is made after experimenting with several functions and selecting the one that gets the best results on the validation data.
 
There are several architectures of a NN, and the most common and straightforward type is the Feed Forward Neural Networks \cite{schmidhuber1989local}.
The architecture of this network is composed of the first layer representing the input, the final layer representing the output, and the intermediary layers are hidden layers as shown in Figure \ref{fig:ANN2}. The layers are composed of neuron units/nodes; a specific layer can have an arbitrary number of nodes called bias nodes. The values of the bias nodes are equal to one, which provides the node with a constant value that is trainable with the inputs \cite{abiodun2018state}. Also, the value of the bias gives an indication to the activation function to move either right or left. 

As stated before the general training/learning process of a neural network requires dividing the data into three sets. The first one is the training set which allows the network to understand the weights. The second one is the validation set; the network uses it to fine-tune the performance. The last one is the test set, which is used to measure the performance and error margin of the network. The training of the network starts by forwarding the propagation of the input information to a hidden layer(s) through an output layer to calculate the loss value. Then the errors are backpropagated \cite{rumelhart1988parallel} from an output to an input layer via hidden layer(s). The last step is updating the values of the parameters, such as the weights and biases based on the feedback calculated by training errors during back-propagation. 
The minimization of the error function could be achieved through the gradient-based optimization algorithms \cite{ruder2016overview}. One of the main disadvantages of Feed Forward Neural Networks is that they are not a good option for sequence data as they do not have adequate memory and cannot remember historical input data. 
Different Deep Neural Network architectures are used for various tasks and data modalities. In general, these are three common architecture types: Convolutional Neural Network used for spatial analysis, Recurrent Neural Network used for sequential analysis and the Transformers.





\subsection{Convolution Neural Network}
The Convolutional Neural Networks (CNN) is a deep learning algorithm commonly used in the Computer Vision field \cite{lecun1998gradient}. It could be considered as a specialized Feed Forward Neural Network. 
It takes the input image and specifies learn-able weights and biases to different objects in the image to assess their importance. CNN checks four key ideas which are shared weights, local connections, use of many layers, and pooling layers. The CNN automatically extracts features from the image by applying a collection of filters to create a final hierarchical structure of features. Each filter has a weight that is learned from the training data \cite{behnke2003hierarchical}. 
\begin{figure}[ht!]
  \includegraphics[width=14cm]{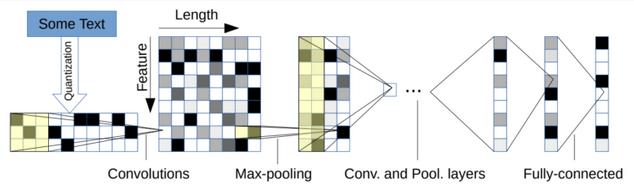}
  \caption{Character-based CNN for text classification \cite{zhang2015character}}
  \label{fig:cnn}
\end{figure}
CNN is also used in different NLP tasks with text input \cite{kalchbrenner2014convolutional}. The process starts by sliding filters of different window sizes over the input that could be word embeddings. Each filter has a weight and generates a new feature for the different windows. A feature map is generated by sliding the filter over each window. As a result of some calculations that consider a small segment of the input sequence and share the parameters with the calculations to its left and right to generate each entry in the feature map. In order to condense a feature map to its most important feature, a pooling technique is used. Max-pooling is one of the most common types of pooling. It takes all maximum values of the feature maps and concatenates them to form a vector. This vector is given to the next layer or output layer at the end \cite{ruder2019neural}. 

Figure \ref{fig:cnn} illustrates an example of CNN model as applied on text. The model takes as input the text/word embeddings and aggregates the local information from the neighbours to save the meaning of a word by applying a convolution operations. 
This network is composed of one large and one small CNNs. These are composed of nine layers deep with six convolutional layers and three fully connected layers.

\subsection{Recurrent Neural Network}
Recurrent Neural Network (RNN) \cite{elman1990finding} is a specific kind of ANN that has internal memory designed to remember its past input every time a new input is given. It is composed of neurons connected by weighted arcs \emph{w} and models sequential data better than Feed Forward Network as it models the relationships between the inputs over time using a feedback mechanism/loops \cite{abiodun2018state}. For every element of a sequence, it performs the same task with the output while depending on previous computations. 
\begin{figure}[!ht]
  \includegraphics[width=\linewidth]{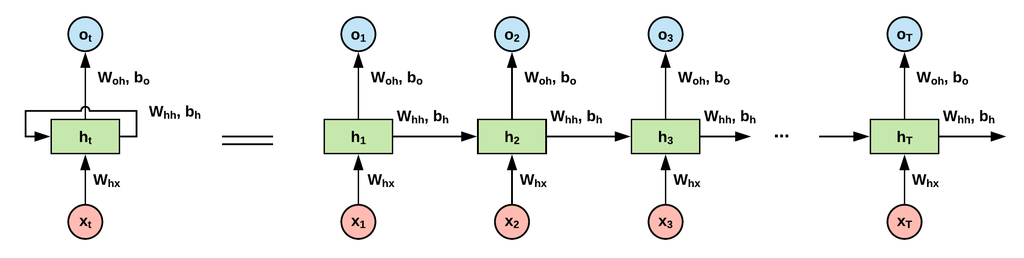}
  \caption{Overall visualization of RNNs \cite{RNN2019}}
  \label{fig:rnn}
\end{figure}
RNN is commonly used in the NER and LID tasks and has showed success due to its capabilities of handling sequential tasks \cite{li2015visualizing,gupta2016table}. As illustrated in Figure \ref{fig:rnn}, the input vector is \emph{x\textsubscript{t}} that represents the current input and \emph{h\textsubscript{t}} represents the current hidden timestamp layer. \emph{W\textsubscript{hx}} is a weighted matrix of current timestamp layer that is used in multiplication of the input vector and it is given to the activation function. The function computes the activation value for the hidden timestamp layer. This hidden layer is used to calculate the corresponding output \emph{o\textsubscript{t}} using weight at output state \emph{W\textsubscript{hy}} and hidden state. This RNN architecture does not require a fixed length limit prior to context because each element is processed one by one at a time. Nevertheless, this affects the parallelism of the execution of this architecture. RNN learns long-distance dependencies because of its maintaining memory-based history information. However, practically, they fail due to the vanishing or exploding gradient \cite{ma2016end}. Thus,
a series of RNN variants have been created in order to solve this problem. 
\begin{figure}[ht]
     \centering
  \includegraphics [scale=0.3]{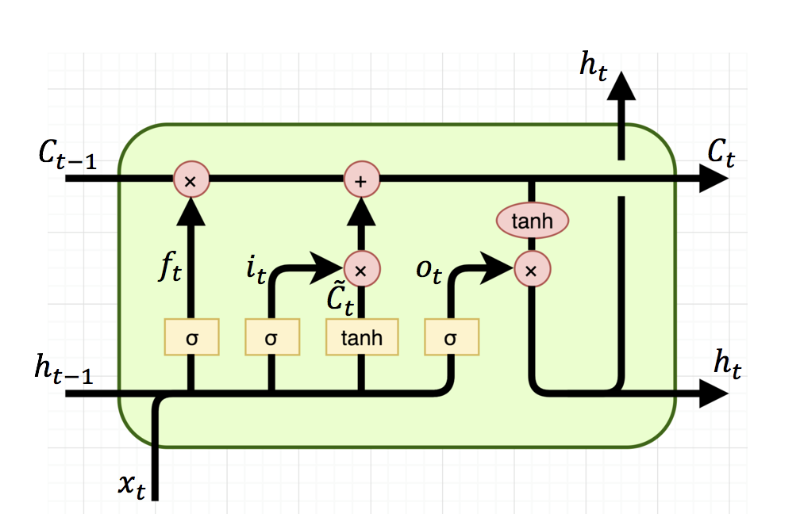}
  \caption{Gates of LSTM unit layer \cite{LSTM}}
  \label{fig:lstm}
\end{figure} 
One of them is Long-Short-Term-Memory (LSTM) that was designed to solve the vanishing problem of RNN. \cite{hochreiter1997long}. LSTM network contains connected memory blocks/cells instead of the traditional nodes of the RNN. They retain information for a longer time, needed in NLP, to model long-term dependencies. LSTM has mechanisms to decide what information should be remembered and what should be forgotten. As shown in Figure \ref{fig:lstm}, the LSTM network augments the RNN architecture with an input gate \emph{i\textsubscript{t}}, forget gate \emph{f\textsubscript{t}} and an output gate \emph{o\textsubscript{t}}. All gates share a common design pattern; they contain a feed-forward layer followed by a Sigmoid activation function and by a point-wise multiplication with the layer being gated. They are all functions of the current input \emph{x\textsubscript{t}} and the hidden previous state \emph{h\textsubscript{t}}. These gates interact with the current input, its cell state \emph{c\textsubscript{t}} and the previous cell state \emph{c\textsubscript{t-1}} and allow the model to either retain or overwrite information.The forget gate \emph{f\textsubscript{t}} is capable of removing context that is not needed. It computes the weighted sum of the previous hidden layers and the current input and passes that to a Sigmoid activation function. The context vector is multiplied to the output to remove the unneeded information \cite{gers1999learning}. However, the disadvantage of using LSTM is that it does not check the future context as it checks the previous one only. 

\begin{figure}[ht]
     \centering
  \includegraphics [scale=0.65]{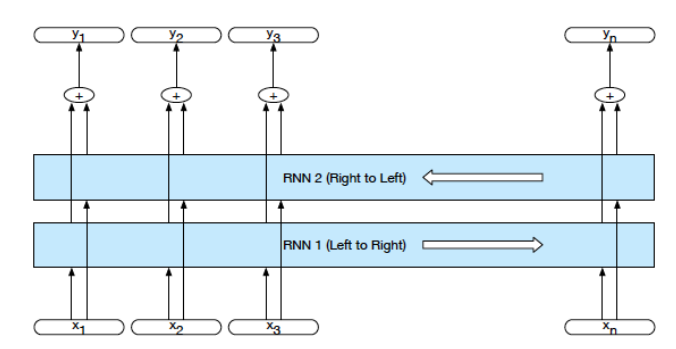}
   \caption{Bidirectional LSTM Architecture \cite{jurafsky2019speech}}
  \label{fig:LSTMBiLSTM}
\end{figure}
To benefit from the previous and from future contexts, Bidirectional Long Short-Term Memory Networks (BiLSTM) was proposed as an extension to LSTM. This represents two separate LSTMs each one representing a sequence forward and backward to save the previous and future information as shown in the Figure \ref{fig:LSTMBiLSTM}. It takes advantage of both left and right contexts and is considered as pair of LSTMs, the first one is trained from left-to-right and the second one is trained from the right-to-left. This has showed promising results in different NLP tasks \cite{yin2017comparative}.

\begin{figure}[ht]
     \centering
  \includegraphics [scale=0.65]{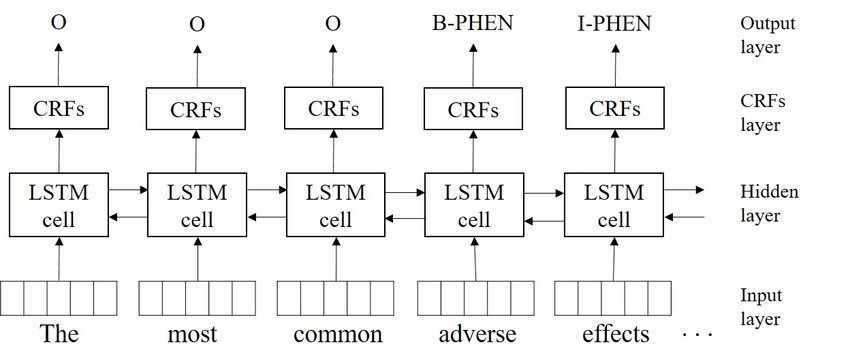}
  \caption{Bidirectional LSTM with CRF layer Architecture for NER \cite{kim2019bootstrapping}}
  \label{fig:BiLSTMCRF}
\end{figure}

For the sake of predicting the current tags, there are two common ways to make use of the previous and the future tag information. One way is to predict a distribution of tags, step by step, and then use beam-like decoding to find the best sequence of tags; this could be achieved using the Maximum Entropy Markov model. The other way is to use the CRF model which focuses on sentence-level instead of individual positions \cite{huang2015bidirectional}. As stated before, CRF is one of the most conventional high-performance sequence labeling models. Combining LSTM/BiLSTM with a CRF layer has an advantage over LSTM/BiLSTM alone and CRF alone. As shown in Figure \ref{fig:BiLSTMCRF} CRF layer on top of BiLSTM will add the sentence level tag information to the model. Then, the CRF Layer can efficiently predict the current tag from past and future tags, equal to the BiLSTM past and future input features. These extra features can boost tagging accuracy \cite{huang2015bidirectional}. 
Combining both LSTM/BiLSTM and CNN networks is also used to solve sequential problems. It takes advantage of both models, the CNN extracting the features from the input text, and the LSTM/BiLSTM saving the chronological order of the input words 

\subsection{Transformers}
\begin{figure}[!ht]
  \includegraphics[width=\linewidth,scale=0.5]{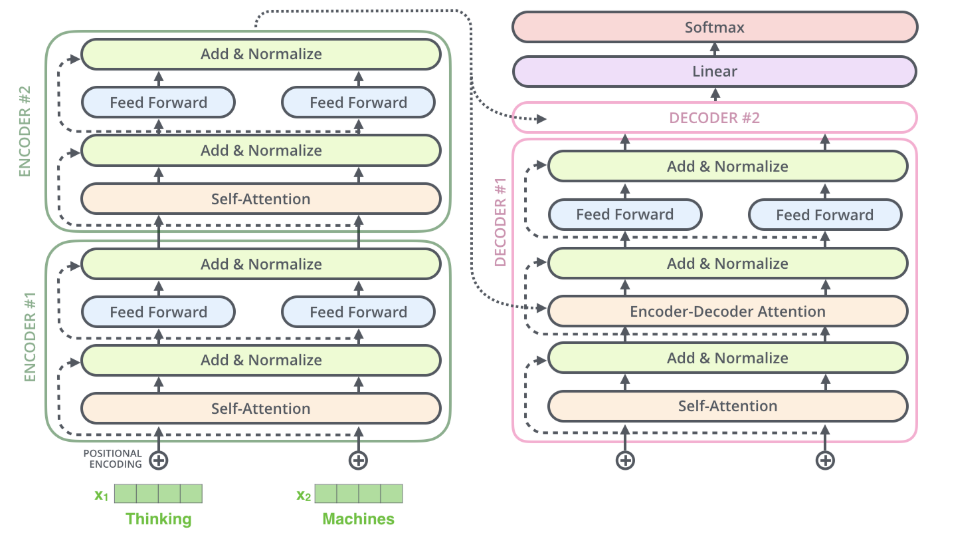}
  \caption{Transformer Architecture\cite{Transf}}
  \label{fig:transformer}
\end{figure}

Transformers are a recent type of neural network architecture \cite{vaswani2017attention}. They have become a mainstream architecture used in several NLP tasks as they are very powerful. The transformer models solve the disadvantages of the LSTM models as they allow parallelism in training and capture long term dependencies of tokens in a sequence. In comparison with BiLSTM models that cover the full context of a sequence by concatenating two models in one vector, the transformers monitor in one model the sequence bidirectionally. This method is better than the concatenation as it gives better representation of sequence. This is achieved by modeling direct dependencies between each two words in a sequence. Transformers model dependencies using an attention mechanism. As shown in Figure \ref{fig:transformer} its architecture depends on stacked layers of self-attention and point-wise in forming encoder and decoder components; the encoder is the component on the left and the decoder is the one on the right. 
Few concepts must be defined in order to better understand  the architecture of the transforms such as, for example, attention, self-attention, encoder and decoder.
\subsubsection{Attention}
As proposed by \cite{luong2015effective}, attention is based on context-encoding mechanism to overcome the drawback of the context vector mechanism of the LSTM model. The idea behind this mechanism is that, each time a model outputs a hidden state, the context from part of the input sequence where most relevant information is used. 
\subsubsection{Self-Attention}
Transformers use the mechanism of self-attention to find words in a sentence that are relevant to the current processed word. This is achieved by calculating a score for the tokens in a sequence to represent its relevance. Then after calculating the scores for every token, all the scores are added to get the output for the first position token. 



Another mechanism used by the transformers is the Multi-Headed Attention which enhances the performance of self-attention. The attention layer is given different heads, each with different parameter values. Using the given values the head computes the output attention. After all the outputs of the heads are concatenated to form the final output of the attention layer.

\subsubsection{Encoder}
One of the main components of the transformer is the encoder. It is created from a stack of N identical layers, with each one being composed of two sub-layers. The first sub-layer is composed of a multi-head self-attention mechanism. The second one is a simple position wise fully connected to a Feed Forward network followed by a normalization layer. A corresponding hidden state is passed to the decoding stage for each input vector. As shown in the Figure \ref{fig:transformer} the vectors of the input pass through positional encoding nodes. An order representation is given to the input vectors by these nodes that add positional encoding vectors.
\subsubsection{Decoder}
The final main component is the decoder which is also created from a stack of N identical layers, with each one being  composed of two sub-layers. However, a third sub-layer performs multi-headed attention on the output of the stack of the encoder. The output of the decoder is converted to word format using a linear layer followed by a Softmax layer.

\subsection{Configuration of Deep Neural Network}

Model design variables determine the network structure, such as the number and size of the hidden
layers, and the hyper-parameters determine how the network is trained, such as learning and dropout rate \cite{diaz2017effective}. The following are some model variables and hyper-parameters that affect the training of deep learning models.

\textbf{Epoch:} It is a random cutoff number, usually defined as "one pass over the whole dataset''. It is used to separate training into different phases, which is useful for logging and for periodic assessment \cite{11}.

\textbf{Batch Size:} It is the number of sentences that will be propagated through the network. The training dataset will be divided by the number of batch sizes. Small batch sizes are attractive since they can make convergence in fewer epochs. However, large batch sizes provide more data-parallelism, which successively enhances computational efficiency and scalability \cite{22}.

\textbf{Activation Function:} 
It is used to establish non linearity to models, which allows deep learning models to learn non-linear prediction. It calculates the output from the summation of the weighted input signals of the neural network and also maps the result between  0 to 1 or -1 to 1, depending on the function. The main reason for using it is to transform the input signal in a network into an output signal. It is a mathematical equation that is responsible for determining the output of a neural network. Each neuron has a function attached that determines whether it should be activated or not. This function is applied to the summation of the product of input nodes and their weights. A Neural Network unaccompanied by the Activation function would directly be a Linear regression model, which will not fulfill learning complicated functional mappings from data \cite{32}. The following are popular activation functions that are experimented with in order  to find the best match for the different deep learning models, Softmax, Softplus \cite{glorot2011deep}, Softsign, Relu \cite{nair2010rectified}, Tanh, Sigmoid, Hard-Sigmoid, and Linear \cite{act}.

\textbf{Optimization Function:}
This function is used to minimize the output of the error function. It depends on the internal learnable parameters of a model applied to the input to compute the predicted output. The internal parameters significantly impact and efficiently train the model and process accurate outcomes \cite{32}. One of the commonly used optimization functions is the Stochastic Gradient Descent (SGD). However, it is not easy to state the learning rate as it affects its performance. To overcome its disadvantages other functions were proposed such as Adagrad \cite{duchi2011adaptive}, RMSProp \cite{hinton2012neural}, Adadelta \cite{zeiler2012adadelta}, Adam \cite{kingma2014adam}, Adamax \cite{kingma2014adam} and Nadam \cite{dozat2016incorporating}.

\textbf{Dropout:}
This function is used to tackle the overfitting problem due to the noise found in the training dataset but not in the test dataset. As deep neural networks have various non-linear hidden layers, which make the model an exceedingly expressive model. They can learn very complicated relationships between the outputs and the inputs. It is an averaging technique to combine the exponential number
of hidden layer architectures, each sharing the same weights. Some of the complicated relationships will be the outcome of sampling noise due to the limited training data \cite{26}. 

\textbf{Learning Rate:} This function specifies how fast a network updates its parameters. A lower value makes the network train faster; however, it can miss the minimum loss function \cite{diaz2017effective}. 

\section{Word Embeddings }

One of the powerful developments in the NLP field is word embeddings. This is used to convert input text to a machine-readable format. One of the traditional techniques to do so is one-hot encoding. This represents each sequence text input in \emph{d} dimensional space, where \emph{d} is the vocabulary size in the dataset. If the term is presented in the document, it will get 1 and 0 otherwise. In the case of a large corpus, this method will generate huge vectors that are very sparse and inefficient. Thus, later word embeddings were introduced where each word is projected to a dense vector; of short length and most elements are non-zero. These word embeddings preserve the semantic distance between words, and semantically similar words are grouped near each other. Word embedding or representation is done by mapping every word \emph{wi} in a sentence to vector \emph{xi} in multi-dimensional space. The word embeddings are stored in a matrix \emph{X}.
Thus, an input represented as a sequence of words \emph{w1,...,wt} is represented as a corresponding sequence of word embeddings \emph{x1,..,xt}, that is given to the neural network. 

As it is hard to identify the similarity between the same Arabic words having different prefixes and suffixes, the usage of word embedding makes it easier to find these similarities as similar words are mapped to nearby vectors. For instance, the words \<الدولة> (the country), \<الدولتان> (the two countries), \<الدول> (the countries) differ in their morphological forms, however, their embedding should be placed near each other in the space. 


Large datasets are used to train and save the embeddings. Then these could be used as pre-trained word embeddings models in various downstream NLP tasks to boost their performance. Moreover, they avoid training an embedding model from scratch as learning independent representations for words from the training data alone is difficult \cite{qiu2020pre}. Thus, adding word embedding to deep learning models enhances the performance by specifying syntactic and semantic word relationships. There are two main categories of word embeddings: classical (non-contextual) and contextual word embeddings. The main difference between these two categories is whether the context of a word affects its embedding and changes it or not. It is challenging to learn high-quality representations because of the different characteristics of the words and changes they undergo based on the linguistic contexts \cite{peters2018deep}.

Classical word embeddings do not consider the context of the words, and the generated vectors are static; they do not change based on the context of the word. For instance, the word \textit{``apple''} has a different meaning in the following two sentences \textit{``I want to eat an apple''} and \textit{``Apple store is very crowded today''}. Thus, using classical word embeddings will generate the same vector for the word ``apple'' in both sentences. Therefore, 
to overcome this problem of ambiguity, contextual word embeddings are being introduced, and several types are being proposed in the NLP field. Contextual word embeddings generate different embeddings for the same words based on their context, which is essential to capture the semantics of the ambiguous words based on their context \cite{peters2018deep}. In this example, ``apple''  will be given different vectors in both sentences. This will help explain that it refers to fruit and in the second one, to a store or organization in the first sentence. The following are some of the embeddings types most used in various NLP tasks.

\subsection{Classical Word Embeddings}


There are two main classes of embeddings, word-level and character-level. The Word2vec and GloVe embeddings will be discussed next from a word-level class and FastText embeddings from the character-level class.


\subsubsection{Word2vec (W2V)}
Word2vec (W2V) learns word representations using neural networks. The architecture of W2V is a Feed Forward Neural Network with one hidden layer. The representations are created by training a classifier to distinguish nearby and far-away words. The two main prediction/learning approaches commonly used in W2V are continuous Bag-of-Words (CBOW) and Skip-gram. CBOW predicts the current word based on the given context. In comparison, Skip-gram predicts surrounding words given the current word. W2V provides multiple degrees of similarity between different words by mapping to nearby vectors \cite{mikolov2013efficient}.
 The most popular pre-trained Word2Vec embedding models were developed by Google. 
It was trained on 100 billion words of Google News data-set.

\subsubsection{GloVe} It is another widely known model for generating pre-trained embeddings. Deriving the relationship between words from global statistics is the main idea of Glove. The training of this algorithm is computed on aggregated global word-word co-occurrence statistics in a large amount of textual data.
One of the English pre-trained models of GloVe \cite{pennington2014glove}, was trained on Wikipedia 2014 and Gigaword 5 data.
Both Word2vec and Glove word embeddings perform similarly in most applications. However, the training of Glove is based on matrix factorization, thus, it could be easier parallelized. 
Besides, both of them share the same main limitations, one of which being the problem of out-of-vocabulary words, which is solved by introducing the character-level embeddings such FastText.

 \subsubsection{FastText} It was created by Facebook as an extension of W2V. FastText encodes character-level representations for a word. It breaks words into N-grams, and the word embedding is computed as the sum of all these N-grams. This capability helps FastText in outperforming W2V in modeling word representations. Several pre-trained models trained on Wikipedia were released for several languages, including English and Arabic. The main advantage of FastText, as compared to W2V, is that it can generate embeddings for unseen words, words not processed during training from their character n-gram features  \cite{bojanowski2017enriching}. 

\subsection{Contextual Word Embeddings}
Contextual word embeddings are considered the new approach for representing the vector of a word in a given text, as compared to Word2Vec and GloVe models. 
These embeddings provide different vector representations of a single word and are derived from pre-trained bidirectional language models on a large text corpus \cite{peters2018dissecting}. The contextual embeddings perform better when being used in a downstream like NER on ambiguous, complex, and unseen languages \cite{arora2020contextual}.  
In the following part several types of contextual embeddings will be discussed, ELMo, BERT, Contextual string embeddings, Pooled FLAIR embeddings, ELECTRA and MUSE. 


\subsubsection{ELMo} 
Embeddings from Language Models (ELMo) was one of the first contextualized word embeddings based on language modeling trained using BiLSTM. The task of language modeling is unsupervised learning used in ELMo in the pre-training phase. ELMo combines all layers with a weighted average pooling operation. A language model generates the following word based on the previous words in the sentence. The purpose is to distribute over sentences in the original training data as close as possible to the resulting distribution. The internal representations from the BiLSTM language model is transferred after pre-training on a large dataset to form ELMo embeddings \cite{peters2018dissecting}. 
Several pre-trained ELMo models for several languages include the Arabic language implemented by \cite{peters2018deep}.

\subsubsection{BERT} 
\begin{figure}[!ht]
\centering
  \includegraphics[scale=0.75]{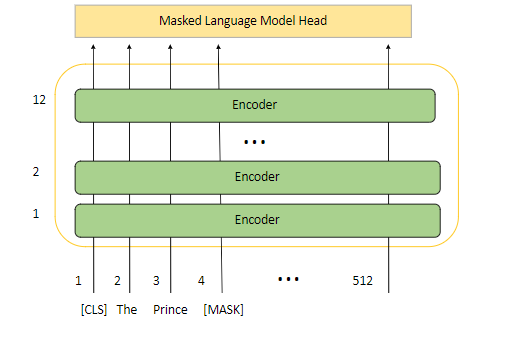}
  \caption{Shows data-path of input tokens inside BERT model \cite{BERTI}}
  \label{fig:path}
\end{figure}

Bidirectional Encoder Representations from Transformers (BERT) is a transformer-based language model. It replaces language modeling with Masked Language Model (MLM) and Next Sentence Prediction (NSP) tasks. MLMs are trained on masking out random tokens by replacing them with a special token [MASK] inside a sentence. The model then tries to predict the masked
tokens using the whole context of a sequence. Besides, NSP models are trained on distinguishing whether two input sentences
are continuous segments or are separate \cite{devlin2018bert}. BERT models sentences as a sequence of tokens. Each input token in a sequence flows through stacked encoders and outputs a hidden state representing the word embedding. This enables BERT to train in parallel. The data-path of input sequence is shown in Figure \ref{fig:path}   

\begin{figure}[!ht]
\centering
  \includegraphics[scale=0.75]{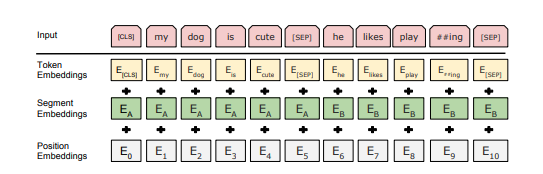}
  \caption{BERT input representation \cite{devlin2018bert}}
  \label{fig:layeremb}
\end{figure}

 The size of the model can vary according to the number of stacked encoders and hidden size. Larger BERT models require more data to train without overfitting. BERT models train on a fixed length sequence. In the tokenization phase, BERT uses the WordPiece model to tokenize words. A token that is not in its vocabulary is broken by the WordPiece model into segments that the model can tokenize. Special tokens [CLS] and [SEP] are added in the tokenization process to split a pair of sentences. Finally, a special token [MASK] replaces the tokens that are being predicted. Then three embedding layers are added to the input vectors before feeding them to the encoders as shown in Figure \ref{fig:layeremb}. A static positional embedding is added to each token to indicate its position in the sequence. Segment embedding designed to help distinguish which sentence a token belongs to when a pair of sentences is inputted, helps in handling a variety of NLP tasks. At last token embedding transforms the input representation into fixed-size vector.
BERT was state-of-the art in 11 NLP tasks, including NER \cite{devlin2018bert}. One of the main pre-trained BERT models is the BERT Multilingual model, which contains 104 different languages, including Arabic and English. 
\subsubsection{Contextual String Embeddings} It is a contextualized character-level/string bidirectional language model. Contextual string Embeddings model sentences as a sequence of characters and trains on the auto-regressive task. Contextual string Embeddings architecture is an Encoder-Decoder architecture variant. The first layer is the encoder layer which is the embedding layer for the model. Then hidden layers are LSTM layers. Finally, a decoder layer which is a fully connected dense layer for output. FLAIR can model sequence bidirectionally by stacking a forward and a backward character language model. Therefore, to get the whole context, the forward and backward FLAIR language models should be trained. Besides considering the context, these language models are trained without the explicit notion of words which lead to modeling the words as a sequence of characters. Its models can produce several embeddings for the same word based on its context and manage dealing with rare and misspelled words by modeling context as characters \cite{akbik2018contextual}.  


A variant of Contextual string Embeddings embeddings is the Pooled FLAIR embeddings. It has an identical architecture of Contextual string embeddings. The difference lies in having an additional memory component. This component solves the  disadvantage of Contextual string embeddings by generating meaningful embeddings of rare strings used in an under-specified context. Pooling operation aggregates the contextualized embeddings of all unique strings it finds and retrives previous embeddings produced from memory. Then, the pool operation is performed and defines one-word embedding for all contextualized instances \cite{akbik2019pooled}.

\subsubsection{ELECTRA} 
\begin{figure}[!ht]
  \includegraphics[scale=0.15]{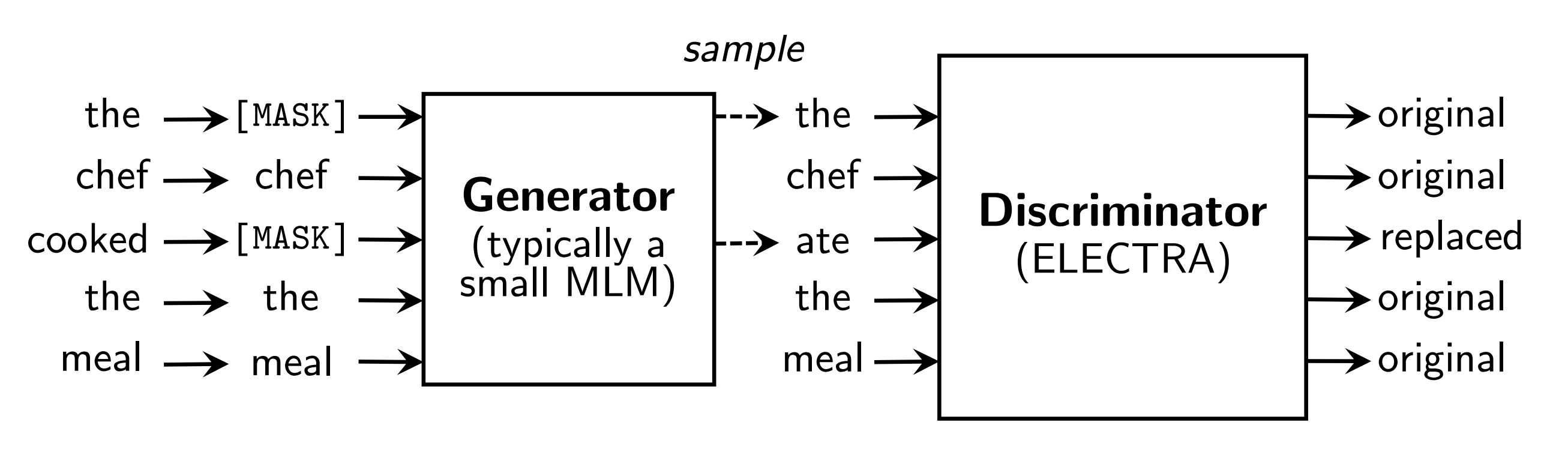}
  \caption{ELECTRA generator and discriminator components \cite{clark2020electra}}
  \label{fig:evis}
\end{figure}

Efficiently Learning an Encoder that Classifies Token Replacements Accurately (ELECTRA) is composed of two neural networks, a generator and a discriminator trained together on MLM and Replaced Token Detection (RTD) tasks. RTD is a particular unsupervised task that trains discriminative models. The architectures of the discriminator and generator are encoder variants of transformers like BERT. The language model is given a sequence of tokens from a generator and tries to predict whether a generator or original token replaces a token. A full modeling context does this. The ELECTRA model is trained to minimize the combined losses of the generator and the discriminator. The generator model is only used in pre-training. In the fine-tuning stage, the generator model is thrown away and the discriminator model is used for downstream tasks. Figure \ref{fig:evis} visualizes the inner components of ELECTRA.

An input sequence first passes through a generator that is identical to BERT. The three embedding layers in Figure \ref{fig:layeremb} are added to the input sequence. The generator inputs are corrupted by replacing random tokens with a special token [MASK] which the generator is trained to predict. Then, using the attention mechanism, the encoder layers output hidden states of the embedding representation. The output layers of the generator predict the tokens in the sequence using the embedding representation computed. Next, the discriminator model takes the computed representation as input. The discriminator model has as well its own three embedding layers that add vectors to the input tokens. The encoder layers then compute attention and finally, discriminative output layers distinguish tokens in the data from those that have been replaced by the generator model. The ELECTRA model is trained to minimize the combined losses of the generator and the discriminator. Compared with BERT, the ELECTRA training mechanism is considered to be more efficient because the task is defined over all input tokens rather than just the 15\% tokens that were masked out \cite{clark2020electra}.

\subsubsection{MUSE}
One of the recent embedding types is the Multilingual Universal Sentence Encoder (MUSE) \cite{yang2019multilingual}. Being one of the members of the Universal Sentence Encoder (USE) embedding models \cite{cer2018universal}, it maps text written in different languages having the same meanings, to nearby embedding space representations. To align the cross-lingual vector spaces, a translation ranking task is used by MUSE. A sentence of any length is mapped by MUSE to a vector with dimensions equal to 512. MUSE has multilingual models that support 16 languages, one of which being Arabic. The available models are trained on generic corpora from different sources such as Wikipedia and contain vocabulary of
200,000 sub-words. 



\section{Evaluation Metrics} 

It is essential to discuss the evaluation metrics commonly used to evaluate and compare the different NER techniques and models. Evaluating a system requires a comparison between the outputs with the gold-standard annotations. The gold-standard corpus contains annotated instances with the same types of named entities, and this corpus is usually tagged manually. The process starts by randomly dividing the corpus into training and testing sets with portions of 70\%-30\% or 80\%-20\% respectively and is followed by the learning process of the model that uses the training set. Then entities are extracted from the test set using the trained model. For each named entity type several evaluation metrics could be calculated.

The different NER forums have suggested various evaluation schemes; in this work, the exact-match evaluation suggested at CoNLL-2003 \cite{tjong2003introduction} is followed. 
A simple method used to analyze the success rate and denote the right and wrong predictions is the confusion matrix. The rows of the matrix represents an actual class and the columns represent a predicted class.  
\begin{table}
\begin{center}
\begin{tabular}{|l|l|l|}\hline
 &  \textbf{True label positive} & \textbf{True label negative}\\
\hline
\textbf{Predicted label positive} & True Positive (TP) & False Positive (FP)\\\hline
\textbf{Predicted label negative} & False Negative (FN) & True Negative (TN) \\ \hline

\end{tabular}
\caption{Confusion Matrix}
\label{table:confusion} 
\end{center}
\end{table}
As shown in Table \ref{table:confusion}, the possible classification cases are denoted as $TP$ and $TN$ representing the correctly classified number of positive and negative named entities as well as $FN$ and $FP$ representing the misclassified negative and positive named entities, respectively. The information given by the confusion matrix is not enough to evaluate the performance and more concise metrics are used. 

There are four common metrics that use numerical values to represent several aspects of the quality of a system, Accuracy, Precision, Recall, and F1-score (micro-averaged). 
The Accuracy is a common evaluation metric in machine learning classification tasks. However, it gives high values to systems that do not return any results. Accuracy is the percentage of correct predictions from the total predictions made by the system. 
\begin{equation}
Accuracy = \frac{TP + TN}{TP + TN + FP + FN}
\label{eq:accuracy}
\end{equation}
The Precision is the percentage of named entities recognized by the correct system. It is calculated by computing the ratio of several correct answers to the total number of answers as shown in Equation \ref{eq:pre} \cite{hossin2015review}.
\begin{equation}
Precision = \frac{TP}{TP + FP}
\label{eq:pre}
\end{equation}
The Recall is the percentage of named entities in the corpus/golden annotations found by the system. It is calculated by computing the ratio of the number of correct system answers to the expected total number of answers as shown in Equation \ref{eq:rec}. A named entity is considered correct if it matches the corresponding entity in the corpus. 
\begin{equation}
Recall = \frac{TP}{TP + FN}
\label{eq:rec}
\end{equation}
The F1-score combines those two values, as shown in Equation \ref{eq:fscore}. It is the harmonic mean of Precision and Recall used to balance their results.

\begin{equation}
F1-score = \frac{2 * Precision * Recall}{Precision + Recall}
\label{eq:fscore}
\end{equation}

%% file: RelatedWorkNew.tex
\chapter{Related Work}\label{chap:relatedwork}
Many efforts have been made to improve the performance of several NLP tasks for the English language and develop similar Arabic language techniques. This chapter discusses some related work for the NER task on monolingual and code-switched data. Then, we discuss previous work of other related NLP tasks on CS data.   

\section{Named Entity Recognition}
The NER task was officially coined for the first time in the Message Understanding Conferences (MUC-6) in 1995 \cite{sundheim1995sixth}. Detecting and classifying entities in the text was one of the first steps needed for most information extraction applications. 
Information extraction refers to the task of converting unstructured information in text into structured data. One of the earliest Arabic NER research was done in 1998 \cite{maloney1998tagarab}, and more work was done for the Arabic language starting 2007 \cite{shaalan2007person}.
First, in this section, an overview of some of the previous work related to NER on monolingual English and Arabic data is presented. Second, we give an overview of NER approaches applied to code-switched data.  
\subsection{Named Entity Recognition on Monolingual Data}

Much research has been conducted on NER for monolingual text. Due to the morphological complexity of the Arabic language, fewer attempts have been made to tackle the problem of NER in Arabic as compared to other languages such as English. The two main approaches mainly used in NER systems are the rule-based and machine learning-based approaches.

The rule-based approach is one of the earliest approaches used and depends on grammar/hand-crafted rules, usually represented as regular expressions. The advantage of such an approach is that it relies on lexical resources and does not require annotated training data. For instance, this approach has been used for English NER in \cite{mikheev1999named,hanisch2005prominer,quimbaya2016named}.

For the Arabic NER task, we present some of the systems that used rule-based approaches. In \cite{maloney1998tagarab}, they developed TAGARAB, an Arabic name recognizer that uses a pattern‐recognition engine integrated with morphological analysis. The role of the morphological analyzer is to decide where a name ends, and the non-name context begins. They randomly selected fourteen documents and tagged them manually. The system achieved an F1-score of 85\% for recognizing four different entity types. 

In addition, in \cite{mesfar2007named} used NooJ\footnote{NooJ is available at http://www.nooj4nlp.net.} linguistic environment to process the Arabic text and build NER tagger. Their system was composed of a tokenizer, morphological, and named entity finder. They used a set of gazetteers and lists to find the named entities and support the constructed rules. They achieved F1-score equal to 85\% for Person, 76\% for Location, and 84\% for Organization.

In \cite{shaalan2009nera}, the Arabic NER system relied on a whitelist containing names and a set of grammar rules. It was evaluated using their data-set, the results of the evaluation accomplished a high F1-score of 87.7\% Person, 85.9\% for Location, and 83.15 \% for Organization.

The system of \cite{al2012real} was implemented using GATE\footnote{GATE is available at http://gate.ac.uk/.} and an Arabic morphological analysis is provided. They used several gazetteers in their system, and it was evaluated using ANERcorp. The system achieved an F1-score of 76.27\% for Person, 70.87\% for Location, and 57.30\% for Organization.

However, the rule-based approach is domain-specific, and lexicon resources are not always available. Therefore, creating such resources and maintaining them is time- and effort- consuming especially, if the linguists required knowledge is not available \cite{yadav2019survey,shaalan2014survey}. Thus, most recent studies have started moved to use machine learning-based approaches.

The second main approach is machine learning; there are three machine learning methods, supervised, semi-supervised, and unsupervised. The unsupervised and semi-supervised (bootstrapped) require very little training data for learning to predict using both labeled and unlabeled data. 
The main idea of supervised approaches is learning to predict by training on annotated examples with features. We will focus on the supervised learning approach similar to our approaches for NER taggers. 

Among the different supervised approaches that have been conducted on English NER are Hidden Markov Models (e.g., \cite{zhou2002named}), Conditional Random Fields (CRF) (e.g., \cite{mccallum2003early}), Support Vector Machines (SVM) (e.g., \cite{li2004svm}), and decision trees (e.g., \cite{schapire2013explaining}) \cite{nadeau2007survey}.

In \cite{benajiba2007anersys}, they used the Maximum Entropy (ME) and N-grams based algorithms, built a system (ANERsys), and created the freely available corpus (ANERcorp) and gazetteer (ANERgazet) for training and testing. This system achieved an F1-score of 55.23\%. The system was enhanced in \cite{benajiba2008arabic} by comparing two different techniques, SVM and CRF. They have also explored various combinations of contextual, lexical, and morphological features on different data-sets, but they could not show which one, SVM or CRF was better as it differs for different entity types. The best results achieved were 83.5\% F1-score for the ACE 2003 BN data \cite{doddington2004automatic}. In \cite{benajiba2008arabic2} the probabilistic model was replaced from ME to CRF, and the result as 79.21\% F1-score was obtained. 

A set of features was investigated in \cite{abdul2010simplified} to be used for CRF sequence labeling. These showed that character N-grams of leading and trailing characters of a word could be represented as lexical features. This could help in the NER task without the use of linguistic analysis. Their system achieved 81\% F1-score on ANERCorp dataset \cite{benajiba2007anersys} and 76\% F1-score on ACE 2005 dataset. However, they have only considered three NE types (persons, locations, and organizations). 

In \cite{abdelrahman2010integrated}, they implemented an Arabic NER task using bootstrapping semi-supervised pattern recognition and CRF. Their system extracts 10 types of NEs. It outperformed the LingPipe recognizer\footnote{http://alias-i.com/lingpipe} when both systems were evaluated on the ANERcorp data-set. 

Also, a combination of both rule-based and machine learning-based approaches was used as a hybrid approach. In \cite{mikheev1999named} and \cite{bajwa2015hybrid} English NER systems using hybrid approaches were implemented.

A hybrid approach for Arabic NER was also used in \cite{oudah2012pipeline} they used a hybrid approach for Arabic NER. The rule-based part of their system is similar to the one presented in \cite{shaalan2009nera}, regarding the machine learning part started by features and classifiers selections. Their approach showed promising results on the ANERcorp data-set; however, it still had the problems of the rule-based approaches. The system performance was 90.1\% F1-score for Location, 94.4\% for Person, and 90.1\% for Organization.

Afterward, using word representations for NER tasks
\cite{seok2016named} was experimented with. Since labeled data is costly and many unlabeled data exists, a new technique was introduced to cluster words and then use the clusters as features in supervised approaches. The first type used was Brown clustering \cite{brown1992class}, which improved the performance when used in \cite{liang2005semi} as features in semi-supervised English NER. The idea of Brown clustering is minimizing the bi-gram language model perplexity for a text corpus. However, its disadvantage lies in its inability to cluster tens of millions of phrases. K-means clustering algorithm was used in \cite{lin2009phrase} to include the clusters of phrases as features in the CRF classifier for NER. The advantage of using the k-means clustering algorithm is the ability to cluster tens of millions of words. Afterward, the focus moved to using word embeddings in linear NER models and enhancing the results (e.g., \cite{turian2009preliminary,turian2010word}).

In \cite{guo2014revisiting}, three approaches for incorporating word embeddings with CRF were presented to apply the NER task, binarization, clustering, and a new proposed distributional prototype method. A high performance using the three techniques compared to using the continuous embedding as heterogeneous features were achieved. It was demonstrated in \cite{passos2014lexicon} that plugging phrase embeddings in a log-linear CRF system improved the performance.

In the NER biomedical domain \cite{wu2015study}, word embedding is also used for NER in clinical texts. The clinical texts contain a lot of noise and unstructured sentences compared to general English texts. Different word embedding algorithms were investigated and compared since they could represent hidden meanings and capture relations in the real value matrix. Word2Vec and the ranking-based neural word embedding algorithms and three different strategies were compared for deriving and distributing word representation features from their embeddings. Also, the NER system presented in \cite{liu2015effects} achieved high results using a CRF classifier with features like lexicon resources, with one of them being word embeddings.

For the Arabic language, the first NER system that used word representations was presented in \cite{zirikly2014named}. Their system was applied on Dialectal Arabic, and the Brown clustering feature was used in addition to the classical features.  
They continued work, and they proposed in \cite{zirikly2015named} a technique to use word embeddings by clustering them and using the clusters as features along with the other set of features in the CRF system.  
Their approach showed promising results; nevertheless, it was designed to recognize entities extracted from social media texts written in dialectal Arabic. It achieved a 72.68\% F1-score on a Dialectal data-set. Besides, few details were given concerning the dimensions of the generated vectors and the numbers of clusters. The second one presented in \cite{laachfoubi2016arabic} mainly focused on comparing two different word embedding algorithms, Word2Vec and Global Vectors. The AQMAR corpus, which consists of 74k tokens, was used. The best performance of the system achieved was 67.22\% F1-score. However, the effect of the number of clusters was not studied.

Supervised or semi-supervised traditional machine learning approaches require domain-specific resources and a lot of feature engineering. That is why neural network systems have been proposed for the NER task, and this improved performance significantly \cite{yadav2018survey}. One of the early neural network architectures for NER was presented in \cite{collobert2008unified}. They constructed feature vectors from lexicons, dictionaries, and orthographic features. Later in \cite{collobert2011natural}, they implemented the first-word embedding model instead of the manually created feature vectors along with the neural network system. This model showed the importance of having word embeddings in several NLP tasks, such as the NER. The input of the model was given as a sequence of embeddings of each word in a sentence. They used a Convolution layer connected to a CRF layer and achieved an 89.59\% F1-score on the English CoNLL 2003 data-set \cite{tjong2003introduction}. 

In \cite{huang2015bidirectional}, they used BiLSTM instead of Convolutional Neural Networks. They proposed several models for sequence tagging. The model that produced accurate tagging performance 
was BiLSTM-CRF; a CRF layer was connected on top of BiLSTM to decode labels for the entire sentence. They scored 84.26\% F1-score on English CoNLL 2003 data-set using random embedding.
In \cite{yan2016multilingual}, they introduced different Neural Network-based models for NER on three data-sets in English, German, and Arabic. For the English, they used the CoNLL 2003 data-set, for German, they used GermEval 2014 NER shared task \cite{benikova2014germeval}, and for Arabic, they used ANERcorp. The different models they experimented with were BiLSTM, window BiLSTM, and a word-level feed-forward. They also added other features such as CRF, Part-of-Speech tagging, and word embedding. The best results for English were 88.9\%; for German 76.1\%, and for Arabic, 71.3\% F1-score.

Learning character embeddings has proved to help tackle out-of-vocab words, especially for morphologically rich languages. Therefore, combining the characters of a word and its context has been shown to enhance NER systems \cite{lample2016neural}. For instance, a hybrid BiLSTM and CNN architecture was created in \cite{chiu2016named}. Its advantages are to detect word and character features with no need to use most feature engineering. For modeling character-level information, CNN was used. A model using BiLSTM-CRF was introduced in \cite{ma2016end}. A CNN was added to the model to benefit from its capability to convert character-level data of a word into its character-level presentation. They combined word and character level representation to be the input to BiLSTM followed by CRF layer to generate the labels. They obtained a 91.21\% F1-Score on CoNLL 2003 data-set. One of the advantages of their model is easily applying data from several domains to the model because it does not require data from a specific domain or task-specific knowledge.

In \cite{lample2016neural}, developing resources and features for new languages and domains were tackled. Two models using BiLSTM were created, and output label dependencies via CRF were added and another one using a transition-based approach inspired by a shift-reduce parser for NER on four languages. The best performance using the BiLSTM and CRF models was achieved. Their model does not require any language-specific resources or features. A character-based word representation model was also used to capture orthographic sensitivity. For the English and German languages, CoNLL 2003 data-set was used and achieved a 90.94\% and 78.76\% F1-score. Moreover, for the Dutch and Spanish languages, CoNLL 2002 data-set \cite{tjong2003introduction} was used and achieved an 81.84\% and 85.75\% F1-score.

Another bidirectional recursive Neural Network connected to a Convolutional Network was explored in \cite{li2017leveraging}. This approach divides each sentence into chunks of meaningful sentences holed by nodes, then the model categorizes each node by these hidden features and evaluates hidden state features of every node. They got an F1-score equal to 90.91\% for BiLSTM-CNN and added to it word embedding only. Also, they achieved an F1-score equal to 91.55\% with lexicon and capitalization features on OntoNotes 5.0 data-set \cite{hovy2006ontonotes}. 
They also scored an F1-score equal to 91.62\% using BiLSTM-CNN along with the lexicon and word embedding.

As traditional word embeddings do not consider the context of the words, recently, new types of contextual word embeddings have been proposed and used in several NLP tasks. Some work has been conducted for the NER task using contextual embeddings for the English language. In \cite{peters2017semi} they introduced a semi-supervised approach using bidirectional language models for adding contextual pre-trained embeddings to different NLP tasks. They evaluated their model using CoNLL 2003 English data-set for NER. It achieved an F1-score of 91.93\%, which is an improvement of 1\% from the baseline system implemented with the regular pre-trained embeddings. 

Moreover, a new type of embeddings was proposed in \cite{peters2018deep}, which is deep contextualized word embeddings (ELMo) that could be added to existing models. This new type considers the syntax and semantics of the words and their variations in different linguistics contexts. They tested their new type of embeddings on different NLP tasks, including NER. The CoNLL 2003 data-set was used, and a baseline using character-based representation, pre-trained word embeddings, two BiLSTM layers, and a CRF layer were built. The baseline got an F1-score of 90.15\%, and the model was enhanced by adding ELMo to it. It got an F1-score of 92.22\%. 

The use of ELMo embeddings improved the performance of most NLP tasks. However, it is not easy to integrate it into neural network architectures, but there are several ways to do it: weighting the three layers or using only the first or last one. A task-specific architecture that considers the embeddings as additional features can also be used  \cite{reimers2019alternative}.
Another new type of embeddings was presented in \cite{devlin2018bert}; it is called Bidirectional Encoder Representations from Transformers (BERT). This new type uses an unlabelled text by conditioning the left and right context in all layers to train bidirectional word representations. Several approaches using BERT on CoNLL 2003 data-set for NER were compared, and the best one got an F1-score of 92.8\%. 

In \cite{akbik2018contextual}, the newly proposed type of embeddings called Contextual string embeddings for the NER task on CoNLL 2003 English and German data was used. The embeddings combine the advantages of all other contextual embeddings, which are training on a large unlabelled data-set,  and consider the context. They model the words as a sequence of characters to better handle misspelled words and prefixes and suffixes of words. An F1-score of 93.09\% and 88.33\% for English and German languages were achieved, respectively.

A Pooling contextualized embedding was proposed in \cite{akbik2019pooled} for the NER task. The open-source FLAIR framework \footnote{https://github.com/zalandoresearch/flair} to build the NER system was used and implemented using the BiLSTM-CRF sequence labeling. The system aggregates contextual embeddings of all unique strings by a pooling technique. CoNLL 2003 data-set and WNUT-17 task \cite{derczynski2017results} were used to evaluate their model. The system achieved high results of a 93.18\% F1-score on CoNLL 2003 English data, an 88.27\% F1-score on CoNLL 2003 German, a 90.44\% on CoNLL 2003 Dutch, and a 49.59\% F1-score on WNUT-17. The results achieved on the CoNLL 2003 data are considered state-of-the-art results in the NER task for these languages.

All the previous work presented mainly tackled the English language. Regarding the Arabic, in \cite{gridach2016character} they used deep learning for the NER task on the Twitter data-set. This takes advantage of both character- and word-level representations by applying them to integrate between BiLSTM and CRF. Not only did they use unannotated corpora, but their model also depends on unsupervised word representations learned from their corpora. Their system obtained an 85.71\% F1-Score for Arabic NER in social media. The data-set used in this paper is unsupervised, and it was collected from Twitter. In addition, this paper did not add the CNN layer, which was observed in our work to boost performance. 

As contextual embeddings achieved state-of-the-art results in several NLP tasks such as English NER, pre-trained Arabic contextual embeddings were proposed. One of the most recent embeddings, AraBERT \cite{antoun2020arabert} was trained on 70 million sentences, corresponding to around 24GB of the text of the news domain. AraBERT was used to evaluate different tasks, and its performance was compared in several baselines, one of which was the NER system implemented by \cite{el2019arabic}. The NER model was based on Bi-LSTM-CRF and BERT multilingual embeddings, the use of AraBERT increased the performance and achieved an F1-score of 84.2\%, which is the highest result on ANERCorp. 
BERT was trained in \cite{helwe2020semi} in a semi-supervised learning approach for Arabic NER using labeled and semi-labeled data-sets. They relied on the pre-trained model of AraBERT and followed the approach of teacher-student learning mechanism proposed by \cite{yalniz2019billion}. They compared their approach to three other available techniques presented in \cite{pasha2014madamira,abdelali2016farasa,helwe2019arabic} using the same MSA data-sets (AQMAR, NEWS, and TWEETS), and they achieved higher results on the first two sets as 65.5\% and 78.6\% F1-score respectively. 

\subsection{Named Entity Recognition on Code-Switched Data}
Concerning the NER on code-switching data, no work has been conducted in this direction for Arabic-English CS text before the present work. 
Regarding other language pairs, in \cite{gupta2016hybrid}, they introduced a hybrid approach for NER from CS English-Hindi and English-Tamil. A classifier based on CRF was used, and an F1-score of 62.17\% for English-Hindi and 44.12\% for English-Tamil were achieved.
In \cite{banerjee2017named}, they proposed a Bengali-English code-mixed data-set in the domain of sports and tourism for the NER task. They also compared four machine learning approaches for NER on code-mixed. The best performance achieved was a 92.31\% F1-score for sports data using CRF and a 70.63\% F1-score using SVM for the tourism data.   

In FIRE'2015, a shared task was established to collect and recognize entities from CS social media data for Hindi, Malayalam, Tamil, and English languages \cite{rao2015esm}. Lately, the CALCS 2018 shared task for the third workshop on Computational Approaches on Linguistic Code-switching \cite{AAS+18b} and was established for the NER task on CS data from social media for English-Spanish (ENG-SPA) and MSA-Egyptian (EGY). Of the 9 participants, 8 submitted work on ENG-SPA, and 6 submitted work on MSA-EGY \cite{ASM17,JKD+18,janke2018university,GCB18,WCK18}. The best performance of the language pair MSA-EGY was achieved by \cite{WCK18}; their model was implemented using BiLSTM-CRF and an embedding layer. An F1-score equal to 71.62\% was achieved.

Furthermore, a NER tool for Hindi-English code-mixed data was proposed in \cite{singh2018language}. They implemented two different models, one using a CRF classifier and another using an LSTM model composed of two bidirectional layers. The performance of their models was 72.06\% and 64.64\% F1-score, respectively.
A benchmark for Linguistic Code-switching Evaluation (LinCE) using deep learning and pre-trained ELMo and BERT-based models was proposed in \cite{AKS20}. The results obtained by the researchers can be submitted and compared to others in real-time. LinCE covers four CS language pairs, one of which being MSA-EGY and including several NLP tasks. One of them was the NER covering the language pair MSA-EGY.

In \cite{aguilar2019named}, they addressed NER for code-switched texts using 50k Spanish-English, and 10k Modern Standard Arabic-Egyptian annotated tweets on nine entity types. Other researchers showed different approaches for code-switched corpus acquisition techniques. Also, in \cite{gao2019code}, they presented a code-switching data generator system using a pre-trained BERT monolingual model and Generative Adversarial Networks. A Bert-based Chinese model was applied to monolingual data and configured to generate Mandarin-English code-switching data. A discriminator compared generated data and actual code-switched data, assuring that generated data was similar to real code-switching data. To test the effectiveness of the generated data, it was used to train a new language model. The trained language model showed lower perplexity of 3591.66 on monolingual corpus while it achieved 1935.97 perplexities on code-mixed data. Finally, GLUECoS \cite{khanuja2020gluecos} presented an evaluation benchmark containing data-sets in English-Hindi and English-Spanish for six NLP tasks and using pre-trained multilingual embedding.

\section{Related NLP Tasks on CS Data}

An overview of some of the most relevant work concerning collecting CS data by transcribing speech data or by using social media and web documents is provided in this section. Besides, the available work on CS Contextual embeddings, data augmentation techniques, and the automatic language identification task focusing on word-level identification are also discussed.
\subsection{Code-Switched Data Collection}
Due to the increasing importance of CS data, several researchers have been recently working on collecting CS data for different languages and NLP tasks (e.g., NER, Automatic Speech Recognition (ASR)). Several approaches have been used to collect such data, one of them being gathering (e.g., audio recordings from interviews) and transcribing speech data, as people tend to code-switch more while talking. One popular approach to gather data is using the power of the crowd using, for example, Amazon Mechanical Turk or Games With A Purpose (e.g., \cite{DBLPesp5} and \cite{osman2015building}). Another approach to collecting CS data was gathering data from social media platforms, web documents, or news commentaries. This approach was faster and easier than the first one and could lead to a vast amount of collected data. 
The following are some example of CS corpora:
\begin{itemize}
 \item Arabic-English corpus \cite{HEA18} collected by recording and transcribing informal interviews. 
 \item CSCS Egyptian-Arabic CS corpus \cite{BHA+20} containing 1,153 sentences, gathered from speech transcriptions \cite{HEA18}, that have been tokenized, lemmatized, and tagged with POS.
 \item MSA-DA (Egyptian) and English-Spanish corpora \cite{AAS+18b} proposed a CS NER data-set to benchmark NER approaches containing data gathered from Twitter and nine entity types.
  \item MSA-DA (Levantine, Gulf, and Egyptian) corpus \cite{ZC11} collected from news commentaries. It has sentences annotated on Amazon Mechanical Turk. Each sentence has three labels: dialectal content, how much dialect there is, and the type of Arabic dialect.
     \item Arabic-English parallel corpus \cite{MLJ+19} gathered from official documents of United Nations with two reference translations; monolingual Arabic monolingual English.
     \item MSA-English corpus \cite{Hamed2017} gathered by harvesting a collection of documents (books, tutorials, and notes) from the web in the domain of computers. The corpus contains 2,361,002 sentences, with  240,874 code-mixed sentences.
       \item Arabic-English corpus \cite{Hamed2017} collected from web documents of online libraries and search engines were presented. A language model for Arabic-English CS was built using the corpus.
    \item Romanized Algerian Arabic-French corpus \cite{cotterell2014algerian} collected from Algerian news website was created and annotated with word-level language type.
    \item MSA and Moroccan Arabic corpus \cite{samih2016detecting} created to use for language identification. Data about varying subjects and blogs from Moroccan Internet discussion boards were collected.
     \item MSA-DA (Egyptian) EMNLP 2014 Shared Task corpus \cite{SBM+14} created from unique tweet ids and character offsets of tweets and posts from blogs for the LID task. 
    \item MSA-DA (Egyptian and Levantine) corpus \cite{ED12} manually annotated for the language identification task, two ways of annotation were applied on each token, one based on the context and one ignored the context of the word.
    \item Turkish-German corpus \cite{Cetino1997} created with a focus on intra-sentential switches using Twitter. 
    \item Spanish-Wixarika corpus \cite{Mager2019} created from comments and posts from public pages on Facebook and manually annotated with intra-sentential switches as well. The last two corpora are similar to our data collection for the LID task, focusing on intra-sentential switches.
\end{itemize}

\subsection{Code-Switched Contextual Embeddings}

Some studies implemented different word embedding models to deal with various Arabic NLP tasks. For example, Arabic pre-trained word embedding models were implemented \cite{soliman2017aravec}. Their first version contained six distinct models built using either Continuous Bag-of-Words (CBOW) or Skip-gram (SG) techniques for the different text domains. They measured the similarity of word vectors of a subset of sentiment words and named entities to evaluate their models. Then they applied the clustering technique to see if, in each set, the words with the same polarity will be clustered together or not. SemEval-2017 Semantic Textual Similarity was also used to check how equivalent paired snippets of text were. 

Another set of word embedding models for the Arabic language was presented in \cite{fouad2020arwordvec}. The models were implemented using CBOW, SG, and GloVe techniques and generated from a set of Arabic tweets. They proposed a new way to measure the similarity of Arabic words to evaluate the performance of their proposed models. Besides, the performance in the multi-class sentiment analysis classification task was tested.

Bilingual models have been explored to bridge the gap between languages in CS word embedding and building models for low-resource languages. Several bilingual models have been proposed to align cross-lingual data using different alignment techniques such as word-level \cite{hermann2014multilingual,faruqui2014improving}, sentence-level \cite{hermann2014multilingual,gouws2015bilbowa}, both word and sentence level \cite{luong2015bilingual} and document-level alignments \cite{vulic2015bilingual,vulic2016bilingual}.

Besides, in \cite{upadhyay2016cross} they presented a comparison between four different cross-lingual embeddings models \cite{luong2015bilingual,hermann2014multilingual,faruqui2014improving,vulic2015bilingual}, varying in terms of the amount of supervision. \cite{pratapa2018word} compared between the three bilingual models \cite{hermann2014multilingual,faruqui2014improving,luong2015bilingual} for enhancing the downstream tasks of sentiment analysis and POS tagging on English-Spanish CS data. Moreover, they proposed an approach for training CS text generated by \cite{pratapa2018language} using Skip-grams. 

In \cite{lachraf2019arbengvec}, they proposed several Arabic-English cross-lingual word embedding models trained on pairs of Arabic-English parallel sentences. It is essential to mention that training cross-lingual and multilingual embeddings require monolingual data. The set of syntactic structures and semantic associations of code-switched text are not shown in monolingual sentences. Thus, learning CS text analysis using cross-lingual and multilingual embeddings is not optimal. Code-switched embedding should be trained using CS text \cite{pratapa2018word}. 

In \cite{gao2019code}, they proposed an approach to employ the BERT model and Generative Adversarial Net model for CS text generation. The developed system, capable of generating full CS sentences by training on low resourced CS data, the BERT model learns contextual embedding representation through this process. They evaluated their generated data in an ASR system on Mandarin-English CS data. A Multi-Encoder-Decoder Transformer model was proposed in \cite{zhou2020multi} for CS data. This model involves two language-specific encoder modules, each trained on monolingual data. These two modules are then integrated and trained on low-resourced CS data. Eventually, this module is capable of representing contextual embedding.  

Related to our work for Arabic-English CS data, in \cite{hamed2019code}, they compared different bilingual embeddings \cite{hermann2014multilingual,faruqui2014improving,luong2015bilingual} having different cross-lingual supervision. They also proposed two extensions, one of which depends on monolingual and small CS corpora, combining the first two approaches. They evaluated the effect of using different embeddings in language modeling. However, the proposed embeddings are classical ones and do not consider the context of the words.

\subsection{ Data Augmentation Techniques}

Due to the scarcity of data, several approaches were devised to overcome this issue. One way to produce more data is to use data augmentation techniques, i.e., automatically create new labeled training data from available ones. Another way is using active learning, i.e., selecting the most informative parts of an unlabeled data-set and manually labeling it \cite{tran2019bayesian}. Data augmentation aims to slightly modify/transform the already existing data in a relevant way to produce more data that is similar to the original one. 

It is a well-known technique for augmenting image data-set, where it was proven to be effective \cite{szegedy2015going,krizhevsky2012imagenet}. Data augmentation involves creating new images by cropping, padding, and horizontal flipping the original ones in the data-sets to train neural network models with larger sets \cite{chatfield2014return,krizhevsky2012imagenet}. Also, it is a successful technique in speech recognition \cite{cui2015data,ko2015audio}. The increase of data boosts the performance of the learning model with a limited number of training examples. However, applying data augmentation to the NLP field is harder and less studied, and this is due to the complexity of the different languages and the diversity of the various NLP tasks. 
Lexical substitution (substituting words in a sentence without changing its meaning) is one often used text data augmentation approach with many different techniques using, thesauri such as, for example, WordNet \cite{wei2019eda,shim2020data}), Classical word embeddings such as, for example, Word2Vec, FastText and GloVe \cite{wang2015s,hailu2019pre}), or Contextual embeddings (e.g., BERT \cite{kobayashi2018contextual}).
Some studies used back-translation to do data augmentation for different tasks while keeping the semantics of the sentences~\cite{luque2019atalaya,xie2019unsupervised} more, but none of them handled CS data. Several works have combined different techniques to augment the data \cite{lunmultiple,hernandez2018data}. 

Recently, large generative models have been used to artificially synthesize new labeled data based on fine-tuning a language model and having a filtering technique for selecting from the generated data \cite{anaby2020not,kumar2020data}. 
Many studies have been applied to augment data using unlabeled data and small labeled sets because of the lack of training data concerning the NER task. Very few works used data augmentation techniques to augment the training data for NER,  \cite{abacha2016meta,mathew2019biomedical} and none did on CS data. 

 In~\cite{abacha2016meta}, they proposed a selective data augmentation approach, which is based on selecting the most relevant data to augment a target training data from another. They also proved that selective data augmentation is better than combining several corpora. Also, in~\cite{mathew2019biomedical} and~\cite{kim2020construction}, they used bootstrapping approaches to generate machine labeled data and improving NER performance. 
 However, all the techniques mentioned above relied on having several labeled and unlabeled corpora, which are rarely available for CS data.
Data augmentation performed to generate realistic textual data for the NER task is a challenging task in CS, as it requires creating switch points and tagging them. Some existing data augmentation techniques, such as, for example, the random swap of words, are not beneficial for the NER task, as no new entities would be added.

\subsection{Automatic Language Identification on Code-Switched Data}

LID is the task of identifying the language of sentences or tokens. 
Previously, the automatic Language Identification (LID) task was known on the document-level, either monolingual or multilingual \cite{hughes2006reconsidering,baldwin2010language,lui2012langid,king2013labeling}. However, the focus moved to the word-level LID to process code-switched data.
Between the document-level and the word-level LID, there were several shared tasks on sentence-level LID in the Discriminating between Similar Languages \cite{malmasi2016discriminating} in particular, Arabic dialect identification.

LID is the most extensively covered task on CS data. This is due to the available LID-annotated corpora, and the LID shared tasks \cite{solorio2014overview,AD16,molina2019overview} which significantly contributed to the research on CS LID. Work on LID is also motivated because it is a pre-processing task for other tasks \cite{CSV16}. For Arabic CS, LID is a needed in several cases; MSA-DA, DA-Foreign when Arabic words are written in Arabizi, and DA-Foreign when the foreign language is written in Arabic script. Most of the work in this field covered the first case, followed by the second. The last case is the least frequently used and, as such, the least studied.

The first and second shared tasks held for language identification on code-switched data were in CodeSwitch shared tasks \cite{solorio2014overview,shirvani2016howard}. The best systems presented in \cite{shirvani2016howard} achieved high-performance results for all language pairs \cite{samih2016multilingual}.
However, most of the participants failed to recognize and assign a mixed label for intra-word CS. Different approaches are being implemented to tackle the CS LID problem in various languages. For instance, in \cite{bar2014tel} they focused on the language pair Spanish-English using SVM classifier. 

Besides, in \cite{barman2014dcu} they identified mixed words of Nepali-English data using various approaches such as linear kernel SVMs, dictionary-based methods, and k-nearest neighbor approach. An unsupervised word-level LID approach for CS data of any language pairs was implemented in \cite{rijhwani2017estimating}, without the need of having annotated training data. A Feed Forward network and a constrained decoder for LID of CS and monolingual data were presented in \cite{zhang2018fast}. In \cite{Jhamtani2014}, researchers implemented a model for LID on word-level Hindi-English data.


Additionally, an RNN system was implemented in \cite{chang2014recurrent} to detect the language of code-switched data such as English-Spanish, English-Nepali, Mandarin-English, and Modern Standard Arabic-Egyptian Arabic. They used Twitter data provided by the EMNLP Code-switching Workshop \cite{solorio2014overview}. Several other works have been conducted for CS identification for Egyptian Arabic and MSA data, such as in \cite{al2015aida2} that used CRF classifier and in \cite{samih2016multilingual} that used an RNN model.

In \cite{elfardy2013code,Chittaranjan2015} they tackled the problem of identifying the CS point in MSA-DA data.
Another work that has been conducted for CS identification for Egyptian Arabic and MSA data using the CRF classifier was presented in \cite{al2015aida2}. 
In \cite{eskander2014foreign} and \cite{al2014automatic}, they focused on distinguishing between English and Arabizi in the same sentence using a finite state transducer, morphological analyzer, and POS disambiguation tool, and a decision tree based on a language model. Moreover, a system for detecting the CS point was presented \cite{Al-Badrashiny2016} using CRF classifiers. They tested their system on several language pairs, including the ones similar to our Arabizi-English and Arabic-Engari. The best systems achieved an F1-score equal to 97.0\% and 98.9\%, respectively.

Segmentation of words is a significant step in the subword-level LID before tagging. A popular technique for labeling unsegmented data is the connectionist temporal classification (CTC) presented in \cite{graves2006connectionist}. Nevertheless, they assume a monotonic alignment between the inputs and the outputs and do not predict the segmentation boundaries. Later, the SegRNN model was proposed and used for segmentation, and labeling \cite{Kong2016}. Several machine learning methods segments the words into morphemes \cite{hammarstrom2011unsupervised,ruokolainen2013supervised,gronroos2014morfessor,cotterell2015labeled,kann2018fortification}.
 
Unfortunately, all of the above studies did not focus on detecting the language of code-switched intra-word. Identifying the language of the sub-word is a more challenging task and not widespread yet. Only two research papers tackled this issue, and not one did for AR-EN CS data. The first one was in \cite{Nguyen2016}, which focused on detecting intra-word CS for Dutch–Limburgish. They used the Morfessor \cite{creutz-lagus-2002-unsupervised} to segment all words into morphemes. For each morph, the model computed its probability in each language. The second one was in \cite{Mager2019}, which focused on German-Turkish (DE-TR) and Spanish-Wixarika (ES-WIX) CS data. Several models for segmentation and tagging of sub-words were implemented. The Segmental recurrent neural network (SegRNN) \cite{Kong2016} achieved the best F1-score of 98.7\% for DE-TR segmentation and 92.5\% for tagging DE-TR. Nevertheless, the model of BiLSTM and sequence-to-sequence got 98.1\% for segmentation of the language pair ES-WIX and 95.1\% for tagging of DE-TR. A similar architecture of the SegRNN was followed in this work to build LID models for AR-EN CS data.

%% file: NERMSA.tex
\chapter{Named Entity Recognition on MSA Data} \label{chap:NerMSA}
\begin{figure}[ht!]
    \centering
    \includegraphics[width=11.5cm,keepaspectratio]{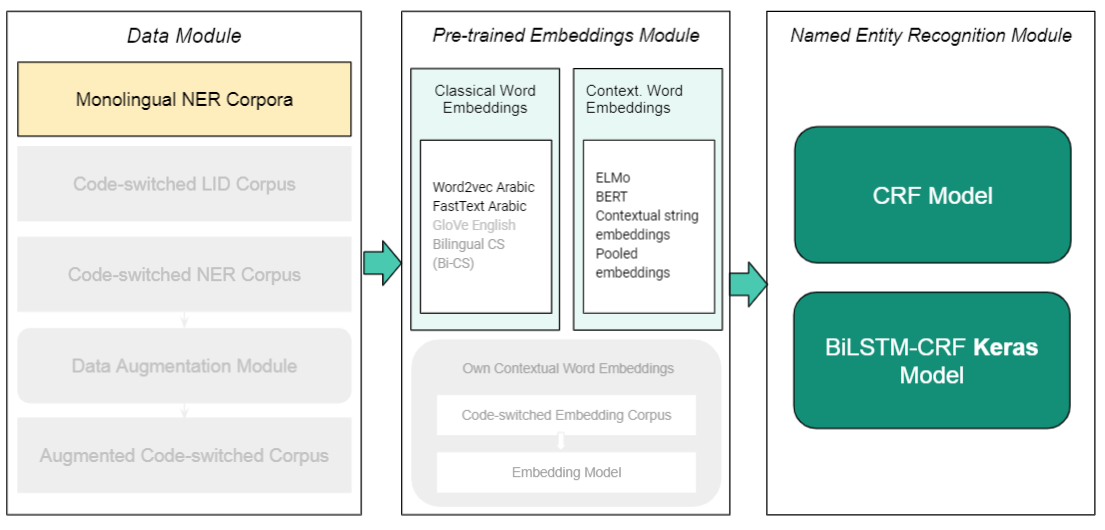}
    \caption{NER proposed approaches on MSA}
    \label{fig:nermsa}
\end{figure}

\chaptermark{NER on MSA Data}
 In the first part of this work we investigated applying NER task on MSA data before moving to the CS data \cite{sabtyarabic,awad2018arabic}. This chapter introduces the two different supervised NER approaches that we investigated for the MSA data as shown in Figure \ref{fig:nermsa}. It starts by the traditional machine learning approach CRF and then moving to the deep learning approach that we implemented.
The proposed NER approaches recognize named entities with three types of proper names, Location (LOC), Person (PER) and Organization (ORG), also known as the ENAMEX types. In addition, following the convention proposed in the CoNLL conferences \cite{Conll2002,tjong2003introduction}, a Miscellaneous type was used to include the proper names not belonging to the ENAMEX types and the ones not considered NE were tagged with other (O).

\section{Corpora}

 Two main available corpora were used in our introduced NER approaches for MSA; ANERCorp and AQMAR. The ANERCorp (Arabic Named Entity Recognition Corpus) was created by \cite{benajiba2007anersys}. It is the largest free annotated corpus for Arabic NER. The domain of the corpus is news and consists of 150,286 tokens and 32,114 entities which is a ratio of 4.67 tokens to types. The percentage of entities for each type is presented in Table \ref{tab:ANERCorp}. The labels of ANERCorp follows the IOB format (classes) that is used in MUC-6 tasks\footnote{http://cs.nyu.edu/cs/faculty/grishman/muc6.html}. Each class has two types: B-class and I-class. The B-class denotes the beginning of an entity and the I-class denotes the inside of a class.

 \begin{table}[ht]

\begin{center}
 \caption{Ratio of Entities per Type (\%)}
\begin{tabular}{l|r}
      \hline
     \bf Entity Type & \bf Ratio\\
      \hline
      Person & 39.0  \\
     
      Location & 30.4 \\
    
       Organization & 20.6 \\
      
       Miscellaneous  & 10.0\\
      \hline
   
\end{tabular}

\label{tab:ANERCorp}
 \end{center}
\end{table}
 The AQMAR (American and Qatari Modeling of Arabic) is an Arabic Wikipedia Named Entity Corpus and Tagger. It is composed of 74,000 tokens and around 5,854 entities of 28 Arabic Wikipedia articles hand-annotated for named entities \cite{19}. Unlike the first corpus, the domain of AQMAR is diverse and includes topic areas of interest such as, for example, history, technology, science, and sports.  
 In the following NER approaches, different training sets were investigated but unified so the testing data will be composed of 40k tokens.
 


\section{NER using Conditional Random Field}\label{sec:sectionCRF}
  
The first NER technique introduced for MSA data was using the supervised approach of Conditional Random Field and word embeddings. CRF, being one of the main statistical machine learning techniques, labels a sequence of tokens instead of classifying each token alone. Also, fundamental improvements in the NLP field and in the NER task took place because of developments in the word embeddings. Unsupervised word embedding with CRF showed that it can be integrated to perform NER task for MSA. This combination could not be done directly as CRF systems only allow categorical features and not continuous features such as word embedding vectors. 
The proposed solution was to cluster the generated vectors and plug the generated cluster IDs in the feature vector of the CRF system along with other lexical and contextual features. However, it was hard to know the optimal number of clusters. Thus, we compared different numbers of clusters to know which one achieves the best performance.



\subsection{Model Architecture}

\begin{figure}[!ht]
\begin{center}
\includegraphics[scale=0.7]{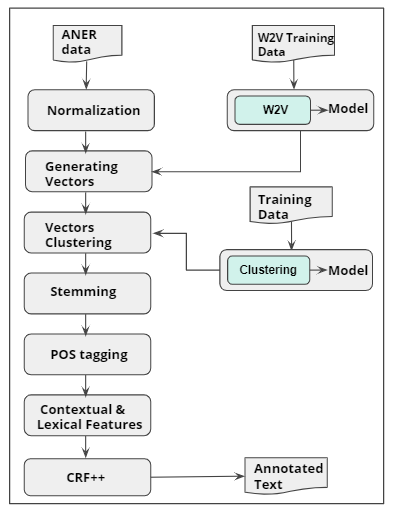}
\caption{A block diagram for the proposed CRF system}
\label{fig.1}
\end{center}
\end{figure}

As shown in Figure \ref{fig.1}, the system starts by normalizing the data. 
As the Arabic letters have different shapes, a normalization process was needed to unify some letters written differently. For instance, '\< آ >' and '\< أ >' are replaced with '\< ا >' and some punctuation such as '.' and ',' were removed. We trained our own Word2Vec model using an independent Arabic news-wire data-set. The normalized training data and the model were used to generate vectors for the training data. Afterwards, the vectors were clustered using a clustering model. The output from the previous step was used to generate IDs based on the cluster number for the training data. These IDs were added as features in the CRF system. Several features were added to the training data as will be explained later. In the final step, the new generated training data with all the features was fed as inputs to the CRF algorithm. The toolkit used to apply the CRF algorithm was CRF++ \cite{parker2009arabic}, open source tool used mainly for sequence classification. The main advantage of this tool was its ability to handle large feature sets while having the option of using multi-threading which makes it much faster than other existing CRF tools.

\subsection{Initial Baseline Features}
Features are considered the characteristic attributes of words that should be used with ML algorithms \cite{nadeau2007survey}. The main selected features are: stemming, POS tagging, and some contextual and lexical features. 
\subsubsection{\textbf{Stemming Features}}
In order to get the single representation of the words which is called Stem, a new approach should be implemented or used  \cite{taghva2005arabic}. Adding the stems of the words as a feature to the CRF algorithm can help matching similar words with different morphological representations. Thus, the "ISRI stemmer" algorithm that is described in \cite{taghva2005arabic} for word stemming on the whole data-set was used.  
\subsubsection{\textbf{Part-Of-Speech tag Features}}
The Part-Of-Speech is a category each word is assigned to represent its type such as, for example, Nouns, Verbs, etc.
Several POS tagging tools were investigated and the one that resulted in the best tagging performance was RDRPOSTagger. A language independent tool,
it identifies part-of-speech tags. RDRPOSTagger applied an error-driven approach to construct a Single Classification Ripple Down Rules tree of transformation rules \cite{nguyen2014rdrpostagger}. 
\subsubsection{\textbf{Contextual and Lexical Features}}
Contextual features are automatically generated based on the context of the NE. The context can be any number of previous or following words. Several sets of features for this class of Contextual features to get the optimal setting were investigated. Concerning the lexical features, these were the character N-grams of the tokens, or, in other words, they could be considered as the fixed length prefix and suffix of a word. In this current system, the first and last letters of a word are added as two lexical features.                 
\subsection{Word Embeddings based Feature}
The Word2Vec algorithm was used to produce the word embedding vectors. 
W2V provides multiple degrees of similarity between different words by mapping to nearby vectors, which is very useful for Arabic language. A dataset is used to train the W2V model that consists of 84M words. The generated embeddings vectors are indirectly integrated in the current system by adding them as a feature to the CRF algorithm to recognize Arabic NE. In order to convert continuous vectors into categorical features, a clustering technique is used to cluster the vectors and add their clustering IDs as a features. K-means, an unsupervised learning algorithm, is the clustering algorithm used and is one of most popular clustering algorithms. It partitions the unlabeled dataset into \textit{k} pre-set distinct clusters. Each item from the data will belong to only one group \cite{bock2007clustering}. Two of the main factors that affect the performance of the system are the vector size and the number of clusters. The vector size chosen was 100. Several numbers of clusters were investigated as shown in Section \ref{sec:ev}.

\subsection{Evaluation and Results}
\label{sec:ev}
The set of evaluations started with the usage of the ANERCorp corpus and was divided into training and testing data of 110,286 and 40k tokens, respectively. The evaluation process was divided into four experiments. First, the best combination of contextual features was checked. Second, the performance of the system was evaluated by adding different features to build the baseline. The third experiment integrated the word embedding with the selected set of features and compared the different number of clusters for the word embedding. Finally, combining the coarse and fine grained clusters IDs was investigated. 

\subsubsection{Contextual Features}
The first experiment started by evaluating the baseline with only the current word as a feature. Then, the next and the previous words were added to the current word separately. Furthermore, left and right contexts were combined together. In the combination the left or right context consisted of either one or two words. The performance measures used in the evaluation were precision (P), recall (R) and F1-score.

\begin{table}[t!]

\begin{center}
\caption{The performance measures results in (\%) due to using different combinations of contextual features}
\begin{tabular}{l|r|r|r}
      \hline
     \bf Feature& \bf Precision & \bf Recall &\bf F1-score\\
      \hline
      Current & 97.0 & 41.7 & 58.3\\
      Current-1Next & 96.3 & 37.9 & 54.3\\
       Current-2Next & 95.5 & 35.3 & 51.5\\
       Current-1Previous & \textbf{98.0} & \textbf{43.0} & \textbf{59.7}\\
       Current-2Previous & 97.1 & 40.0 & 56.6 \\
      Current-Previous-Next & 96.0 & 40.0 & 56.4\\
      Current-2Previous-2Next & 96.9 & 37.5 & 54.0\\
      \hline
   
\end{tabular}

\label{tab:table1}
 \end{center}
\end{table}

In Table \ref{tab:table1}, the results of the different contextual features are listed. According to Table \ref{tab:table1}, the best setting achieved by using the current and the previous word was equal to 59.7\% F1-score. Whereas, the lowest results came from using only the current and the 2 following words as 54\% F1-score. 

\subsubsection{Baseline Features Set}

In Table \ref{tab:table2}, the performances of the CRF system using the different types of features are illustrated. The Table shows an increase in the value of the F1-score by appending the features together. For instance, the addition of the lexical feature to the word stem and the current word increased the F1-score from 64.1\% to 66\%. Moreover, by adding all the features to build the baseline, the F1-score increased to 68.4\%.

\begin{table}[t!]

\begin{center}
\caption{The performance measures results in (\%) due to using different set of features}
\begin{tabular}{l|r|r|r}
      \hline
     \bf Feature& \bf Precision & \bf Recall &\bf F1-score\\
      \hline
      Current-Stemming & \textbf{96.0} & 48.2 & 64.1\\
      Current-Stemming-Lexical & 94.2 & 50.9 & 66.0\\
       Current-Stemming-Lexical-Contextual  & 95.1 & 51.2 & 66.5\\
      Current-Stemming-Lexical-Contextual-POStag & 93.2 & \textbf{54.1} & \textbf{68.4}\\
      \hline
\end{tabular}

\label{tab:table2}
 \end{center}
\end{table}



\subsubsection{Cluster Granularity}
In order to add the cluster ID of the word embedding as a feature, an experiment was conducted to get the relevant cluster granularity. Several numbers of clusters were experimented on in order to get the cluster size with the best performance. The result of this experiment is shown in Table \ref{tab:table3}. The evaluation started by adding the IDs of a small size cluster 50 to the set of features of the baseline. The size of the cluster was increased and the performance increased as well till reaching the cluster with size 500 and after that the performance started to decrease. The figure indicates that the cluster with size 50 is the one with the worst results being 72.7\% and the cluster with size 500 achieved the best result of a 76.1\% F1-score. Thus, adding the IDs of the cluster 500 to the feature set enhanced the performance of the baseline from 68.4\% to 76.1\%.        
\begin{table}[t!]

\begin{center}
\caption{The performance measures results in (\%) due to using different number of clusters}
\begin{tabular}{l|r|r|r}
      \hline
     \bf Number of Clusters & \bf Precision & \bf Recall &\bf F1-score\\
      \hline
      50 & 90.0 & 61.1 & 72.7\\
     100 & 90.0 & 62.6 & 73.8\\
      200 & 90.7 & 64.2 & 75.1\\
       500 & 92.0 & 64.9 & \textbf{76.1}\\
       1,000 & 90.6 & 64.1 & 75.0\\
      4,000 & 90.6 & 63.7 &  74.8\\
      \hline
      50 \& 500 combined & \textbf{91.4} & \textbf{65.7} & \textbf{76.4}\\
      \hline
\end{tabular}
 \label{tab:table3}
 \end{center}
\end{table}

Furthermore, the coarse and fine grained clusters were investigated by adding the IDs generated by the number of clusters 50 and 500 to the feature set. The output demonstrated that the performance had slightly increased to a  76.4\% F1-score. We interpret the improvement after the combination as coarse grained clustering could help more in modeling rare words. On the other hand, fine-grained clustering can results in better performance for common words.

\section{NER using Deep Learning on MSA Data}\label{sec:Deep_Learning}



In this Section we introduced the second NER tagger implemented for MSA data using Recurrent Neural Network.
 Recently, the usage of Neural Network has shown better recognition results than the CRFs. Over the past few years, deep learning models proved to be effective in solving a wide range of NLP tasks including NER and successively reached state-of-the-art performance \cite{lample2016neural,survey2020}. Deep learning models learn automatically non-linear combinations of features compared to the CRF that learns linear combinations \cite{habibi2017deep}. Moreover, it does not require handcrafted features or specific resources and it can learn all features from the data without having them set in advance \cite{yadav2018survey}. Since it is a new and promising approach, there was only one tagger for the Arabic language in Neural Networks. We investigated several RNN variations to reach the final model with the best performance. In addition, we evaluated the model with different contextual embeddings.




\subsection{Model Architecture}

\begin{figure}[!ht]
\centering
\includegraphics[scale=0.7]{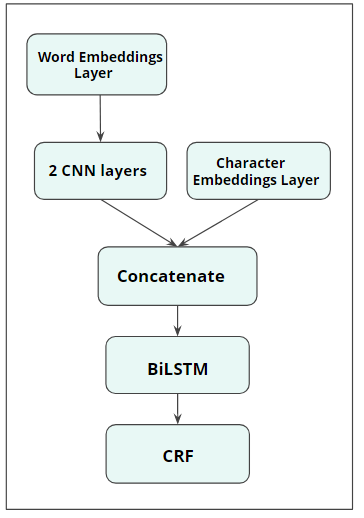}\\[2mm]
\caption{Proposed Arabic NER Deep Learning Model}
\label{Figure:model}
\end{figure}

Our model had several layers that we modified to understand their impact on the overall performance. The first two layers that come after the input layer are the Word Embedding and Character Embedding. After computing their outputs, these outputs are concatenated together to get the best result. Then two Convolutional Neural Network layers were added on top of the word embedding before concatenation of word embedding with character embedding. Now the main layers of our models follow; these are the BiLSTM then the CRF layers. They are mainly responsible for training the model on the input sequence and connecting the tags of a sequence together as shown in Figure \ref{Figure:model}.
We implemented the model using Keras framework\footnote{\url{https://keras.io}}, a neural network Python-based library that is used in many NLP tasks. This franework supports both recurrent and convolutional neural networks and can combine multiple types of neural networks.  

\subsubsection{Word Embedding}
In the first stage of BiLSTM-CRF model was using index encoding for word sequences but the performance of the model was not high as compared to most models that used Word2Vec. Thus, Word2Vec was fed to BiLSTM-CRF. The advantage to our model of using Word2Vec was capturing the characteristics of the
neighbors of a word and similarities between words.
The same Word2Vec Arabic model that we trained and used in the first approach of CRF, which contains 600k vocabulary and 100 dimensions has been used for getting the word embeddings.
\subsubsection{Convolutional Neural Network}
A multiple of convolutional layers with nonlinear activation function form a Convolutional Neural Network. 
In our model, we used 2 CNN layers after the pre-trained word embedding to improve the performance. Each with filter equal to 800 which is the best filter count after tuning the hyper-parameters. The kernel size had to be equal to 1 because any number greater than 1 will produce the problem of decreasing the number of the word embedding sequence. The use of pooling layers is a key aspect of Convolutional Neural Networks, typically defined after the convolutional layers. Pooling layers sub-sample their input \cite{23}. The pooling layer was not used for the model because it decreases the input sequence of the word embedding sequence while the tag sequence and character embedding sequence have not changed their sizes.

\subsubsection{Character Embedding}
Bidirectional LSTM was applied to create a character embedding model. First, a list of characters and their embedding were randomly initialized. Then the character embedding matching to each character in a word was given in straight and reverse order to the bidirectional LSTM. Then, the concatenation of both forward and backward forms of character embedding is formed to derive the word.


\subsection{Model Hyper-Parameters}
One of the most important properties that affect the learning of our model is the activation function which calculates the output from the summation of the weighted input signals of the neural network.
Another function that affects the training of deep learning models is the optimization function. The optimization function is used to minimize the output of error function \cite{32}. Thus, several optimization functions were tried to improve the output. Nadam is the optimization function that was used in our model and it outperformed other optimization functions. 

Deep neural networks have various non-linear hidden layers, which make the model an exceedingly expressive model to enable users to learn very complicated relationships between the outputs and the inputs. Some of the complicated relationships will be the outcome of sampling noise due to the limited training data. The dropout rate is used to tackle the problem of overfitting, which is due to the noise found in the training dataset but not in the test dataset \cite{26}. The dropout used in the model is equal to 0.2, it reduced over-fitting slightly and improved the F1-score.

\subsection{Evaluation and Results}
Both datasets of ANERCorp and AQMAR are added together forming a bigger dataset of more than 200K words. Since tags across the two corpora do not follow the same labeling guideline, they have been normalized. The different tags in the data are PER, LOC, ORG and MISC. For each type of tag there are two different forms, one for indicating the beginning of a name entity and the other for indicating the inside of a named entity. The whole dataset was divided into chunks of 150 words. The vocabulary was created using all distinct words. For tags, we used one hot encoder because every tag needs to have a different vector. Otherwise, the model might get confused when predicting the tag of a word. The dataset has been divided into training, validation, and testing of 72\%, 8\% and 10\%. 

The Word2Vec Arabic model with dimensions equal to 100 has been used for word embedding. Out-of-vocab words have been assigned an embedding vector of zeros. The main usage of character embedding is to give another representation for words. This should help our model to learn the word embedding out-of-vocab rather than simply ignore it. Hence, the words are represented by an array of characters and these characters are represented by one hot encoder. The model also takes a second input sequence of words in the form of character embedding. Finally, after granting a format for words, characters and tags, we used these forms to represent three different sequences to be trained and tested.


\subsubsection{Model Results}
In this part, we will compare and discuss the performance of the different models composed of various combinations of layers as shown in Table \ref{table:mr2}. The first model consisted of BiLSTM and CRF. The input of the model was the sequences of words with each word represented by an index. The result of the precision was equal to 86.07\% and recall equal to 34.5\%. Due to the very low recall, the F1-score and accuracy were the lowest, equal to 50.17\% and 91.93\%, respectively.

Then, we added the Word2Vec to the model, as it boosts deep learning models. The model showed remarkable improvement after adding the Word2Vec layer. We used the Word2Vec on the LSTM layer first; then, we showed the difference between LSTM and BiLSTM. Word2Vec on LSTM and CRF had an F1-score equaling 70.2\% and accuracy 96.26\%. The F1-score of the model of BiLSTM was equal to 72.65\%. The model accuracy had increased by 0.32\%. Thus, BiLSTM is better than regular LSTM. Afterwards, we added two CNN layers with 800 Filters on the latest model architecture. The F1-score of the model increased by 1.05\%. The model had a character embedding concatenated to its word embedding to solve the problem of out-of-vocab and enable the model to enhance prediction. The model achieved the highest results equaling a 75.68\% F1-score, 68.74\% on recall and 95.71\% on accuracy.

\begin{table}[!ht]

\begin{center}
\caption{Results of different models}
\resizebox{\textwidth}{!}{
\begin{tabular}{l | r | r | r | r | r} 
\hline
\bf Models & \bf Precision & \bf Recall &\bf F1-Score  & \bf Validation & \bf Testing \\ 
&  & & & \bf Accuracy & \bf Accuracy \\\hline

BiLSTM-CRF & 86.07 & 34.50 & 50.17 & 91.93 & 94.33\\ 
LSTM-CRF-WE & 76.20 & 65.10 & 70.20 & 95.26 & 97.28\\
BiLSTM-CRF-WE & \textbf{86.78} & 62.47 & 72.65 & 95.58 & \textbf{98.85}\\
BiLSTM-CRF-WE-CNN & 80.50 & 68.00 & 73.70 & 95.54 & 98.17\\
BiLSTM-CRF-WE-CNN-CE & 84.15 & \textbf{68.74} & \textbf{75.68} & \textbf{95.71} & 98.06\\ [1ex] 
\hline
\end{tabular}}
\label{table:mr2}
\end{center}

\end{table}
\subsection{Tuning Hyper-Parameters}
Tuning the hyper-parameters is a critical process to help choose the best values for each parameter that gets the best performance on a validation set.
The tuned parameters are the optimizer, activation function, epoch number and batch size. These parameters affect the performance of the model as explained below. The epoch of our model was tuned between 5, 10 and 50. The best results achieved were 75.86\% by the number of epochs equal to 10 as shown in Table \ref{table:ep2}.

\begin{table}[!ht]

\begin{center}
\caption{F1-score versus number of epochs}
\begin{tabular}{l | r} 
\hline
\bf Number of Epochs & \bf F1-Score \\ [0.5ex] 
\hline
5 & 72.15 \\ 
10 & \textbf{75.86} \\ 
15 & 73.84 \\ 
20 & 74.84 \\ 
25 & 74.21 \\ 
30 & 73.97 \\ 
35 & 74.97 \\ 
40 & 74.13 \\ 
45 & 74.17 \\ 
50 & 74.24 \\ 
\hline
\end{tabular}
\label{table:ep2}
\end{center}

\end{table}
The training dataset is composed of 1252 sequences, that are divided based on the value of the batch size. Table \ref{table:bs2} shows that the highest performance equal to 76.05\% is achieved by the batch size equal to 10 followed by 32 that got an F1-score equal to 75.15\%.

\begin{figure}[h!]
  \centering
  \subfloat[Dropout rate equal to 0.1]{\includegraphics[width=7cm]{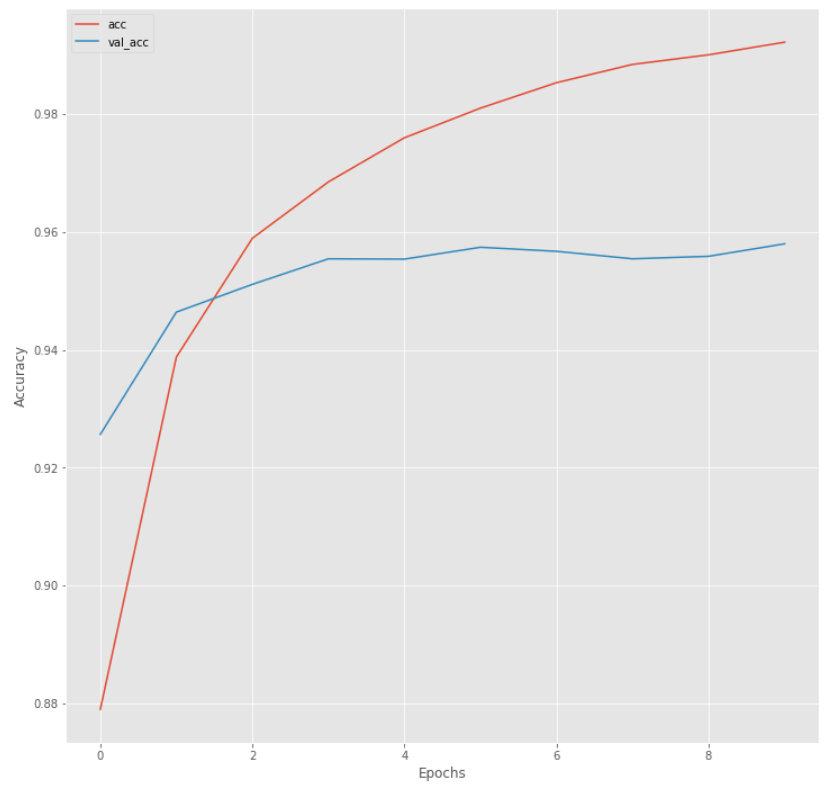}} 
  \hfill
  \subfloat[Dropout rate equal to 0.2]{\includegraphics[width=7cm]{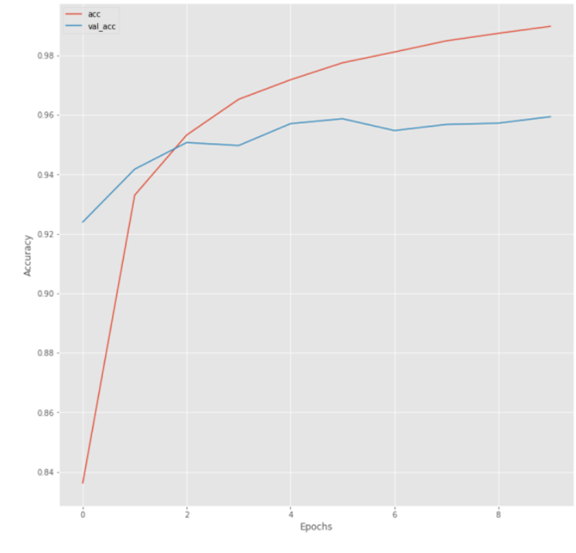}} 
  \caption{Dropout Rate Comparison}%
   \label{fig:dropOut}%
\end{figure}

\begin{table}[!ht]

\begin{center}
\caption{F1-score versus Batch sizes}
\begin{tabular}[center]{l | r} 
\hline \bf Batch Size & \bf F1-Score \\ [0.5ex] 
\hline
10 & \textbf{76.05} \\ 
20 & 73.64 \\ 
32 & 75.15 \\
40 & 74.81 \\ 
60 & 74.39 \\ 
80 & 74.62 \\ 
100 & 73.25 \\ 
\hline
\end{tabular}
\label{table:bs2}
\end{center}

\end{table}
The model has two different layers, CNN and BiLSTM, each one of them needing an activation function. We tried the following ones, Softmax, Softplus, Softsign, Relu, Tanh, Sigmoid, Hard-Sigmoid and Linear to reach the best performance. As shown in Table \ref{table:af} the CNN layer the Tanh function achieved the highest F1-score equal to 76.7\% and the BiLSTM layer the Linear function got the highest F1-score equal to 75.6\%.
\begin{table}[!ht]

\begin{center}
\caption{F1-score (\%) for different Activation Functions applied on CNN and BiLSTM}
\begin{tabular}[center]{l | r | r} 
\hline \bf Activation Functions & \bf CNN & \bf BiLSTM\\ [0.5ex] 
\hline

Tanh & \textbf{76.7} & 75.3 \\
ReLu & 72.8 & 71.8 \\
Linear & 69.8 & \textbf{75.6} \\
Softmax    & 43.4 & 19.6 \\
Sigmoid & 68.2 & 75.2 \\
Softplus & 72.2 & 75.3 \\
Softsign & 74.7 & 74.7 \\
Hard-Sigmoid & 74.6 & 66.4 \\
\hline
\end{tabular}
\label{table:af}
\end{center}

\end{table}
In addition to the previous parameters the optimization function has been tuned. The different optimization functions that were tried out were RMSprop, Adagrad, Adadelta, Adam, Adamax, and Nadam. As shown in Table \ref{table:of} Nadam was the best optimizer and it improved the F1-score of the model and achieved 75.43\%.
\begin{table}[!ht]

\begin{center}
\caption{F1-score for different Activation Functions applied on CNN and BiLSTM}
\begin{tabular}[center]{l | r} 
\hline
\bf Optimization Function & \bf F1-Score \\ [0.5ex] 
\hline

Adam & 72.28 \\ 
Nadam & \textbf{75.43} \\ 
Adamax & 73.84 \\ 

Adagrad    & 72.63 \\ 

Adadelta & 63.40 \\
RMSProp    & 71.41 \\

\hline
\end{tabular}
\label{table:of}
\end{center}

\end{table}
The final tuned parameter was the dropout. Its initial value was equal to 0.1 resulting in an F1-score equal to 75.11\% and after tuning the value of the drop out selected was 0.2, which increased the performance to 76.65\%.
The model used a dropout equal to 0.1 which has a high over-fitting between the accuracy and the validation accuracy. After tuning the dropout rate, we selected the value equal to 0.2 to be the best performing dropout rate. The F1-score of the 0.1 value was 75.11\% and of the 0.2 value equal to 76.65\%. Figure \ref{fig:dropOut} shows the difference between both dropout rates.


The final model that achieved the best performance after tuning the hyper-parameters was composed of BiLSTM-CRF, Word Embedding, two CNN layers and Character Embedding. Its activation functions are Tanh for CNN layers and Softplus for the BiLSTM layer. In addition, Nadam was used as the optimization function and the dropout rate was equal to 0.2. Its accuracy was equal to 95.94\%, and the F1-score 76.65\%.
\begin{table}[ht!]
\begin{center}
\caption{F1-score of the different models using various training datasets}
\begin{tabular}[center]{l|r|r|r} 
\hline
\bf Model & \bf AQMAR & \bf ANERCorp & \bf Both Datasets\\\hline
BiLSTM-CRF & 39.61 & 63.68 & 52.24\\
LSTM-CRF-WE & 57.67 & 80.39 & 74.80\\
BiLSTM-CRF-WE & 61.97 & 81.90 & 75.38\\
BiLSTM-CRF-WE-CNN & \textbf{69.31} & 81.30 & 75.76\\
BiLSTM-CRF-WE-CNN-CE & 67.22 & \textbf{82.18} & \textbf{76.65}\\
\hline
\end{tabular}
\label{table:all}
\end{center}
\end{table} 
After tuning the model, we trained the different types of models with the same tuned parameters but using AQMAR and ANERCorp datasets, each one alone, as shown in Table \ref{table:all}. We used the same testing data and made sure there was no overlap with the training data. The use of AQMAR alone reduced the performance as the F1-score of the model was equal to 67.22\%. However, the use of ANERCorp alone increased the performance by 5.58\% and got the highest results equal to 82.18\% F1-score. 


 \subsection{Contextual Embeddings}


In order to enhance the performance of the NER tagger on MSA data, we evaluated the usage of several contextual embeddings in the final architecture of the model instead of the classical W2V. We used the pre-trained models of ELMo, Pooled embeddings, AraBert and Contextual String embeddings. We also tried to combine two embeddings together of (FastText and Contextual string) and (AraBERT and Contextual string). We trained the model with the different pre-trained embeddings with the ANERCorp alone and with both datasets together. We did not train it using AQMAR alone as already the performance of the RNN model trained using AQMAR alone with classical embedding got lower results than the CRF. This could be due to having different domains for training rather than for testing, as the domain of the testing data was news and taken from the ANERCorp dataset. As shown in Table \ref{table:contall} all the results of training the model with ANERCorp dataset alone were higher than for both together. Also, the usage of ELMo model achieved the lowest results as the Arabic model used is small. Besides, AraBert embeddings achieved the highest results for ANERCorp equal to 83.71\% using ANERCorp, reflecting an increase of 1.53\% an absolute F1-score from the best results of the NER tagger with the classical embedding W2V. 
\begin{table}[!ht]
\begin{center}
\caption{F1-score (\%) after the usage of different embeddings in our NER system on the different training datasets}
\begin{tabular}[center]{l|r|r} 
\hline
\bf Embedding Model & \bf ANERCorp & \bf Both Datasets\\\hline
ELMo &57.57 &51.42 \\
AraBERT & \textbf{83.71}& 76.63\\
Pooled embeddings &75.03 &69.81 \\
Contextual String embeddings & 75.78&71.98\\
FastText \& Contextual String embeddings  & 79.86&73.76\\
AraBERT and Contextual String embeddings  &83.23& \bf 78.40\\
\hline
\end{tabular}
\label{table:contall}
\end{center}
\end{table} 

\section{Summary}
We presented two approaches for NER task on MSA data. First one an effective integration between CRF and word embedding was presented. Word embedding was integrated into the CRF classifier by clustering the vectors and adding the cluster-ID as a feature. It can be concluded that the system achieved the best performance by the following features: Current word, Stemming, Lexical, Contextual, POS tagging, fine- and coarse-grained word embedding cluster IDs. Then, as deep learning models proved useful in solving NER tasks in different languages, we investigated state-of-the-art DL models for NER on MSA data. We introduced various architectures such as LSTM-CRF, BiLSTM-CRF, Word Embedding classical or contextual, CNN, and Character Embedding to reach the highest performance model. 

\begin{figure}[ht!]
    \centering
    \includegraphics[width=13.5cm,keepaspectratio]{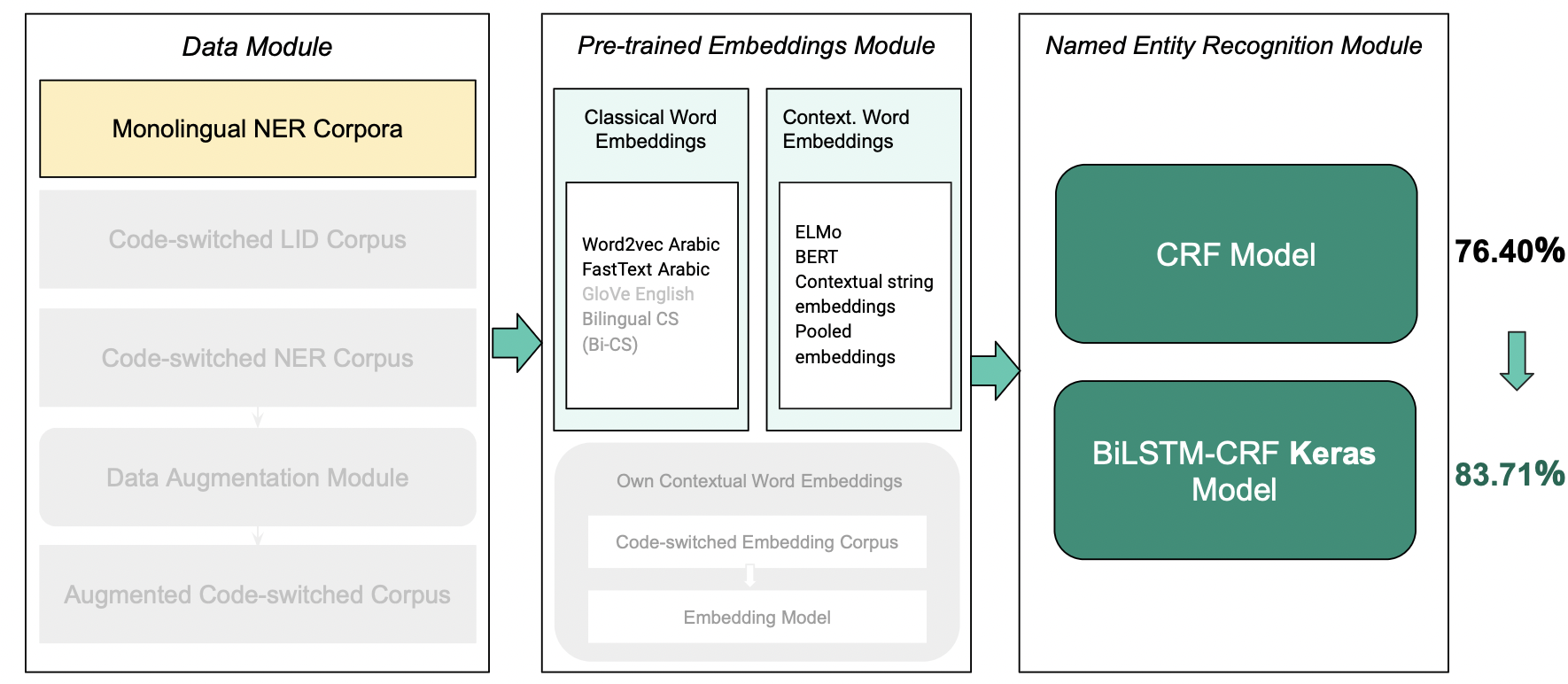}
    \caption{Results of Deep Learning MSA Taggers}
    \label{fig:sumResults}
\end{figure}

The final model with the highest F1-score consisted of BiLSTM-CRF with word Embedding, CNN, and Character Embedding after tuning the model hyper-parameters and adding contextual embddings instead of the classical one. Using the same training dataset, the NER tagger achieved an F1-score equal to 83.71\% which is an increase of 7.31\% from the CRF model as shown in Figure \ref{fig:sumResults}. Even while using different datasets in training, for the first approach using the ANERCorp and the second using both ANERCorp and AQMAR, the performance improved by 2\%. 

%% file: NERCS.tex
\chapter{Named Entity Recognition on Code-Switching Data}\label{chap:NERCS}
\chaptermark{NER on Code-Switching Data}

Social media reflects the changes we have in our daily lives. It shows the changed structure and type of the generated data. The language used in such posts is dialectal Arabic in addition to code-switching. Recently, code-switching became a widespread behavior in Arabic countries as Arab tend to use English words while speaking in Arabic. For such reasons we were motivated to analyze such data and explore applying the NER task to it.

\begin{figure}[ht!]
    \centering
    \includegraphics[width=13cm,keepaspectratio]{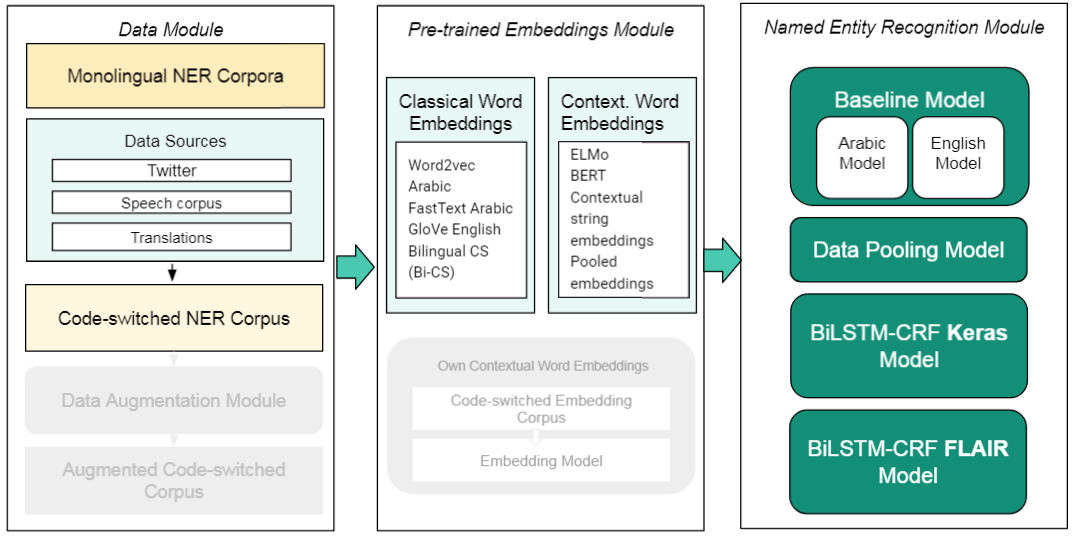}
    \caption{NER approaches on code-switched Arabic-English data}
    \label{fig:NERcs}
\end{figure}

This chapter introduces our second NER tagger for code-switching Arabic-English data. It first presents the first collected and annotated corpus for code-switched Arabic-English data for NER tasks. To the best of our knowledge no work has been conducted in this direction for the task of NER for Arabic-English CS text. This chapter also presents the first proposed RNN baseline NER system and the pooling technique introduced for this kind of CS data \cite{sabty2019named} as shown in Figure \ref{fig:NERcs}. It then discusses the proposed approaches to enhance the performance and have an NER system with higher results \cite{sabty2019techniques}. 

\section{Data Collection and Annotation}
One of the old preliminary approaches we investigated to create a NE list containing dialectal Arabic was using human computation techniques such as Game With A Purpose (GWAP). Human computation is the idea of utilizing human efforts to perform tasks that cannot be done or solved by computers satisfactorily or enjoyably for the individuals involved \cite{quinn2011human}. We created a prototype of a GWAP called ”3arosty” presented in \cite{sabtygamified}, to collect from users Arabic entities along with their categories and some related tags. The implemented prototype was tested by a diverse sample of players from different educational backgrounds. The number of players who played the game was 113 users; 43\% were males, and the rest females. The age of the players varied from 16 to 40 years old. The testing was held for a short time, and we were able to collect 220 words (entities), which means that each player played two games on average. The category with the most significant number of collected entities is Person, as 108 entities were collected. The location category followed them as 66 entities were collected, and the last one is the object category as 46 entities were collected. 

Later, we realized a more robust technique is needed to collect DA and CS sentences containing entities to create a large-scale corpus. We collected the corpus in two phases. In the first phase, we collected 1,331 sentences. In the second phase, we collected an additional 5,194 sentences. Thus the final corpus, composed of 6,525 sentences, contains 136,574 tokens. It has 22,705 (16.6\%) English words and 113,869 (83.4\%) Arabic words. The data was gathered from three different sources. The first one was Twitter collecting 2,303 Egyptian-English CS sentences. The second one took 1,150 sentences from the transcribed speech corpus for conversational Egyptian Arabic \cite{HEA18}. The last one translated 3,072 sentences from the Arabic ANERCorp and AQMAR data sets for Named Entity Recognition.

After collecting the data, the annotation of the entities started by identifying the boundaries of the named entity and then assigning the correct NE type. Our annotations followed the Named Entity annotation guidelines for the shared task of CoNLL-2003 \cite{tjong2003introduction} and was concerned with four types of entities, Persons, Locations, Organizations, and Miscellaneous that do not belong to any of the three types. Words that are not named entities were tagged with O. 
\begin{table}[!ht]
\begin{center}
\caption{Number of sentences containing entities in each data-set}
\begin{tabular}{l|r}\hline
\bf Data-set &  \bf Number of Sentences \\
\hline
Twitter & 1,477 (64.1\%) \\
Translated & 3,059 (99.6\%) \\ 
Transcribed Speech  & 412 (35.8\%) \\ 
Total & 4,948 (75.8\%) \\
\hline
\end{tabular}

\label{table:total} 
\end{center}
\end{table}
Table \ref{table:total} shows the number of sentences containing entities in each data-set and their percentages. The total number of sentences in the corpus containing entities was 4,948 sentences, about 75.8\% of the total number of sentences. 

The total number of NEs in the corpus was 17,577 tokens. Table \ref{tab:totalInClass} shows the total number of words under each NE class. The Person class contains the highest number of entities, 6,534 entities, and the Organization class contains the minimum number of entities, 3,100 entities.
\begin{table}[!ht]

\begin{center}
\caption{Number of entities in each entity type in the final data-set}
\begin{tabular}{l|r|r} \hline
\bf Entity Type & \bf Words & \bf \% of Total Words \\ \hline
Person & 6,534 & 4.8  \\
Location & 4,219 & 3.1 \\
Organization & 3,100 & 2.3\\
Miscellaneous & 3,724 & 2.7 \\
Total   & 17,577 & 12.9 \\ \hline
\end{tabular}
\label{tab:totalInClass}
\end{center}

\end{table}
\subsection{Twitter Data-set}
The first part of the corpus was composed of data harvested from Twitter by querying the Twitter API\footnote{https://developer.twitter.com/en/docs/api-reference-index}. We implemented different approaches to collect tweets containing CS data. One of the approaches was randomly selecting some tweets by gathering them using a query requiring the tweets to have a hashtag. A big percentage of hashtags was written in the English language. The tokenization of the hashtags was automatically done while collecting the tweets. Moreover, the query required that the tweets should contain the Arabic language. Other filtering criteria were followed by making sure that the tweets contained at least two English words to guarantee that the sentence contained CS text. However, words such as HTTP or via were not counted. Another approach used some of the named entities found in the ANERCorp data set as keywords in the search queries to collect more tweets containing NEs. Moreover, in order to guarantee having sentences containing entities, we created a list of famous person names and used them as seeds to collect tweets. 

The total number of sentences collected using Twitter was 2,303 sentences,  containing 38,281 tokens. As shown in Table \ref{tab:twitter2}, this part of the corpus had 5,810 entities, a total of 15.2\% of the total number of words and consisting of 7,405 (19.3\%) English words and 30,876 (80.7\%) Arabic words. The Miscellaneous class contained the highest number of entities, and the Location class contained the lowest number of entities.
\begin{table}[!ht]
\begin{center}
\caption{Number of entities in each entity type in the Twitter data-set} 
\begin{tabular}{l|r|r} \hline
\bf Entity Type & \bf Words & \bf \% of Total Words \\ \hline
Person& 1,907 & 5.0 \\ 
Location & 550 & 1.4\\ 
Organization & 1,098 & 2.9\\ 
Miscellaneous & 2,255 & 5.9 \\ 
Total  & 5,810 & 15.2 \\ \hline
\end{tabular}
\label{tab:twitter2}
\end{center}

\end{table}
\subsection{Translated Data-Set}
The second part was gathered by translating some of the existing annotated Arabic NER data. In the first collection phase, the selected sentences were randomly chosen from the Arabic ANERCorp data-set. In some sentences, we translated all the entities they contained. In other sentences, we translated one or two entities only in order to have Arabic entities in addition to the English entities in this part of the corpus. During the second phase of collection, more sentences were chosen from the same data-set in addition to sentences selected from the AQMAR data-set.  This technique was time-efficient as there was no need to annotate the data. However, sometimes it did not generate correct translations for Miscellaneous or Organization entities as they might have several meanings. Thus, a manual check was done on data to ensure that the words were translated correctly.

The number of translated sentences was 3,072 sentences, containing 78,215 tokens and were composed of 8,549 (10.9\%) English words and 69,666 (89.1\%) Arabic words. As shown in Table \ref{tab:translated}, the total number of named entities was 11,391, or 14.6\% of the total number of words. The Person class contained the highest number of words, equal to 4,516 words, and the Miscellaneous class contained the minimum number of words, equal to 1,348 words.

\begin{table}[!ht]

\begin{center}
\caption{Number of entities in each entity type in the translated data-set}
\begin{tabular}{l|r|r} \hline
\bf Entity Type & \bf Words & \bf \% of Total Words \\ \hline
Person& 4,516 & 5.8\\
Location & 3,601 & 4.6 \\
Organization & 1,926 & 2.5 \\
Miscellaneous & 1,348 & 1.7  \\
Total  & 11,391 & 14.6 \\ \hline
\end{tabular}

\label{tab:translated}
\end{center}
\end{table}
\subsection{Transcribed Speech Data-set}
The third part of our corpus, composed of data from the transcribed speech data-set \cite{hamed2018collection}, was gathered spontaneous speech gathered through informal interviews. The interviews topics were technical ones in order to have a higher probability to contain CS. They manually transcribed the corpus and formed
a total of 1,234 sentences and 17,769 tokens. In the original speech corpus, the sentences were divided into 124 monolingual Arabic,
125 monolingual English and 985 mixed. Overall, the data-set contained 79.8\% of code-mixing, 10.1\% of code-switching and 10\% of purely Arabic.
\begin{table}[!ht]

\begin{center}
\caption{Number of entities in each entity type in the transcribed speech data-set} 
\begin{tabular}{l|r|r} \hline
\bf Entity Type & \bf Tokens & \bf \% of Total Words \\ \hline
Person & 111 & 0.6 \\
Location & 68 & 0.3 \\ 
Organization & 76 & 0.4 \\ 
Miscellaneous & 121 & 0.6 \\
Total  & 376 & 1.9 \\\hline
\end{tabular}
 
\label{tab:speech}
\end{center}
\end{table}

As shown in Table \ref{tab:speech}, this part of the corpus contained 1,150 sentences including 20,078 tokens. Some pre-processing was performed on the data and more tokens/sentences were added. The corpus contained 13,327 (66.4\%) Arabic and 6,751 (33.6\%) English words. The Miscellaneous class contained the maximum number of entities, equal to 121 words, and the Location was the lowest, equal to 68 words only.

\section{Named Entity Recognition Model}
In this section, brief descriptions are provided for the different components of our NER model such as pre-trained word embedding and model architecture.

\subsection{Pre-trained Embeddings}
We investigated along with our deep learning model architecture for NER on CS data several types of embeddings from classical and contextual word embeddings. As stated before, the classical word embeddings do not take into consideration the context of the words. We used four different types of classical word embeddings, two for Arabic, one for English, and one for Arabic-English CS data.

The first one is our pre-trained word embeddings model for Arabic. We used the Word2Vec (W2V) algorithm to generate and save our Arabic word embedding model. 
The model of W2V was trained using an independent Arabic news-wire data-set. The second type of embeddings we used for the Arabic language is the pre-trained Arabic FastText embedding that generates vectors with dimensions equal to 300. Regarding the English word embedding model, we used GloVe for obtaining vector representations for words. The pre-trained model of GloVe we used was trained on Wikipedia 2014, and Gigaword 5 data \cite{pennington2014glove}.

As one of our focuses is on the Arabic-English CS data, 
 the pre-trained Bilingual CS Embeddings (Bi-CS), a bilingual Egyptian Arabic-English word embeddings of \cite{hamed2019code} were used. This model produced classical word embeddings by training on CS Arabic-English corpus. It was trained on monolingual data and a small amount of CS data, which acted as a gluing force bringing the monolingual embeddings closer in the vector space. The authors trained several word embeddings using multiple algorithms that rely on different levels of cross-lingual supervision. We used the Bi-CS embeddings as these showed the most promising performance. They have two different Bi-CS models that we tried; one trained using Skip-gram (Bi-CS-skip) and the other using CBOW (Bi-CS-cbow). 

Out-of-vocab words that were not found in W2V, GloVe, or Bi-CS models were represented by a vector of zeros in the embedding layer. However, classical word embeddings compute static vectors for each word, in polysemous words that depend on the context. Classical word embeddings fail to model these words. 
Contextual word embeddings are considered the new approach of representing the vector of a word in a given text compared to the Word2Vec and GloVe model. The following types of Contextual embeddings were used, ELMo, BERT, Contextual String embeddings, and Pooled FLAIR embeddings and their performances in the model were compared.

The first type of embeddings, ELMo was used with the Arabic ELMo representations. The second type was BERT and the main BERT model used during the training was the BERT Multilingual model, which contains 104 different languages, including the Arabic and English languages.
Contextual string embeddings was the third type of embeddings we investigated. 
The last one was the Pooled FLAIR embeddings.

\subsection{Model Architecture}

For the purpose of taking benefit of the previous and future contexts, we used the BiLSTM model. 
As stated before, this model represents each sequence forward and backward as two separate hidden states to save previous and future information. At the end, the two hidden states are concatenated to form the output \cite{schuster1997bidirectional}. 
 \begin{figure}[!ht]
\centerline{\includegraphics[width=10cm]{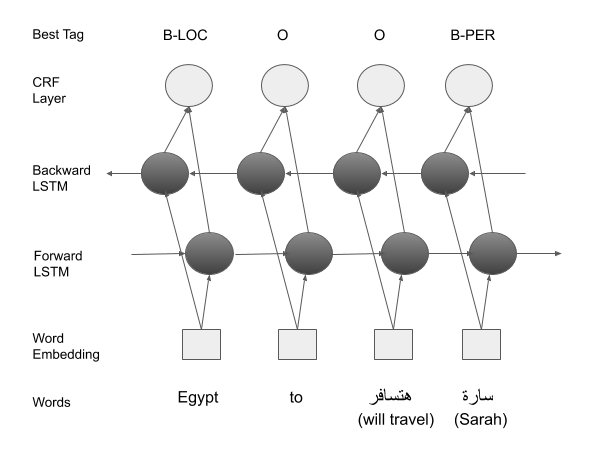}}
\caption{Our main BiLSTM-CRF model Architecture. The example illustrates the input sentence ``Sarah will travel to Egypt'' and its output predicted tags.}
\label{fig:bilstm}
\end{figure}
For the sake of predicting the current tags using CRF model. In order to combine the advantages of BiLSTM and CRF networks, we constructed all our models using BiLSTM-CRF. The model architecture was composed of three layers as shown in Figure \ref{fig:bilstm}. The first one was the input layer. It contained the word embeddings. We tried several types of word embeddings as will be explained later. The second one was the hidden layer of the BiLSTM. The last one was the output layer. It is where the CRF layer calculates the probability distribution over all labels of the previous and future tags to predict the current best tag. As it has been proven, using character-level embeddings is useful as they can handle the out-of-vocabulary words \cite{lample2016neural}. Thus, we constructed a second model architecture containing four layers. The first one was the word embeddings layer. The second one was the character embeddings layer. Then, a concatenation of the first and second layers was done to be given as input for the third layer, the BiLSTM layer. The final one was the CRF layer. 

\section{Experiments and Results}
\begin{figure}[!ht]
\centerline{\includegraphics[width=14cm]{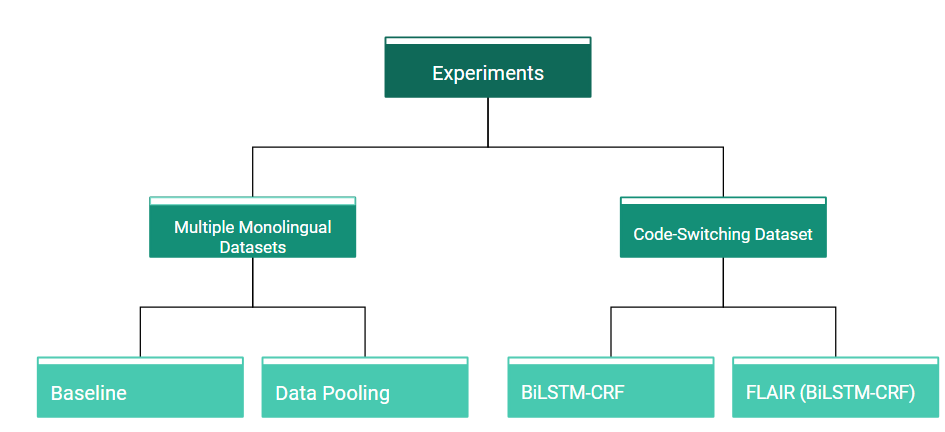}}
\caption{Different experiments applied on different data-sets}
\label{fig:experiments}
\end{figure}

We conducted two sets of experiments as shown in Figure \ref{fig:experiments}, the first one with only monolingual data and the second with the full data-set containing CS data. While the training data-set was different for each experiment, we unified the testing data-set to compare the performance of the different experiments. The testing data contained 1,219 sentences and 28,581 tokens.

\subsection{Multiple Monolingual Data-sets}\label{sec:multipleMono}
We implemented our baseline system and enhanced it using the pooling technique. We did not use in the baseline or pooling systems any CS data in the training process due to the unavailability of sufficient CS data at the time. The training data was combining multiple monolingual data, as will be explained in detail later. 
\subsubsection{Baseline}
In order to build the baseline, two models were created. The first one was the Arabic model. The Arabic data-sets we selected for training were the ANERCorp and AQMAR containing 225,000 words annotated for the NER task. 
The performance of the Arabic model was a 27.7\% F1-score. The second one was the English model. The data we selected for training was CoNLL 2003, containing 206,931 annotated words. The performance of the English model is 7.9\% F1-score. Both systems achieved a low F1-score, which is expected as the training data contained only one of the two languages and the testing data contained mixed sentences.

The baseline started by detecting the language of each word in the testing data-set. Based on the detected language, the predicted tag was taken from either the Arabic or English predictions. The overall F1-score achieved by the baseline system was 52\%, which is still a low expected performance as the predictions do not consider code-switching context. 
\subsubsection{Data Pooling}
We introduced the concept of data pooling to overcome the problem of the lack of large CS data. Pooling is done by combining annotated Arabic and English data-sets to form the training data-set. We used the same Arabic and English data-sets as the baseline, but we combined them; thus, the total number of words in the training file was 431,931 tokens.
The training was performed using Nadam optimization function, SoftPlus activation function, batch size of 32 and epochs number of 10.   

As the corpus contains two different languages, we suggested loading two classical word embedding models to cover English and Arabic words. For Arabic, we used our W2V model, and for English, we used the GloVe model. As a result of using the pooling technique by combining the two training data-sets and using the collected CS corpus for testing, the best performance achieved was a 60\% F1-score. This result is expected, considering that the training and testing data belong to a different context due to the limited amount of CS data in our previous work.
\subsection{Code-Switching Data-set}
As an extension to the experiments explained in Subsection \ref{sec:multipleMono}, we collected more data and were able to train and test with CS data. The total number of sentences in the training file was 5,306 sentences, containing 107,993 tokens. We explored the combination of BiLSTM-CRF architecture with different types of embeddings. The first model was created using using PyTorch, an open source library created by Facebook used in several NLP tasks and providing same deep neural network architectures and features as Keras. We also used the open-source FLAIR framework to create the second model and try the same architecture of BiLSTM-CRF with their proposed different embeddings to identify the best performance for NER on our CS data-set. FLAIR a recent NLP framework designed to train and distribute text classification, language models and sequence labelling, unifies the use of many different word embeddings as well as random combinations of embeddings \cite{akbik2019flair}. The FLAIR system was trained several times using our CS training data-set each time with a different type of embeddings, and we took advantage of its feature of combining two types of embeddings.

The setups of the different models we tried were as follows:\\ \textbf{BiLSTM-CRF}: We experimented using different types of embeddings with the model architecture of BiLSTM-CRF. We implemented this model using anaGo\footnote{https://github.com/Hironsan/anago} library, which is a Python library implemented in Keras for sequence labeling. The first type was the same one we used in the pooling technique, but it was used with the newly collected CS training data and was composed of the two pre-trained word embeddings GloVe for English and W2V for Arabic. The second one used the Bi-CS word embeddings with its two models of Bi-CS-skip and Bi-CS-cbow to see the effect of using CS word embeddings on the performance. The last two are using BERT Multilingual and ELMo Arabic embeddings.


\textbf{FLAIR (BiLSTM-CRF)}: We investigated the use of FLAIR and the model architecture of BiLSTM-CRF with the following different types of embeddings explained before, BERT, ELMo, FastText, Bi-CS-skip, Bi-CS-cbow, Contextual string embeddings, and Pooled embeddings. Besides, we tried combining the highest performing embedding type with other embeddings.

We tuned the hyper-parameters of the models using grid search. The dropout ranged from 0.2 to 0.8 and the batch size ranged from 10 to 32, the dense layer size ranged from 50 to 1024, BiLSTM size ranged from 100 to 1024, and the number of epochs ranged from 50 to 150. Concerning the optimization function, we used Adam optimizers and SGD optimizers with a learning rate of 0.001 to 0.1 and used the tanh function regarding the activation function. 


 



\begin{table}[!ht]

\begin{center}
\caption{Results of the NER models with different types of embeddings} 
 \resizebox{\textwidth}{!}{
\begin{tabular}{l|r|r|r} \hline
\bf Model & \bf Precision & \bf Recall & \bf F1-Score  \\ 
\hline
\bf BiLSTM-CRF  & & & \\
 \quad BERT & 63.05 & 46.81 & 53.73  \\
 \quad W2V \& GloVe & 70.53 & 62.09 & 66.04  \\
 \quad ELMo & 70.00 & 69.66  &  69.83  \\
 \quad FastText & 76.55 & 67.80 & 71.91 \\
 
 \quad Bi-CS-cbow & 78.05 & 67.57 & 72.43 \\
 \quad Bi-CS-skip & 78.08 & 67.71 & 72.53 \\
 \quad Pooled embeddings  & 69.85 &	63.10 & 66.30  \\
 \quad Contextual string embeddings  & 70.05 &	63.38 & 66.55  \\
\quad Contextual string embeddings \& AraBert  & 74.11 & 67.62 & 70.72  \\
\quad Contextual string embeddings \& ELMo  & 73.14 & 64.57 & 68.59  \\
 \quad Contextual string embeddings  \& Bi-CS-cbow   & 73.94 & 65.81 & 69.64 \\
 \quad  Contextual string embeddings \& Bi-CS-skip  & 71.59 & 69.24 & 70.39 \\
  \quad Contextual string embeddings \& FastText  & 74.50 &	72.76 & \textbf{73.62}  \\ \hline
\bf FLAIR (BiLSTM-CRF) & & & \\
 \quad Bi-CS-cbow  & 63.58 & 51.80 & 57.09 \\
 \quad Bi-CS-skip  & 66.48 & 52.52 & 58.68 \\
 \quad ELMo  & 79.34 & 61.47 & 69.27 \\ 
 \quad BERT  & 82.40 & 60.90 & 70.04 \\
 \quad FastText & 75.65 & 66.00 & 70.49 \\
 \quad Contextual string embeddings  & 74.58 & 70.42 & 72.44  \\
 \quad Pooled embeddings  & 76.83 & 72.66 & 74.69 \\ 
 \quad Pooled embeddings \& BERT  & 81.29 & 63.76 & 71.47 \\
 \quad Pooled embeddings \& ELMo  & 77.00 & 69.38 & 72.99 \\ 
 \quad Pooled embeddings \& Bi-CS-skip & 77.93 & 73.85 & 75.84 \\
 \quad Pooled embeddings \& FastText  & 79.15 & 76.28 & \textbf{77.69} \\
\hline
\end{tabular}}
\label{tab:fscoreNER}
\end{center}

\end{table}

As shown in Table \ref{tab:fscoreNER}, the results of BiLSTM-CRF with BERT embeddings were equal to a 53.73\% F1-score, considered to have the lowest performance. This could be due to the type of data BERT multilingual model used in training to generate the pre-trained embeddings, Wikipedia pages. The language used in Wikipedia was MSA, different from the one used in the CS training data. Moreover, the embeddings did not contain CS sentences.

The same model of BiLSTM-CRF and the two pre-trained word embeddings for English and Arabic that we previously trained with the pooling technique got a 6.4\% higher F1-score using CS training data. However, this low F1-score, as compared to other models, could be due to having two different embedding models for English and Arabic that give the same word in both languages different embeddings. Adding ELMo or FastText embeddings in our model enhanced the performance and we obtained an F1-score of 69.83\% and 71.91\%, respectively. Furthermore, when we used Bi-CS-cbow embeddings, the model achieved an F1-score of 72.43\%. We tried combining the Contextual string embeddings as it is the highest character-level embedding with the other types of embeddings. The Contextual string embeddings and FastText together outperformed the results of the other embeddings used with our model and got an F1-score equal to 73.62\%.

The embeddings with the lowest results equal to 57.09\% and 58.68\% F1-score with FLAIR system were the Bi-CS-cbow and Bi-CS-skip, respectively. There was no significant difference in the performance of the FLAIR system with ELMo, BERT, or Arabic FastText embeddings. These achieved an F1-score equal to 69.27\%, 70.04\%, and 70.49\%, respectively. The performance improved more while using FLAIR with Contextual string embeddings and Pooled embeddings as they are the newest types of embeddings, and deal efficiently with unseen words. They achieved an F1-score of 72.44\% and 74.69\%, respectively.  

Also, to take advantage of the FLAIR feature of combining several embeddings, we evaluated combining the Pooling embeddings that achieved the highest F1-score with the other existing types. Adding BERT and ELMo to it did not achieve higher results than the Pooling alone, and F1-score of 71.47\% and 72.99\%, respectively was obtained. However, combining it with Bi-CS-skip enhanced the F1-score by 1.15\%. In addition, the Pool with FastText embeddings performed particularly well on our task of NER on CS data. It produced the highest results among all other models, equal to 77.69\%. 
We selected the two models with the highest F1-score to check the effect of adding character-level embeddings on them. Thus, we first tried the FLAIR and the BiLSTM-CRF models with character-level embeddings alone. Then, we added character-level embeddings to FLAIR with Pooled embeddings \& FastText and to BiLSTM-CRF with Contextual string embeddings \& FastText.


   

\begin{table}[!ht]
\begin{center}
\caption{Results of the highest models with character-level embeddings } 
     \resizebox{\textwidth}{!}{

\begin{tabular}{l|r|r|r} \hline
\bf Model &\bf Precision &\bf Recall & \bf F1-Score  \\ 
\hline
\bf FLAIR (BiLSTM-CRF) & & & \\
\quad Character Embeddings & 71.16 & 64.04 & 67.41  \\
\quad Character Embeddings \& Pooled embeddings \& FastText  & 79.28 & 75.61 & \textbf{77.40} \\
\hline
\bf BiLSTM-CRF  &  & &\\
\quad Character Embeddings & 77.53 & 72.33 &  74.84  \\
\quad Character Embeddings \& Contextual string embeddings \& FastText & 77.30 & 73.47 & 75.34  \\
\hline
\end{tabular}}
\label{tab:character}
\end{center}

\end{table}

It was observed that having character embeddings implemented by FLAIR alone got low results of a 67.41\% F1-score. Having character embeddings with our BiLSTM-CRF model got an F1-score of 74.84\%. Besides, adding character embeddings and the Contextual string embeddings \& FastText embeddings enhanced the F1-score by 1.72\% as shown in Table \ref{tab:character}. Nevertheless, this decreased the performance by 0.26\% while being added to the FLAIR model with the Pooled embeddings \& FastText. This means that character-level embeddings do not benefit the new type of contextual embeddings along with the classical word embeddings in our case. Thus, in the end, the model of FLAIR with Pooled embeddings \& FastText remained the one with the highest performance equal to 77.69\%, and outperformed the previous results of the baseline by a  25.69\% absolute F1-score. 

To further investigate the result of the model with the highest F1-score, the following Table \ref{tab:Highest} shows the results for each entity type. The entity type Person had the highest F1-score, equal to 89.88\%. This result was expected as in our corpus, the maximum number of entities belongs to the Person class, followed by the entity types Location and Organization, equal to an 84.52 \% and 61.66\% F1-score. The minimum F1-score, 37.07\% belongs to the Miscellaneous class. 

\begin{table}[!ht]

\begin{center}
\caption{Detailed results of each entity type for the highest model}
\begin{tabular}{l|r|r|r} \hline
\bf Entity Type &\bf  Precision &\bf  Recall &\bf  F1-Score  \\ \hline
Person & 86.65 & 93.36 & 89.88\\ 
Location & 82.02  & 87.17 & 84.52  \\ 
Organization  & 75.33 & 52.19 & 61.66 \\ 
Miscellaneous & 41.53 & 33.48 & 37.07 \\ \hline
\end{tabular}

\label{tab:Highest}  
\end{center}
\end{table}

\section{Summary}

\begin{figure}[ht!]
    \centering
 \includegraphics[width=13cm,keepaspectratio]{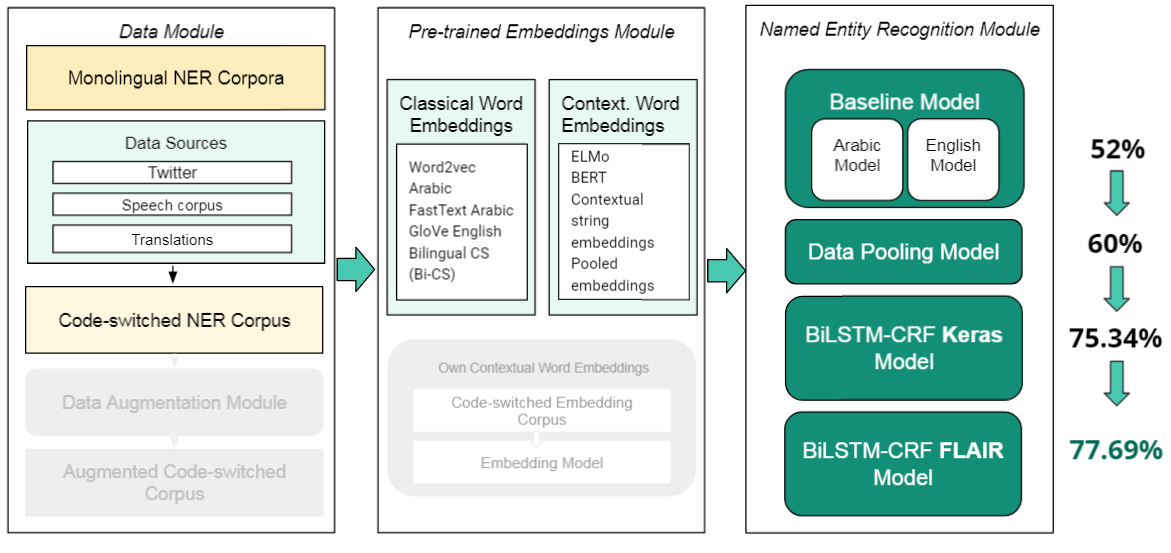}
    \caption{Results of deep learning CS taggers}
    \label{fig:sumRCS}
\end{figure}

We presented the first annotated Arabic-English CS corpus for NER. The corpus contains 6,525 sentences along with different deep learning models for NER on CS Arabic-English data. First, a baseline model was built by training two different models: one for Arabic and another for English, to detect the language of the testing words and get the predicted tag accordingly. As shown in Figure \ref{fig:sumRCS} the performance of the baseline was 52\% F1-score. Second, to improve the results, we introduced a data pooling approach by combining different English and Arabic training data-sets. As initially, the CS corpus size was minimal; it was only kept for testing and evaluation. Moreover, the model used two different pre-trained word embedding models: Arabic and another for English. Then we implemented deep learning models, that were composed of the BiLSTM-CRF network and classical or contextual pre-trained word embeddings models. Pooled embeddings \& FastText as pre-trained word embeddings in the model achieved the highest performance equal to 77.69\% F1-score. 




%% file: Pre-trained_Embeddings.tex
\chapter{Contextual Embeddings for Arabic-English CS Data}\label{chap:Pre-trained-Embeddings}
\chaptermark{Contextual Embeddings for AR-EN CS Data}

\begin{figure}[ht!]
    \centering
 \includegraphics[width=12cm,keepaspectratio]{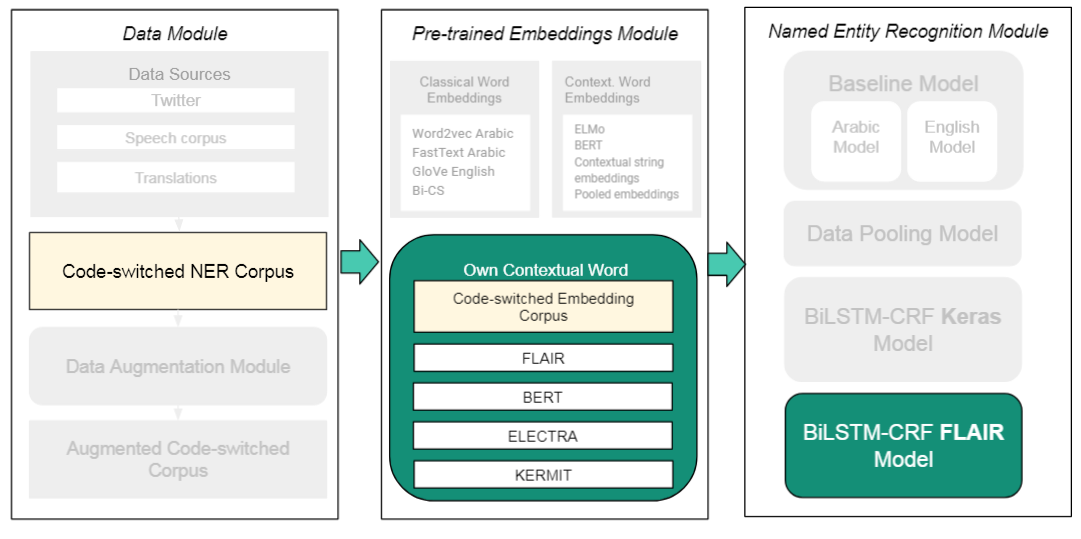}
    \caption{First Enhancement Technique}
    \label{fig:ContEmb}
\end{figure}

To enhance the performance of our NER tagger for CS data, we proposed a solution to train bilingual contextual embedding models using state-of-the-art embedding types generated from the CS Arabic-English corpus we collected. In this chapter, we present our CS created corpora and embedding models using Contextual String embeddings, BERT, and ELECTRA. We also propose a new contextual word embedding model called KERMIT, capable of mapping both Arabic and English words inside one vector space efficiently in terms of data usage \cite{sabty2020contextual}. All our trained and proposed models are available as an open source\footnote{https://github.com/CSabty/Code-Switch-Arabic-English-Contextual-Embeddings}.
We evaluate our embedding models in our NER model as shown in Figure \ref{fig:ContEmb} and in other NLP tasks to check which one will enhance their performance.
\section{Data Collection}
 As the embedding models and especially the Transformers, ones are composed of millions of parameters. Models with such large parameters require a larger corpus to train without overfitting. We started by creating our own CS corpora to train the embedding models using Arabic-English Code-switched text. We collected the data using three different techniques/sources and created two corpora. The first corpus \textit{CS\_TRAIN} is composed of 105 million tokens. The first source of data was using the CS corpus of \cite{hamed2019code} collected from social media platforms. The second technique was generating 30 million Arabic-English CS tokens by translating monolingual Modern Standard Arabic corpus into Arabic-English CS data. We iterated automatically over the corpus and translated tokens according to a set of linguistic constraints using an open-sourced Neural Machine Translation API\footnote{https://rapidapi.com/gofitech/api/nlp-translation}. We followed the linguistic constraints inferred from evaluating real Arabic-English code-switched data in \cite{hamed2018collection}. They defined trigger words such as (\<ال>, the),
(\<في>, in) and (\<و>, and), which are Arabic words preceding a CS point as shown in Table \ref{table:trigger}. 

\begin{table}[!ht]
   
   \begin{center}
       \caption{List of trigger words each with percentage probability of preceding code-switching \cite{hamed2018collection}}

    \begin{tabular}{l|r}
    \hline
\bf    Trigger word   & \bf Percentage \\
    \hline
\< ال >
  (the) 
  & 31.0\%\\
  \< في >
   (in)
   &04.8\% \\
     \< و >
       (and)
  & 03.4\% \\
    \< يعني >
    (means)
   &01.5\% \\
     \< هو >
     (he)
      &01.3\% \\  \hline
    \end{tabular}
        \label{table:trigger}
    \end{center}

\end{table} 

The final part of the corpus is composed from the monolingual Arabic news-wire data. To augment the size of the training data, we created another corpus \textit{CS\_TRAIN\textsubscript{++}} with 20 million tokens from the same Arabic monolingual news-wire data-set by adding to the initial one. Also, we translated them following the same linguistic constraints to form the other 19 million CS tokens. The corpus of \textit{CS\_TRAIN\textsubscript{++}} is composed of 144 million tokens. The corpora statistics are presented in Table \ref{table:datastatistics}.
\begin{table}[!ht]
     
        \begin{center}
    \caption{Detailed data statistics about the number of English tokens, Arabic tokens, the number of sentences}
    \begin{tabular}{ l|r|r|r}
    \hline
  \bf   Data-set     & \bf English Tokens &\bf Arabic Tokens & \bf Sentences  \\
    \hline
\textit{CS\_TRAIN}      & 10M & 95M & 7M\\
\textit{CS\_TRAIN\textsubscript{++}}  & 17M & 127M & 9M\\ \hline
  \end{tabular}
     \label{table:datastatistics}
  \end{center}
    
\end{table}
\section{Trained Embedding Models}
We trained several bilingual contextual embedding models to compare them and have an efficient model for Arabic-English CS data used in several NLP tasks and specially our NER task. We used several state-of-the-art techniques and the type of contextual embeddings for building our models. We started by creating a baseline model using a stack of Arabic pre-trained Pooled Flair embedding and pre-trained FastText. The other models we built used Contextual string embeddings, BERT, and ELECTRA. We also proposed a new model called the KERMIT.

\subsection{Baseline}
The baseline model used in our experiment is a stack of Arabic pre-trained Pooled Flair embedding and pre-trained FastText. Pooled Flair embedding and FastText achieved the best results on our NER task for Arabic-English CS data. It even outperformed the only available Arabic-English CS embedding of Bi-CS. Pooling operation dynamically aggregates each unique word encountered, then retrieves previous embeddings produced from memory. Finally, the pool operation is performed, and all locally contextualized embeddings are concatenated to make the final embedding for the token. 

\subsection{Contextual String Embeddings}
The first model trained on our corpus was Contextual string embeddings.
The pre-processing stage involved removing punctuation symbols and lower casing English characters in the corpus. In the tokenization phase, we listed all the alphabetical English and Arabic characters. Hyper-parameters were configured following the original Contextual string embeddings parameters \cite{akbik2018contextual}. Forward and Backward language models were trained using the same vocabulary. The language models were trained using SGD to perform truncated back-propagation we tracked the training performance on the validation set. Training took ten days for each of the backward and forward models on one GPU or halted when negligible gains were observed. 

\subsection{BERT}
We trained BERT on our CS corpora.
We implemented an additional pre-processing stage to our data set before proceeding to the training phase. This stage involves segmenting Arabic words using Farasa segmenter \cite{abdelali2016farasa} to remove redundant forms of words. After segmenting the corpus, we produced a total of 64k tokens as a vocabulary for our model. We trained BERT\textsubscript{BASE} sized model with 12 encoder layers, a hidden size of 768, and 12 attention heads. Training BERT was done by masking 15\% of tokens in the input, of which 80\% were replaced with a special token [MASK], 10\% replaced with random tokens, and 10\% remaining as an original token. This configuration helped to prevent pretrain-finetune-discrepancy. A learning rate of 2e-5 and Adam optimizer was used in the pre-training. We trained our model for a total of 1,000,000 steps with a batch size of 128. The next 250,000 steps were then trained with a batch size of 256 to speed up the training process. The training took three days to complete on eight cloud TPUs.

\subsection{ELECTRA}
We trained ELECTRA \cite{clark2020electra} on our CS corpora. The pre-training stage was similar to BERT, however, with different configurations. We used the same pre-processing and tokenization mechanism. We trained ELECTRA\textsubscript{BASE} sized model. The discriminator of this model has the same size as BERT\textsubscript{BASE} model. It is composed of 12 encoder layers, a hidden size of 128, 12 attention heads, and outputs embedding of size 768. The generator component of the ELECTRA\textsubscript{BASE} model is 1/3 of the size of the discriminator model. This configuration makes it harder for the discriminator component to distinguish tokens and produces more robust representations. We trained both the generator and discriminator jointly from scratch with a learning rate of 2e-4 and batch size of 256 for a total of 750,000 steps. This training took five days on eight TPUs.

\section{New Proposed Embedding Model: KERMIT}

\begin{figure}[!ht]
  \centering
  \subfloat[ELECTRA model]{\includegraphics[width=0.49\textwidth]{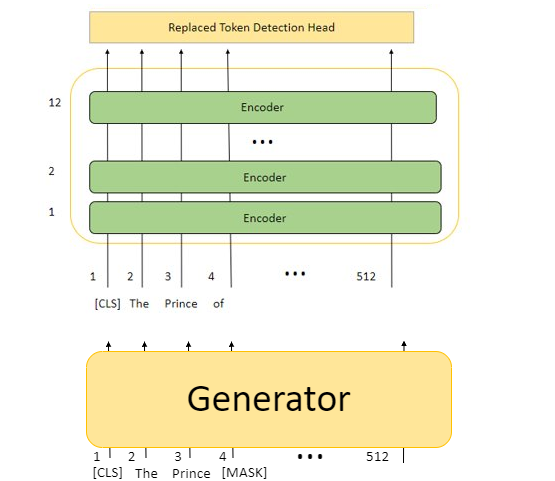}\label{fig:transfer1}}
  \hfill
  \subfloat[BERT model]{\includegraphics[width=0.49\textwidth]{bert_diagram.PNG}
  \label{fig:transfer2}}
 \caption{KERMIT encoder layers trained as discriminator in ELECTRA then as Generator in BERT } \label{fig:transfer}
\end{figure}

We proposed a novel model called KERMIT for producing word embeddings. The architecture of this model is an encoder variant of transformers. The pre-training of this model is divided into two stages, as shown in Figure \ref{fig:transfer}. In the first stage (Figure \ref{fig:transfer1}), KERMIT is trained as a discriminator in ELECTRA architecture using RTD and MLM tasks. After pre-training, the generator is dropped, and the pre-trained discriminator weights are used for the next stage. In the second stage (Figure \ref{fig:transfer2}), we initialized the encoder and embedding layers of the BERT model with the trained discriminator model. Then, we further trained the model as a generator using the MLM task. This training mechanism helps avoid pretrain-finetune-discrepancy and allows the model to learn stronger representations through training on more than one task.
Furthermore, the model is trained on data with different augmentations. A generator replaced tokens at the first stage, and during the second stage, tokens were masked out. This helped increase the diversity of data available for training.
In the two stages of pre-training, we used the same data pre-processing and tokenization vocabulary. We have trained the ELECTRA\textsubscript{BASE} model using the same hyper-parameters used for training the previous ELECTRA model. Using the same vocabulary, we transferred weights of the encoder and embedding layers and trained BERT\textsubscript{BASE} model using the same hyper-parameters used for training the previous BERT model. We investigated different configurations for the transferred model. On the one hand, we trained the BERT model using trained generator prediction layers. On the other hand, we trained the model using randomly initialized generator prediction layers. Furthermore, we have changed the Adam optimizer variables so that loss affects only the encoder layers and applies less effect on the output layers. We also tried to switch training stages through training BERT first then transferring weights to the ELECTRA model. The best configuration observed while training was using both randomly initialized generator prediction layers along with Adam optimizer variables.

\section{Evaluation \& Results}
To compare our bilingual embedding models, we applied intrinsic and extrinsic evaluation methods. In the intrinsic evaluation, the models were tested on their ability to assign near-by vectors to similar tokens. This was achieved by calculating the cosine similarity between the tokens using BERTS\textsubscript{CORE} \cite{zhang2019bertscore}. In the extrinsic evaluation, we evaluated our models on three downstream NLP tasks; Named Entity Recognition, Sentiment Analysis, and Question Answering on Arabic-English CS text. 
To evaluate our models, we used intrinsic and extrinsic methods. We first trained the models using the \textit{CS\_TRAIN} and then using \textit{CS\_TRAIN\textsubscript{++}} for clarity. All models with a suffix ++ have been trained using \textit{CS\_TRAIN\textsubscript{++}}.
\subsection{Intrinsic Evaluation}

The purpose of the intrinsic evaluation was to visualize and test how well the word embedding model can capture similarities in CS Arabic-English text. Thus, to evaluate our embedding models, we leveraged cosine similarity between different tokens using BERTS\textsubscript{CORE} \cite{zhang2019bertscore}. 
This is a new automatic evaluation metric for the generated text that computes token similarity using contextual embeddings. We evaluated the similarity between the tokens of several sentences, and the models performed the same in almost all of them. The following is an example of two CS sentences used to test the similarities stated by the models:\vspace{0.5cm}\\
\begin{tabular}{ccc}
day \<كل> news \<احمد يحب قراءه>\\
'Ahmed loves reading news everyday'
\end{tabular}
\quad
\begin{tabular}{ccc}
\<ينهي كتاب في أسبوع> Mohamed \\
'Mohamed finishes a book in a week'
\end{tabular}
\vspace{0.5cm}

\begin{figure}[h!]
  \centering
  \subfloat[Contextual string embeddings and FastText]{\includegraphics[width=200 pt,height=200 pt]{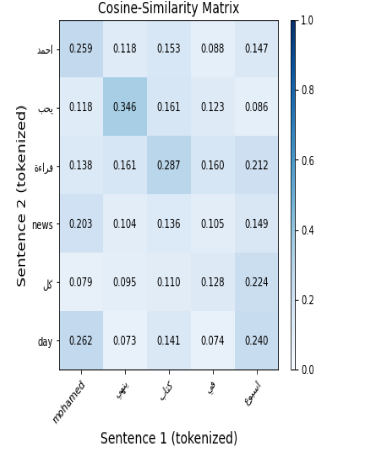}  \label{fig:cos_simFLAIRFast}}
  \hfill
  \subfloat[Pooled Flair and FastText]{\includegraphics[width=200pt,height=200 pt]{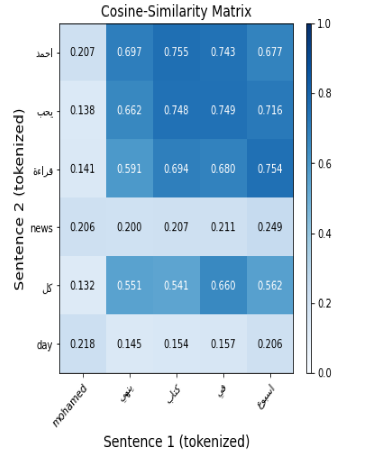} \label{fig:cos_simFLAIR}}
\caption{Results of comparing the tokens of two sentences of different embeddings by calculating the cosine-similarity of BERTS\textsubscript{CORE}}

\end{figure}

\begin{figure}[h!]
  \centering
  \subfloat[BERT]{\includegraphics[width=200 pt,height=200 pt]{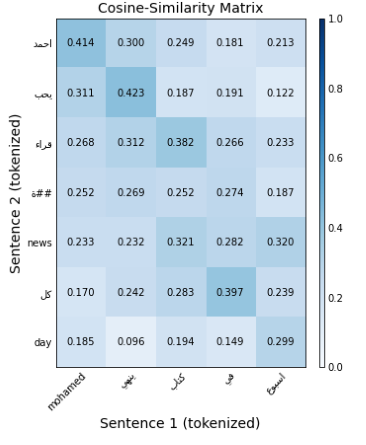}}
  \hfill
  \subfloat[BERT\textsubscript{++}]{\includegraphics[width=200pt,height=200pt]{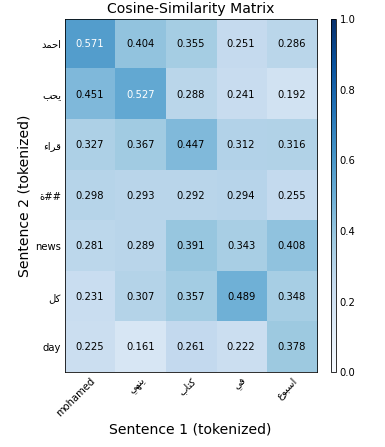}}
\caption{Results of comparing the tokens of two sentences using BERT embedding by calculating the cosine-similarity of BERTS\textsubscript{CORE}}
 \label{fig:cos_simBERT}
\end{figure}

\begin{figure}[h!]
  \centering
  \subfloat[ELECTRA]{\includegraphics[width=200 pt,height=200 pt]{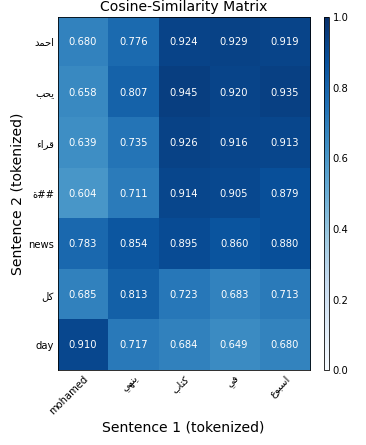}}
  \hfill
  \subfloat[ELECTRA\textsubscript{++}]{\includegraphics[width=200pt,height=200pt]{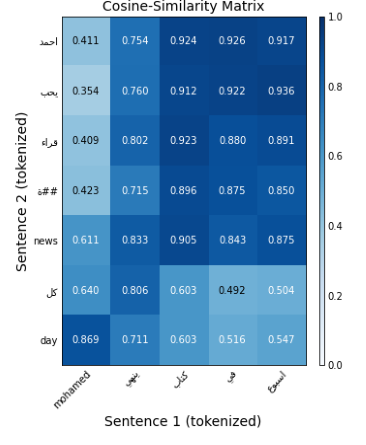}}
\caption{Results of comparing the tokens of two sentences using ELECTRA embedding by calculating the cosine-similarity of BERTS\textsubscript{CORE}}
 \label{fig:cos_simELECTRA}
\end{figure}

\begin{figure}[h!]
  \centering
  \subfloat[KERMIT]{\includegraphics[width=200 pt,height=200 pt]{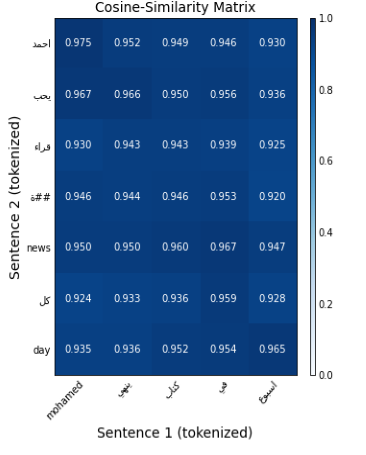} \label{fig:cos_simKERMIT2}}
  \hfill
  \subfloat[KERMIT\textsubscript{++}]{\includegraphics[width=200pt,height=200pt]{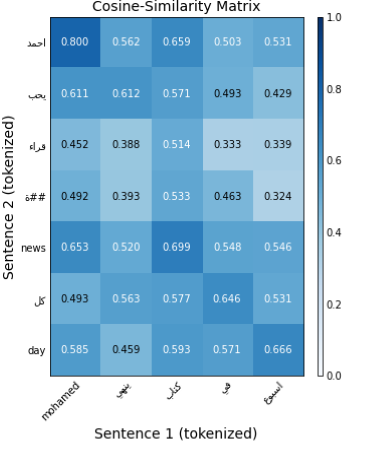} \label{fig:cos_simKERMIT1}}
\caption{Results of comparing the tokens of two sentences using KERMIT embedding by calculating the cosine-similarity of BERTS\textsubscript{CORE}}
 \label{fig:cos_simKERMIT}
\end{figure}

Examining the results in Figures \ref{fig:cos_simFLAIR} to \ref{fig:cos_simKERMIT}, we can see how word embeddings can help state the similarity between CS words. All the models trained with \textit{CS\_TRAIN\textsubscript{++}} performed better than those trained with the smaller data-set \textit{CS\_TRAIN}. We also noted in Figures \ref{fig:cos_simFLAIRFast} that the baseline model using Contextual string embeddings and FastText did not get the similarities of the English words. It is an expected result as it was trained on monolingual Arabic data. On the one hand, the similarities of the Contextual string embeddings model between all words were low. 


 On the other hand, as shown in Figure \ref{fig:cos_simBERT} BERT\textsubscript{++} is capable of capturing small similarities between tokens, including cross-lingual. As illustrated in Figure \ref{fig:cos_simELECTRA}, the similarities obtained by the ELECTRA model were high.

 Moreover, training KERMIT with the smaller corpus did not achieve good results as it modeled wrong similarities between unrelated words as shown in Figure \ref{fig:cos_simKERMIT2}. Also, as shown in Figure \ref{fig:cos_simKERMIT1} KERMIT\textsubscript{++} is the best one; as it was capable of modeling higher cosine-similarity with similar words and lower for unrelated ones. 

\subsection{Extrinsic Evaluation}
The extrinsic evaluation is based on the possibility of using the word embeddings in downstream NLP tasks. We evaluated the impact of our models in three tasks, Named Entity Recognition, Sentiment Analysis and Question Answering on Arabic-English CS text. As there is no available contextual embeddings trained on Arabic-English CS data, we evaluated the recent Arabic model AraBERT \cite{antoun2020arabert} in the three tasks to compare its performance with our models.

\subsubsection{Named Entity Recognition}
 We used two NER approaches to evaluated our embeddings. The first one is based on feature extraction approach, where a set of features are extracted from the pre-trained embedding model. The NER tagger we used is our system as stated in Chapter 5 and that we want to enhance. It is based on BiLSTM and CRF along with the pre-trained word embedding layer. Previously, the best embedding types were using Pooled Flair and FastText pre-trained Arabic embeddings, which we call our baseline. The best performance achieved an F1-score equal to 77.69\%. While training the model we used Adam optimizer and a learning rate equal to 3e-5 with a transformer-based embedding and 0.1 for Contextual string embeddings with an annealing factor equal to 0.5. Regarding the number of epochs, we trained the model a total of 50 epochs while using the transformers embeddings and 110 epochs for the Contextual string embeddings on a batch size equal to 16. We used the same model and the training and testing data to compare the effect of using our different embeddings models on the performance. 
 
 The second approach uses the NER system presented by the HuggingFace NLP library that is based on fine-tuning \cite{Wolf2019HuggingFacesTS}. The fine-tuning approach relies on adding a simple classification layer to the pre-trained embedding model and all parameters are fine-tuned together on the downstream task which is here the NER. Our model was fine-tuned by adding a simple feed-forward layer with Softmax function to get the probability distribution over the output classes predicated. We trained this model for a total of 4 to 10 epochs using Adam optimizer. We evaluated the best configuration of the learning rate from [1e-4, 5e-5, 3e-5, 2e-5], batch size from [16,32] and max sequence length of 64. Concerning the rest of the hyper-parameters they were configured as recommended in BERT \cite{devlin2018bert}. Also, to initialize the linear variable, we tested 30 random seeds and chose the best model out of the 30 models. We tested this system using the same training and testing data. 
\begin{table}[!ht]
  
  \begin{center}
\caption{The F1-score results (\%) of using the different types of embeddings in the two NER systems}
 \resizebox{\textwidth}{!}{
    \begin{tabular}{l|r|r}
    \hline
  \bf   Embedding Model   &\bf  NER Model &\bf  NER HuggingFace \\
    \hline

BERT      & 64.50 &76.50\\
BERT\textsubscript{++} & 68.20 &77.10\\
AraBERT & 58.00 & 66.70\\

KERMIT   & 65.00&78.70\\ 
KERMIT\textsubscript{++} &69.00&  \textbf{79.40}\\
ELECTRA &67.00&75.40\\
ELECTRA\textsubscript{++} &68.30&76.29\\
Pooled Flair \& FastText  & 77.69 & -\\
Contextual string embeddings   &72.55 &-\\
Contextual string embeddings  \& FastText &\textbf{78.20} &-\\
KERMIT\textsubscript{++} \& Contextual string embeddings & &\\
\& FastText & 76.60&-\\\hline

    \end{tabular}}
     \label{table:ner}   
  \end{center}
    
\end{table} 
As shown in Table \ref{table:ner}, we used all the models one by one as the embedding layer in the NER system, and we also combined some models. We combined Contextual string embeddings with Arabic FastText embedding, and it achieved the best results and enhanced the NER model by 0.51\%. All the models trained on \textit{CS\_TRAIN} achieved lower F1-score than the ones trained on \textit{CS\_TRAIN\textsubscript{++}}. The results of using BERT\textsubscript{++}, ELECTRA\textsubscript{++}, and KERMIT\textsubscript{++} are all less than the baseline model, but the three are almost the same with minimal differences, achieving an F1-score equal to 68.2\%, 68.3\% and 69\% respectively. We combined the highest one of them, KERMIT\textsubscript{++}, with the highest model of Contextual string embeddings and FastText. The results improved but were still less than the baseline. 

However, in the HuggingFace system, it was hard to add the Pooled Flair and Contextual string embeddings to their fine-tuning model since the pooling is over past embeddings. Thus, it is hard to fine tune all of them due to memory limitations. As a result, KERMIT\textsubscript{++} outperformed all other transformer-based models and even achieved higher results than the ones presented by the first NER approach by 1.2\% and achieved an F1-score of 79.4\%, setting new state-of-the-art results on code-switch Arabic-English NER task. However, not all model architectures could be used or represented by a transformer architecture. Also, the computational cost is expensive compared to the feature extraction-based approach.

\subsubsection{Sentiment Analysis}
We also evaluated our models on sentiment analysis (SA) tasks using the methods proposed. This is a text classification task, where parts of the text are identified and marked with binary labels to represent the sentiment type. SA can capture the context, in a given text, and predict a label to summarize the context. The labels are based on the annotated corpus that the model was trained on. The models are trained as a multiclass labeling. For instance, we can use a SA model to state whether a text review is negative, positive or neutral in a book reviews.

We created our data-set by translating a monolingual Arabic data-set of book reviews \cite{aly2013labr} to have CS text using the same machine translation API and the linguistic constraints previously described in the data collection section. 
The results of experimenting sentiment analysis are stated in Table \ref{table:sentiment}.


We used two approaches to apply this task. The first one used the framework of FLAIR \cite{akbik2019pooled}. Based on the Recurrent Neural Network and linear classifier layers. It is a feature extraction approach, where only the output layers are trained. We added an RNN layer, a linear dense layer as output and a Softmax layer. The hyper-parameters were configured such that the learning rate was equal to 0.1 and trained for 10-15 epochs. While we added again the pre-trained FastText embedding to the Contextual string embeddings, the performance of the sentiment analysis model improved and the model scored the highest F1-score of 88.8\%. 

In the second approach, we used the sentiment analysis system presented by HuggingFace based on fine-tuning (Linear Classifier) \cite{Wolf2019HuggingFacesTS}; all transformer-based models were fine-tuned by adding a linear classifier layer. We added a feed-forward layer with standard Softmax to get the probability distribution over the predicted output classes. We fine tuned the models for 2-3 epochs with a learning rate of 2e-5 and Adam optimizer. BERT model outperformed ELECTRA by about 1.6\% and KERMIT by 0.9\%. Furthermore, BERT\textsubscript{++} scored higher than ELECTRA\textsubscript{++} by 0.4\%. Finally, KERMIT\textsubscript{++} scored the highest in the second approach. However, using the first approach with Contextual string embeddings and FastText outperformed all models experimented with and used. 

\begin{table}[!ht]
   
       \begin{center}
            \caption{The F1-score results (\%) of using the different types of embeddings in the two sentiment analysis systems}   
     \resizebox{\textwidth}{!}{
    \begin{tabular}{l|r|r }
    \hline
\bf     Embedding Model  & \bf SA HuggingFace & \bf SA FLAIR Framework \\
    \hline

BERT     & 77.70 &-\\
BERT\textsubscript{++}& 77.38 &-\\
AraBERT & 78.20 & -\\

KERMIT   & 76.80 & -\\ 
KERMIT\textsubscript{++}   &\textbf{79.90}&-\\ 
ELECTRA  & 76.39 & -\\
ELECTRA\textsubscript{++}  & 77.18 & -\\
 Pooled Flair and FastText & - & 87.84 \\
 Contextual string embeddings   & - & 87.80  \\
 Contextual string embeddings  and FastText & - &\textbf{88.80}\\
\hline
    \end{tabular}}
      \label{table:sentiment}
   \end{center}
  
\end{table}
\subsubsection{Question Answering}
Another important task where we tested the effect of using our embeddings models as the question-answering approach which refers to finding the part of the text in a context that answers a given question. This could be done by predicting the special tokens indicating the start and end surrounding the answer sequence. A model is given a context and is required to predict a sequence in the context that has the correct answer to a defined question.  

There are very few code-switched corpora for a  question-answering task and none for Arabic-English language pairs. For instance,
\cite{raghavi2015answer} collected 3000 Hindi questions from a TV show and added some questions from the Central Board of Secondary Education (CBSE). These questions were then converted to Hindi-English CS data with the help of 30 students volunteer. The volunteer students were asked how to pose these questions in a CS language. Finally, these questions were filtered and a total of 1000 questions were collected. Also, \cite{banerjee2016first} relied on social media for producing such low resource corpus of Bengali-English CS corpus. For social media tweets, blogs and forums were all used to acquire question-answer sets and have fine-tuned a language mix ratio to ensure only CS data are collected.

We created our own data-set by translating text from the questions of the monolingual Arabic Question Answering (QA) corpus presented by \cite{mozannar2019neural} following the same approach stated in the sentiment analysis task. We could not translate words from the answers, as this would lead to changing their order, which could be misleading for the question-answering task. 

We used only one approach for this task, which uses the question answering system of HuggingFace, which is based on fine-tuning (Linear Classifier) \cite{Wolf2019HuggingFacesTS}. The model was fine-tuned by adding a linear dense layer, a normalization layer and a Softmax function. The hyper-parameters were configured such that the learning rate was equal to 3e-5, batch size equal to 12 and number of epochs equal to two.

As shown in Table \ref{table:question}, KERMIT\textsubscript{++} achieved the highest score as compared to all other fine-tuned models with an F1-score equal to 39.9\%. It outperformed all models because, as stated, before training KERMIT as both generator and discriminator, it helped the model to compute better representation of the data. All the results of using the different embeddings are low as the answers in the training data did not contain CS behaviour, which could be enhanced using more accurate translation techniques without affecting the order of words. As shown in the results our models achieved higher results than AraBERT embedding in all the three tasks. This is due to the fact that AraBert is trained on monolingual Arabic data.


\begin{table}[!ht]

    \begin{center}
        \caption{The F1-score results (\%) of using the different types of embeddings in the question answering system}
    \begin{tabular}{l|r}
    \hline
   \bf   Embedding Model &\bf  QA HuggingFace  \\
    \hline

BERT     & 37.8 \\
BERT\textsubscript{++}     &38.1 \\
AraBERT & 30.0\\
KERMIT   &32.8 \\ 
KERMIT\textsubscript{++}   &\textbf{39.9}\\
ELECTRA  & 34.1 \\
ELECTRA\textsubscript{++}  &37.2 \\
 \hline
  \end{tabular}
    \label{table:question}
\end{center}
     
\end{table}

\subsection{Discussion}
Throughout the pre-training process using downstream tasks, we evaluated the different training configurations. Concerning BERT, it achieved better results in downstream tasks while training it, using randomly initialized output layers compared to training it using trained output layers. In addition, using for the output layer an Adam optimizer to decrease momentum and concentrate on training encoders did not improve the performance of the downstream tasks. Furthermore, in KERMIT, switching the stages by training BERT first then ELECTRA has shown very low performance on downstream tasks. The best configuration was to train the ELECTRA model then transfer it to BERT with randomly initialized output layers.
\begin{figure}[!ht]
\subfloat[Example A]{\includegraphics[height=6.5cm]{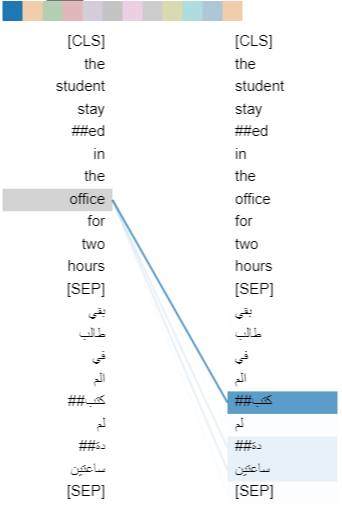}}
\subfloat[Example B]{\includegraphics[height=6.5cm]{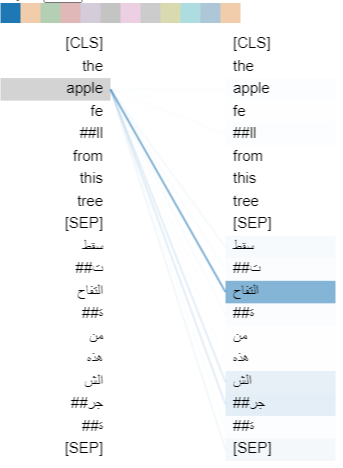}}
\subfloat[Example C]{\includegraphics[height=6.5cm]{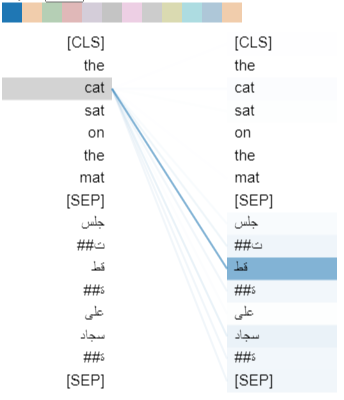}}
\caption{Visualizing dependencies between different language tokens using KERMIT\textsubscript{++}}
\label{fig:examples}
\end{figure}
After training, we evaluated the resulting embedding representation using attention visualization tool \cite{vig2019multiscale}. The KERMIT\textsubscript{++} model has been capable of outputting strong word representation and could map cross-lingual vectors together as shown in the randomly selected examples in Figure \ref{fig:examples}. Examples A, B and C show different context in English and Arabic. Computing the attention of KERMIT has shown the most similarity between vectors of English words and their equivalent Arabic words of the following sentences:
\newline
\textbf{Example A:}
\newline
The student stayed in the office for two hours \newline
 \< بقى طالب في المكتب  لمده ساعتان>\newline
\textbf{ Example B:}
\newline
The apple fell from the tree \newline
 \<
سقطت التفاحه من هذه الشجره>\newline
\textbf{ Example C:}
\newline
The cat sat on the mat \newline
\<
جلست قطه على سجاده
>
\section{Summary}
We proposed a solution to compute a large corpus for pre-training CS word embeddings through mainly translation techniques to enhance our NER model. We used this corpus to train different language models: Contextual string embedding, BERT, and ELECTRA. We also proposed our new language model KERMIT. This model is trained as both a discriminator and a generator where each task has a different data augmentation. The different data augmentations helped the KERMIT model to train efficiently on the collected corpus. We have pre-trained and evaluated the KERMIT model on NER, sentiment analysis, and question answering.
\begin{figure}[ht!]
    \centering
 \includegraphics[width=12.3cm,keepaspectratio]{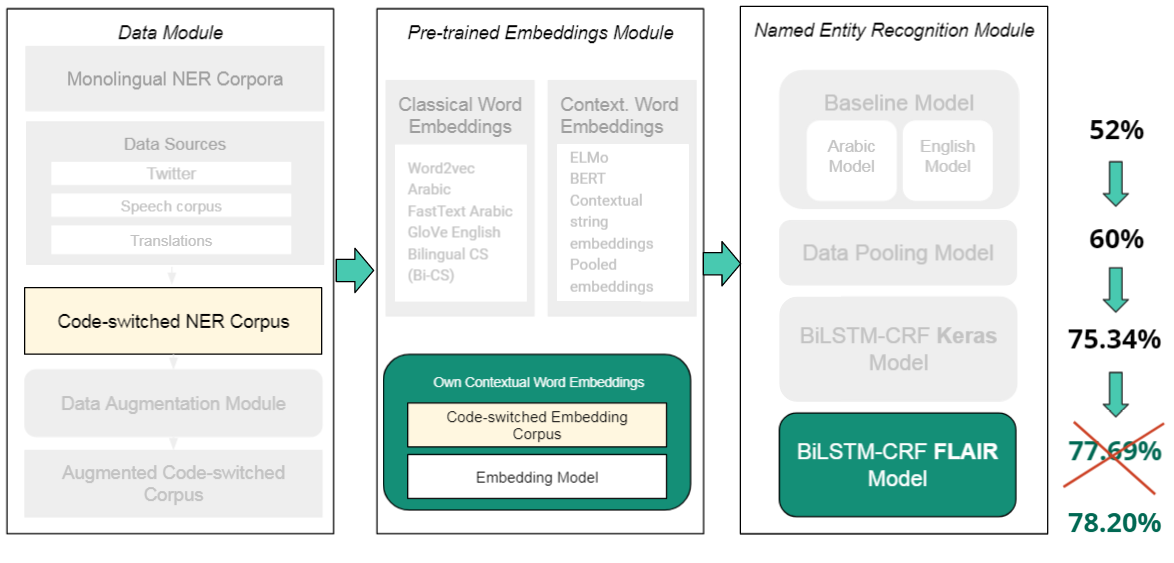}
    \caption{Results of First Enhancement Technique}
    \label{fig:ContEmbR}
\end{figure}
Compared with other models, KERMIT has scored the highest F1-score on both NER and question answering tasks. As shown in Figure \ref{fig:ContEmbR} KERMIT has also outperformed the previous NER baseline with an F1-score of 79.4\%.

%% file: Data-Augmentation.tex

\chapter{Data Augmentation Techniques on CS Data for NER}
\label{chap:dataaugmentation}
\chaptermark{Data Augmentation Techniques}
The main contribution in this part is tackling the lack of Arabic-English CS annotated data and enhancing the performance of our NER task on CS data. Most of the state-of-the-art NER approaches rely on deep neural network models, which require having large training data-sets. Also, large labeled data is needed to avoid over-fitting the models \cite{ratner2017learning}, which is often not available. In addition, the need for a large amount of data is related to the complexity of the task to be solved and is not only specific to deep learning. 
Producing large annotated data for the NER task is challenging, as the collection and labeling processes tend to be expensive and time-consuming \cite{qiu2020easyaug}. 
\begin{figure}[ht!]
    \centering
 \includegraphics[width=12.5cm,keepaspectratio]{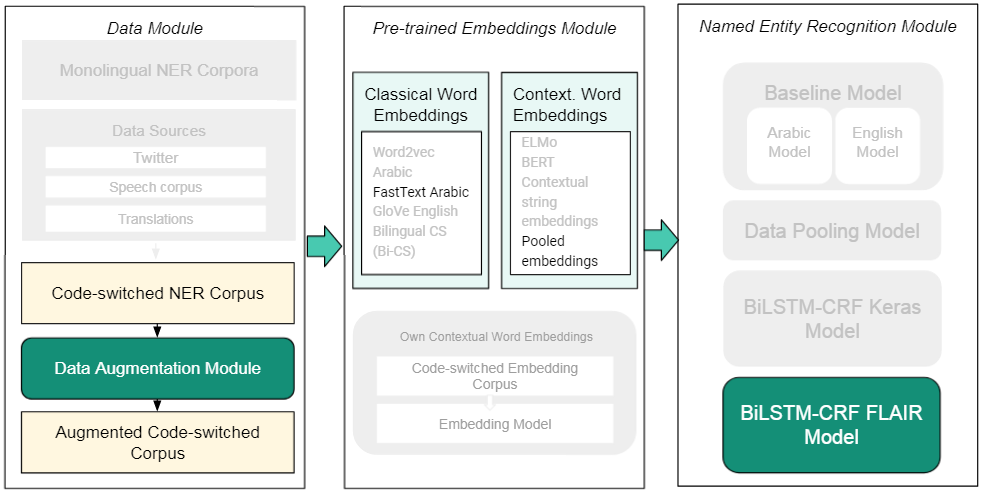}
    \caption{Second Enhancement Technique}
    \label{fig:DataAug}
\end{figure}

In this chapter, we present our proposed data augmentation techniques and show the effect of applying these techniques on the CS training data by evaluating the performance of the NER tagger as shown in Figure \ref{fig:DataAug} \cite{sabty2021Data}.


\section{Data Augmentation Techniques}
We introduced three data augmentation techniques used to substitute entities and synthesize new annotated contexts. We used our original training data that we created and discussed in Chapter 5, with a size of 5,306 sentences containing 14,34 entities. We used this data-set as the source of the data to which we applied the data augmentation techniques. The same deep learning NER model we introduced in Chapter 5 was used to test the effect of different data augmentation techniques and compare the performance with and without DA. The first technique was Word Embedding substitution to replace entities with new similar ones.
Hence, the label of new words will be the same as the initial entities. In the second technique, we modified some of the operations of the Easy Data Augmentation (EDA) technique \cite{wei2019eda} to generate new annotated sentences. In the last technique, we explored applying back-translation (BT). We translated the sentences into one or two intermediate languages and translated them back to form new sentences. If the newly added words resulting from the last two techniques are available in our original data, we take the same label. In case the word is a new one, we use two different NER taggers for Arabic and English. We evaluated different setups for these techniques and checked their effect on the performance of the NER model.



The three techniques start by pre-processing the data to be given as input to the augmentation module, and at the end, the augmented sentences are tagged in the tagging module. In the first technique, the augmented word is assigned the same NER tag as the original word. In the other two techniques, in case the word is a new one, we used StanfordNER tagger~\cite{finkel2005incorporating} for English data and an implemented NER model that was trained on ANERCorp data for the Arabic language. Otherwise, we take the same label from the original data-set.


\subsection{Word Embedding Substitution}
Inspired by the capabilities of pre-trained word embedding models, we implemented two word embedding substitution techniques to enrich our training data using the classical embedding (FastText) and two contextual embeddings (BERT and KERMIT).
\subsubsection{Classical Embedding}
The first technique uses FastText embedding. 
We created two variations of this technique using FastText. The first one is $Full\_WE_{sub}$, which replaces all the words of the sentences. The second one is $Analogies\_WE_{sub}$, which replaces the entities only. 
\begin{figure}[!ht]
    \centering
 \includegraphics[scale=0.48,keepaspectratio]{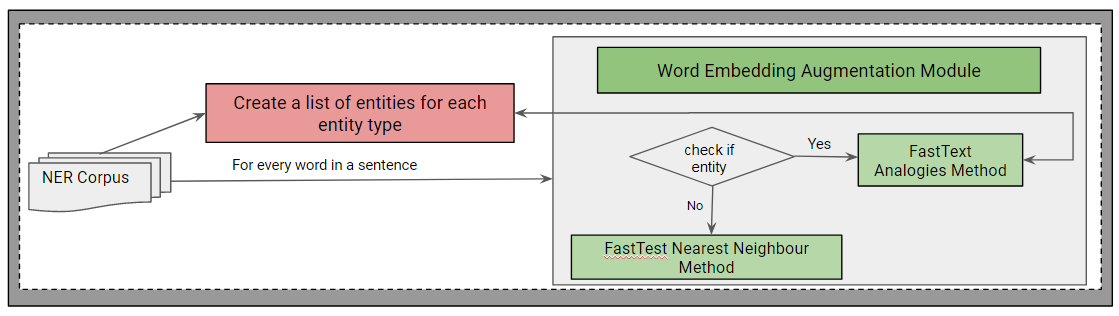}
    \caption{Architecture of the Classical Word Embedding Substitution Technique}%
    \label{fig:DAWE}%
\end{figure}
As illustrated in Figure \ref{fig:DAWE}, the main architecture of this technique starts by creating a list for every entity type and collects all entities with the same type from the original data. Then, for each word in a sentence, it checks whether it is an entity or not. In case it is an entity, its language is detected. Afterward, it uses the word analogy method of FastText of the corresponding detected language. The method input is the current word, and two randomly selected words from the list containing the entities of the same type as the current one. The method then predicts a fourth word that might be with the same entity type. For instance, if we have the word \textit{John}, which is an entity of type Person, we randomly get from the list containing all persons, two other words \textit{David} and \textit{Martin} that are all given to the embedding method, then as output, we get the word \textit{Luther}. In the end, we replace the original word with the new predicted one. In case it is not an entity, in the model of $Full\_WE_{sub}$, it will be replaced with a similar word in the vector space. After investigating the most similar word (n=1), especially for the Arabic language, we did not find a remarkable change in the replaced words. This is due to having almost the same word but with an added prefix or suffix. Thus, we selected the fifth similar word (n=5) for replacement. In the second model $Analogies\_WE_{sub}$, no replacement was applied if the word was not an entity.
\subsubsection{Contextual Embedding}
The second technique uses two pre-trained contextual transformer-based embedding models BERT and KERMIT presented in Chapter 6 each alone $BERT_{sub}$ and $KERMIT_{sub}$; they are both trained on AR-EN CS data. 


Both models have a transformer-based architecture trained using an attention mechanism that helps capture features and supports long sequences.  
This technique augments entities only. It works auto-regressively by augmenting one entity at a time, which helps to preserve dependencies between augmented tokens. This approach utilizes knowledge captured by a pre-trained language model without any further fine-tuning. First, we mask an entity to augment while a word piece model tokenizes the sequence for the contextual models. A word piece tokenizer works by splitting words into subsections to model out-of-vocab words. The contextual model predicts the masked token by outputting the embedding with the highest probability from the vocabulary of 64000 consisting of both Arabic and English tokens. 
The model has been configured to output the top 10 tokens with the highest probability. To validate the fact that the predicted token has the same type as the entity, we added a module that selects from the top 10 tokens similar to the original masked one while ignoring sub-tokens. This module comprises our selected NER taggers for both languages. The added module conditions the outputted selected token to be of a specific entity.

\subsection{Modified Easy Data Augmentation Technique}
We implemented a modified version of the Easy Data Augmentation (EDA) technique \cite{wei2019eda}. The new model handles two languages in the same sentence and supports the Arabic language. The initial EDA technique consists of four different text editing operations:
\begin{itemize}
    \item \textbf{Synonym Replacement (SR)} refers to choosing \textit{n} random words to be replaced by one of their synonyms in the sentence.
    \item \textbf{Random Insertion (RI)} refers to choosing a random word to insert one of its synonyms in a random index in the sentence. It is repeated \textit{n} times.
    \item \textbf{Random Swap (RS)} refers to choosing two random words and swapping their positions in the sentence. It is repeated \textit{n} times.
    \item \textbf{Random Deletion (RD)} refers to choosing random words to be removed from the sentence based on a certain probability.
\end{itemize}

The operations could be repeated \textit{n} times. The value of \textit{n} is calculated based on the length of the sentence (l) with the formula \(n=\alpha \times l\), where \(\alpha\) indicates the percentage of words to be changed in a sentence. The number of iterations per operations on original sentence is $N$, it is calculated using the total number of needed augments $Num\_Aug$ (number of required augments for one sentence) divided by the selected operations. For example, if the selected operations are two, then each one will be repeated \(Num\_Aug/2\).

\begin{figure}
     \centering
  \includegraphics[scale=0.65]{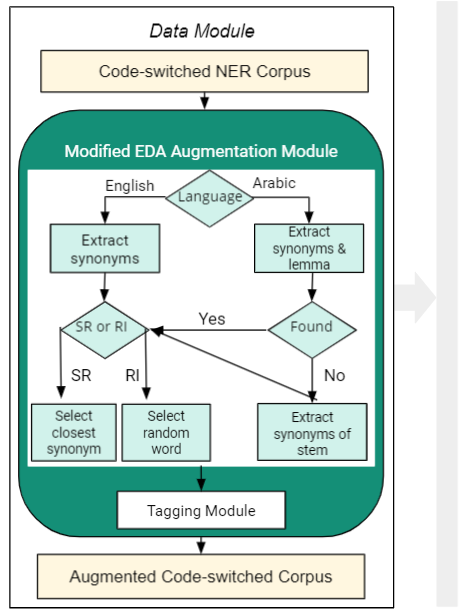}
    \caption{Architecture of the Modified EDA Augmentation Technique}%
    \label{fig:ArchitectureEDA}%
 \end{figure}

The main modifications we applied were in the two operations of SR and RI. 
As shown in Figure \ref{fig:ArchitectureEDA} synonym extraction process is done for each of the SR and RI operations.
This process starts by detecting the language of each word in the sentence. Based on the detected language, if it is English, then the actual extraction process of~\cite{wei2019eda} is used. They rely on WordNet~\cite{miller1995wordnet} to search for synsets for a given word. However, we added another source to search for more synonyms, WikiSynonyms,\footnote{\url{http://wikisynonyms.ipeirotis.com/}} which relies on extracting synonyms from Wikipedia pages. If the detected language is Arabic, we implemented synonym extraction using WikiSynonyms and Arabic WordNet (AWN)~\cite{elkateb2006building}. AWN represents an XML file acting as a database containing Arabic words along with their synsets, antonyms, and hyponyms. Hyponyms are sometimes considered as synonyms for a word, e.g., ( \<شارع, طريق   >) which means (road, street). We removed the diacritics from all Arabic words in the AWN as our original data is without diacritics.
As words can be composed of prefixes, stem, and suffixes. Extracting the stem of the words is not a straight-forward approach as articles, pronouns, prepositions, or coordinating conjunctions could be found as part of the word. Thus, to find the words and get their synonyms using the previously mentioned approaches, we had to extract the stem and lemma of the words using the lemmatizing method presented in~\cite{zerrouki2010tashaphyne} and the stemmer algorithm described in~\cite{taghva2005arabic}. For example, we will not find the word in the AWN the word (\<مصرنا>) (\textit{our Egypt}), and we should extract its lemma first which is (\<مصر >) (\textit{Egypt}).

\vspace{-0.18cm} 
After applying the synonym extraction, we combined all returned lists of synonyms in the SR operation. We used a word similarity method\footnote{\url{https://bitbucket.org/yunazzang/aiwiththebest_byor/src/master/}} that obtains the similarity between words by calculating the cosine similarity of their word embeddings, and we modified it to work on Arabic words as well. We get similarity values for each synonym compared to the original word. The word with the highest similarity is the one used for replacement. If a word is selected to be replaced by synonyms multiple times in the same sentence, then each time, the synonym is selected with a lower similarity than the previous one to have new augmented words. 
For RI, a random word is selected from the sentence to get its synonym using the synonym extraction method. A random synonym is selected and placed randomly in a generated random index in the original sentence from the list of synonyms.

\subsection{Back-Translation}
The third technique implemented was based on back-translation to paraphrase a text while retaining the meaning. This was done by translating a text to a different language and then translating it back to the original one. This technique mostly preserves the same semantics of the sentences but generates a different syntax. We implemented three models to apply BT to our CS data. The first model $BT_{T2T}$ used the deep learning library of tensor2tensor~\cite{vaswani2018tensor2tensor}. We used the publicly available data-set for Arabic-French of the United Nation~\cite{ziemski2016united} of size 3M parallel sentence for training our BT model. In the second model $BT_{GT}$, we used Google Translate, and the intermediate language was French.

\begin{figure}[!ht]
    \centering
 \includegraphics[scale=0.65]{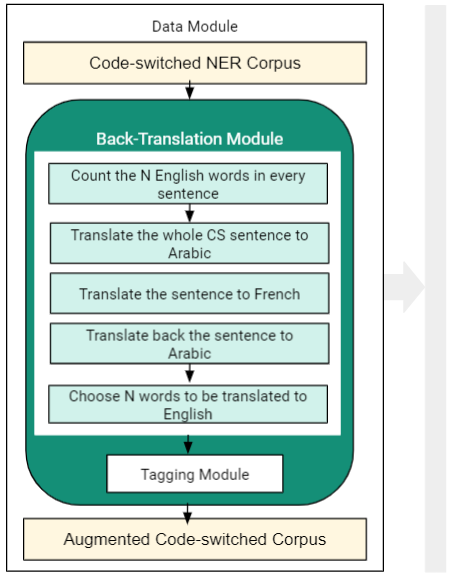}
    \caption{Architecture of the Back-Translation Technique}%
    \label{fig:ArchitectureBT}%
\end{figure}

The same process was applied in both models. 
As shown in Figure \ref{fig:ArchitectureBT}, the models start by translating every English word in our CS data to Arabic. Then, the Arabic sentences are translated into French, and then the French sentences are translated back into Arabic. To preserve the CS behavior in the data, the same number of English words originally existing in the source sentence is maintained in the augmented sentence. The words selected for the translation follow a set of linguistic constraints. The constraints are a set of Arabic words preceding the CS points. Afterwards, we used this new sentence as an augmented version of the original text. 
The third model, $BT_{GT}$+$2L$ used Google Translate as well but using more than one intermediate language to generate more variations from the sentences. All the steps were similar to the previous models. However, after translating the English words of the original sentence to Arabic, we translate the Arabic sentence to French, French to German, and from German back to Arabic. 

The third model, $BT_{GT}$+$2L$ is using Google Translate as well but using more than one intermediate languages to generate more variations from the sentences. All the steps were similar to the previous models. However, after translating the English words of the original sentence to Arabic, we translated the Arabic sentence to French, French to German, and from German back to Arabic.

\section{Experiments \& Results}
We used an extrinsic evaluation, which assesses the system output based on its effect on an external task. The quality of the generated data can be judged based on the improvement in a specific task resulting from combining the new data with the original one \cite{hailu2020intrinsic}. In our case, this task was the NER on CS data. The only available labeled corpus for this task was the one we created and presented in Chpater 5. 
We used the same deep learning NER model presented in Chapter 5 to test its performance with and without data augmentation techniques. This model was based on BiLSTM and CRF along with the pre-trained word embedding layer. We used the same test data they used to compare the performance of their model. Their model trained on the AR-EN CS data without applying data augmentation achieved an F1-score equal to 77.69\%.


\subsection{Modified Easy Data Augmentation (EDA)} 
In the first set of evaluations, we trained the model with the data after applying our modified EDA technique. We investigated different setups for the parameters and operations based on recommendations stated in \cite{wei2019eda}. We applied $EDA_{SR}$ and $EDA_{SR,RI}$ only as the other EDA operations would not guarantee a different sentence with new entities.
\begin{figure}[!ht]
    \centering
  \includegraphics[width=9.5cm]{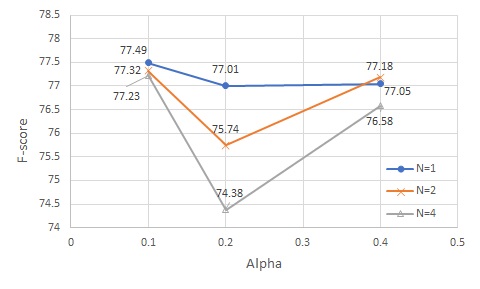} 
    \caption{Results of applying different set of EDA operations with different parameter set-ups}%
        \label{fig:EDA1}%
\end{figure} 
Figure \ref{fig:EDA1} illustrates the usage of SR operation $EDA_{SR}$ with values of $\alpha$ in (0.1,0.2,0.4) and values of number of augments $N$ in (1,2,4). All the models using the SR operation only got results lower than the initial model. This could result due to losing the meaning of the sentence by replacing the words with their synonyms as the SR process does not consider the context. 
\begin{figure}[!ht]
    \centering
    \includegraphics[width=9.5cm]{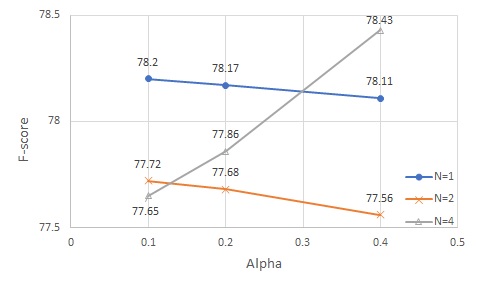} 
    \caption{Results of applying different set of EDA operations with different parameter set-ups}%
     \label{fig:EDA2}%
\end{figure}  

 Figure \ref{fig:EDA2} presents the results of using the SR and RI operations $EDA_{SR,RI}$ with the same range of values for the $\alpha$ and $N$. It can be noticed that the majority of the models combining both operations achieved higher results than $EDA_{SR}$. The highest model of the EDA is $EDA\_(SR,RI)_{\alpha=0.4,N=4}$, which achieved an F1-score of 78.43\%, an increase of 0.74\%. This combination resulted in four variations for each sentence, two of them after applying the SR operation and the other two after applying the RI operation.
 As explained before, the SR operation will replace some of the existing words with their synonym. The RI operation adds to the sentence new words similar to existing ones but does not change the original words, which could be why combining both achieved the highest results.     

\subsection{Word Embedding Substitution \& Back-Translation}

Table \ref{table:Results} gives an overview of the results after applying the set of data augmentation techniques to the data. We can observe that none of the three classical word embedding substitution techniques managed to improve the performance of the model since these techniques replace the words without considering their context. However, the technique of $KERMIT_{sub}$ got the highest result among the word embedding substitution techniques with an improvement of 0.65\%. $BERT_{sub}$ and $KERMIT_{sub}$ results have been restricted by two causes. First, the performance of the model is affected by the accuracy of the NER taggers as NER taggers choose from the predictions the ones with the same entity type. Furthermore, any sub-word prediction is ignored by the taggers. 
We applied the BT techniques alone on the data. They did not enhance performance. This is due to having paraphrased sentences only. Besides, $BT_{T2T}$ got the lowest results, as the data used to train the translation model is not diverse enough, and its domain does not match well the data of the NER corpus. A high chance of errors exists in the tagging process as the taggers miss the context of the words due to the CS behavior.
One possible reason for using the $Full\_WE_{sub}$ is losing the semantics of the sentences, as similar words replace all word occurrences without considering their context. On the other hand, using $Analogies\_WE_{sub}$ alone did the opposite and changed very few words/entities in the sentences, causing the augmented sentences to have significant similarity to the original ones.
\begin{table}

\begin{center}
\caption{Comparison between the NER model with and without applying the different data augmentation techniques (\%)}
\begin{tabular}{l|r|r|r}
\hline
\bf Data Augmentation Techniques  & \bf Precision & \bf Recall  & \bf F1-score \\
\hline
Original Data Without DA &   79.15      &  76.28       & 77.69  \\ 
$Full\_WE_{sub,n=1}$       & 78.47  & 75.14 & 76.77  \\ 
$Full\_WE_{sub,n=5}$      & 80.24   & 74.43 & 77.22  \\ 
$Analogies\_WE_{sub}$   & 78.13  & 73.81 & 75.91    \\ 
$BERT_{sub}$ &  79.81 & 76.63 & 77.64 \\
$KERMIT_{sub}$ & 80.96 & 77.24 & 78.34 \\
\hline 
$BT_{T2T}$                    & 52.35 & 46.78 & 49.41\\
$BT_{T2T}$+$Analogies\_WE_{sub}$       & 75.37   & 69.67 & 72.41  \\ 
$BT_{GT}$              &  81.29  &  74.29   & 77.63         \\ 
$BT_{GT}$ + $Analogies\_WE_{sub}$      & \textbf{81.37}   & 77.14 & \textbf{79.20} \\ 

$BT_{GT}$+$2L$ + $Analogies\_WE_{sub}$     &  81.30 &    76.62   &   78.41  \\ 
\hline 
$EDA\_(SR,RI)_{\alpha=0.4,n=4}$    &     80.50      &   76.47      &    78.43      \\ 
$EDA\_(SR,RI,RD,RS)_{\alpha=0.4,N=4}$ & 75.42 & 78.02  & 76.70\\
$EDA\_(SR,RI,RD,RS)_{\alpha=0.1,N=4}$ & 75.76 & \textbf{78.52} & 77.12 \\
\hline
\end{tabular}

\label{table:Results}
    
\end{center}
\end{table}

Thus, we created a new technique using BT models $BT_{GT}$ and $BT_{T2T}$ along with the word embedding substitution model of $Analogies\_WE_{sub}$ consecutively, to make sure we replace entities with similar ones and introduce new entities to the data. We started by applying the $Analogies\_WE_{sub}$ method in order to be able to replace the entities before they are translated, and then the output sentence is given to the translation module. The performance of the technique $BT_{T2T}$+$Analogies\_WE_{sub}$ is better than the $BT_{T2T}$ alone but is still less than all other models. This is also due to the better quality of the Google Translate output compared to our trained model. 
After applying $BT_{GT}$+$Analogies\_WE_{sub}$ to the training data, the model achieved the highest results with an F1-score 79.20\%, which is an improvement of 1.51\%. This model succeeded in introducing new entities with high accuracy tags as they are mapped from the original ones in semantically correct new sentences.
In addition, we evaluated combining the back-translation technique of two intermediate languages with the word embedding substitution $BT_{GT}$+$2L$ + $Analogies\_WE_{sub}$ as there is no need to try back-translation alone anymore. The model achieved an F1-score equal to 78.42\%, which is an increase of 0.73\%.
\section{Discussion}
As shown in Figure \ref{fig:EDA1}, the highest model of the EDA techniques discussed before was $EDA\_(SR,RI)_{\alpha=0.4,N=4}$, considered to be the second-best data augmentation technique among all techniques in terms of enhancing the F1-score. Thus, we tried to use the same parameters of $\alpha = 0.4$ and $N = 4$ with the full set of operations of the EDA technique. However, the results decreased. This proved our assumptions that not all EDA operations applied to CS data for NER enhance the performance.
The model of $BT_{GT}$ + $Analogies\_WE_{sub}$ achieved the highest results; this is due to the better quality of word embedding substitution compared to the synonym replacement methods. Besides, the usage of BT after introducing new entities in the sentence resulted in having an augmented sentence that is different from the original one. Finally, we trained the model using the two augmented data-sets of the highest two techniques $EDA\_(SR,RI)_{\alpha=0.4,N=4}$ and ($BT_{GT}$ + $Analogies\_WE_{sub}$) combined together. These achieved an F1-score equal to 77.45\%, which means the performance decreased by 0.24\%. We can deduce that enhancing the NER models on CS data is not about having new instances and bigger data-sets only but the quality of the augmented instances in terms of semantics matter.

  \begin{table}[h]

  \begin{center}
    \caption{Total number of entities in each type before and after augmentation using the highest technique and its increase factor}
  \resizebox{\textwidth}{!}{
   
 \begin{tabular}{lrrr|rrr|rrr}
  \hline
  \multicolumn{1}{c}{} & \multicolumn{3}{c|}{Input data }  & \multicolumn{6}{c}{$BT_{GT}$+$Analogies\_WE_{sub}$}    \\ 
  \multicolumn{1}{c}{} & \multicolumn{3}{c}{\textbf{Before}} & \multicolumn{3}{|c}{\textbf{After}}    &\multicolumn{3}{c}{\textbf{Increasing factor}}  \\ 
  \hline
  \textbf{Entities} & \multicolumn{1}{l}{\textbf{English}} & \multicolumn{1}{l}{\textbf{Arabic}} & \multicolumn{1}{l|}{\textbf{Total}} & \multicolumn{1}{l}{\textbf{English}} & \multicolumn{1}{l}{\textbf{Arabic}} & \multicolumn{1}{l|}{\textbf{Total}} & \multicolumn{1}{l}{\textbf{English}} & \multicolumn{1}{l}{\textbf{Arabic}} & \multicolumn{1}{l}{\textbf{Total}}  \\ 
 \hline
  \textbf{LOC} & 2,82 & 563   & 3,38   & 20,62 & 1,95 & 5,29 & 1.18 & 3.47  & 1.56  \\ 
  \textbf{PER}  & 3,70   & 1,52 & 5,23  & 23,20  & 2,69 & 7,46  & 1.29 & 1.76  & 1.43   \\ 
  \textbf{ORG} & 706 & 1,64    & 2,35  & 6,42  & 2,57 & 4,41 & 2.60 & 1.56  & 1.88   \\ 
 \textbf{MISC}  & 1,73     & 1,66  & 3,39    & 11,24   & 2,06    & 4,07    & 1.17  & 1.24   & 1.20\\
  \textbf{Total}  & 8,95  & 5,39    & 14,34   & 61,48 & 9,27   & 21,23   & 1.34     & 1.72  & 1.48 \\
  \hline
 \end{tabular}

 }
    \label{table:AugmentedBT}
   \end{center}

  \end{table}

 \begin{table}

  \begin{center}
   \caption{Total number of entities in each type before and after augmentation using the second highest techniques and its increase factor}
  \resizebox{\textwidth}{!}{
  \begin{tabular}{lrrr|rrr|rrr}
 \hline
  \multicolumn{1}{c}{} & \multicolumn{3}{c|}{Input data}  & \multicolumn{6}{c}{$EDA\_(SR,RI)_{\alpha=0.4,N=4}$}   \\ 
 \multicolumn{1}{c}{} & \multicolumn{3}{c}{\textbf{Before}} & \multicolumn{3}{|c}{\textbf{After}} &   \multicolumn{3}{c}{\textbf{Increasing factor}}  \\ 
  \hline
  \textbf{Entities} & \multicolumn{1}{l}{\textbf{English}} & \multicolumn{1}{l}{\textbf{Arabic}} & \multicolumn{1}{l|}{\textbf{Total}} & \multicolumn{1}{l}{\textbf{English}} & \multicolumn{1}{l}{\textbf{Arabic}} & \multicolumn{1}{l|}{\textbf{Total}} & \multicolumn{1}{l}{\textbf{English}} & \multicolumn{1}{l}{\textbf{Arabic}} & \multicolumn{1}{l}{\textbf{Total}}  \\ 
 \hline
 \textbf{LOC} & 2,82 & 563   & 3,38   & 20,62  & 3,75   & 24,37   & 7.32  & 6.66    & 7.21   \\ 
  \textbf{PER}  & 3,70   & 1,52 & 5,22  & 23,20  & 9,53 & 32,72 & 6.27  & 6.25 & 6.26  \\ 
  \textbf{ORG} & 706 & 1,64    & 2,35  & 6,42   & 10,39   & 16,82  & 9.10    & 6.33  & 7.16   \\ 
 \textbf{MISC}  & 1,73     & 1,66  & 3,39    & 11,24  & 10,49    & 21,73    & 6.514   & 6.32   & 6.42     \\ 
  \textbf{Total}  & 8,95  & 5,39    & 14,34   & 61,48   & 34,16    & 95,63  & 6.87  & 6.34 & 6.67  \\
  \hline
  \end{tabular}}
     \label{table:AugmentedEDA}
   \end{center}

 \end{table}

Table \ref{table:AugmentedBT} and Table \ref{table:AugmentedEDA} illustrate the statistics of the training data before and after applying the highest two augmentation methods along with their  increase factor $y$. The increase factor is calculated based on the formula \(x_{i}' = y * x_{i}, i \in E\), where $y$ is the increase factor by which the number of entities increased after augmentation, $E$ is the set of all named entity types, and \(x_{i}\) is the number of entities for entity type $i$. The size of the original input/train data presented in Chapter 5 before augmentation was equal to 5,306 sentences. The testing data was not augmented for comparison while running the NER model with the new augmented training data. After applying the highest technique of $BT_{GT}$+$Analogies\_WE_{sub}$, the size increased to 10,612 sentences. The total number of entities in the initial data was 14,34; after augmentation using $BT_{GT}$+$Analogies\_WE_{sub}$, it increased by a factor $y$ equal to 1.48. 
Also, after we applied the $EDA\_(SR,RI)_{\alpha=0.4,N=4}$ it increased to 95,63. Again, this highlights the fact that enhancing the performance does not depend on just adding more entities only.
The technique of $BT_{GT}$+$Analogies\_WE_{sub}$ added fewer entities than the $EDA\_(SR,RI)_{\alpha=0.4,N=4}$ but it improved the performance further from 77.69\% to 79.20\%. 

To check the quality of the auto-generated data of the two best techniques, we took a sample of 100 random sentences for each one and manually checked their quality, to see if they are semantically correct or not. Regarding the data generated after applying the $EDA\_(SR,RI)_{\alpha=0.4,N=4}$ model, we found out that 60\% of the sentences are of good quality. This could be because replacing only words with their synonyms or inserting new synonyms will lead to losing the meaning of the sentence. However, the result of checking the data generated after applying the $BT_{GT}$+$Analogies\_WE_{sub}$ model is better resulted in having 70\% of the sentences meaningful and understandable. This could go back to the better quality of the word embedding substitution rather than the synonym replacement method and back-translation usage that changes the sentence but keeps its meaning.    

The following example shows the original tagged sentence and its resulting artificial instance after applying the best back-translation technique along with word embedding substitution.
\newline
\newline
 \textbf{Original sentence:}
\begin{flushright}
  Cairo
 \< في >
 Chahine
 \<
 خلال عرض فيلم المخرج يوسف >

 LOC \hspace{0.08cm}
 O \hspace{0.3cm}
 PER \hspace{0.6cm}
 PER  \hspace{0.6cm}
  O \hspace{0.4cm}
 O \hspace{0.5cm}
 O \hspace{0.5cm}
 O \hspace{0.6cm}
\<  نحاول ان نقول لجمهور مصر وللسينمائيين
نحن
موجودون
>

  O \hspace{0.8cm}
 O \hspace{0.9cm}
 O \hspace{0.6cm}
 LOC \hspace{0.40cm}
O \hspace{0.4cm}
 O \hspace{0.35cm}
 O \hspace{0.3cm}
 O 

 \end{flushright}
 Through displaying the movie of the director Youssef Chahine in Cairo, we try to say to the audience of Egypt and the cinematologists: We exist.
\newline
 \textbf{Augmented sentence:} 
 \begin{flushright}

\<مدغشقر نحاول أن نقول للجمهور >
in
\<مارتن >
Ahmed
\<خلال إسقاط المخرج>

 O \hspace{0.4cm}
 O \hspace{0.15cm}
 O \hspace{0.25cm}
 O \hspace{0.4cm}
 LOC \hspace{0.15cm}
 O \hspace{0.02cm}
 PER \hspace{0.05cm}
 PER \hspace{0.5cm}
 O \hspace{0.9cm}
 O \hspace{0.9cm}
 O \hspace{0.9cm}
\<و صناعي الأفلام: نحن موجودون>
the Egyptian

 O \hspace{0.4cm}
 O \hspace{0.5cm}
 O \hspace{0.5cm}
 O \hspace{0.5cm}
  O \hspace{0.1cm}
MISC \hspace{0.1cm}





\end{flushright}

 Through projection of the director Ahmed Martin in Madagascar we try to say to the Egyptian audience and movie producers: We exist.

\section{Summary}

We proposed two techniques that complement the NER taggers on CS data and enhance their performance. The first one was training existing contextual embedding models using our created corpora for this task. In addition, we proposed a new contextual model called KERMIT. The second one was applying data augmentation techniques, word embedding substitution, modified approach of EDA, and back-translation on CS data. Using each enhancement technique alone increased the performance of the NER tagger.
\begin{figure}[!ht]
\centering
\includegraphics[width=13cm, keepaspectratio]{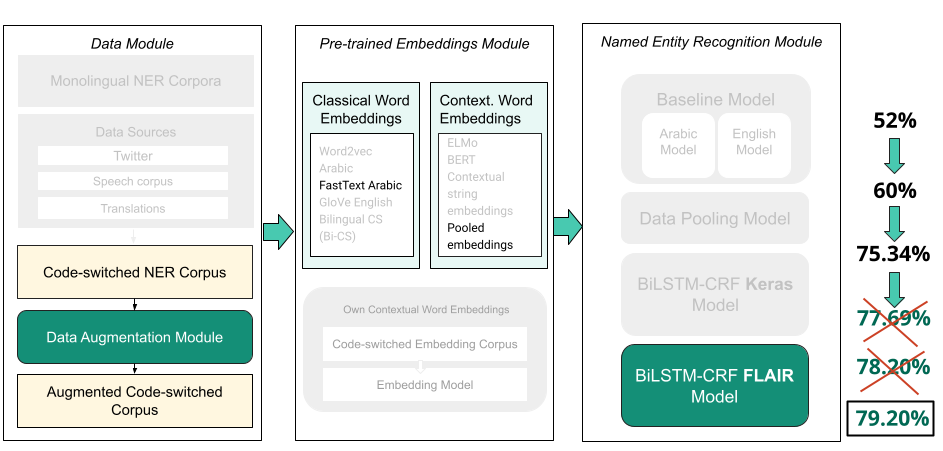}\\[2mm]
\caption{Results of Second Enhancement Technique}
\label{Figure:fg}
\end{figure}

As shown in Figure \ref{Figure:fg}, our trained Contextual string embedding model with Arabic FastText embedding achieved the best results and enhanced our NER model by 0.51\%. Also, when we used KERMIT with the HuggingFace NER tagger achieved higher results by 1.2\% than the one presented by our first NER approach. It achieved an F1-score of 79.4\%, setting new state-of-the-art results for the NER task on AR-EN CS data. 





%% file: LID.tex
\chapter{Language Identification of Intra-Word CS for Arabic-English}\label{chap:LID}
\chaptermark{LID of Intra-Word CS for AR-EN}
One of the essential pre-processing steps in the NLP field is LID. It is the task of determining the language type of a text. The majority of the work that has been conducted in this area was done for the document level. CS has moved the focus to the word level. Nevertheless, very few researchers targeted subword-level language identification, segmenting mixed words and tagging each part with its corresponding language ID. In most of the LID systems, the CS intra-words are labeled as mixed words and, they are not analyzed further, and their internal information is lost. 

We focus on this intra-word information as well as word-level language identification. 
One of the main challenges for LID systems applied to Arabic and AR-EN CS texts is having Arabizi (referring to Arabic written using the Latin/Roman script) and Engari (referring to English written using Arabic script) \cite{Al-Badrashiny2016} tokens. They make it harder to recognize both languages as their usual script/Unicode used is different. This work is a preliminary one to identify named entities in mixed words as this current LID identifies if a token is an Arabic or English entity.

\begin{figure}[!ht]
\centering
\includegraphics[width=13cm,keepaspectratio]{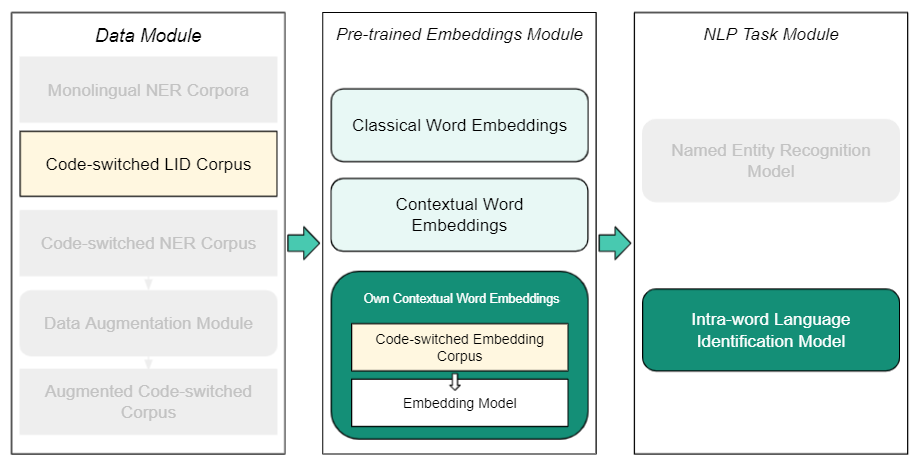}\\[2mm]
\caption{Proposed Pipeline for LID Task }
\label{fig:pLID}
\end{figure}

 In this chapter,
we first present how we created and annotated AR-EN corpus for the CS intra-word LID task as shown in Figure \ref{fig:pLID}. Then, we describe our three implemented models for segmenting mixed words and tagging each part with its corresponding language ID: Na\"{i}ve Bayes, Character BiLSTM, and Segmental Recurrent Neural Networks (SegRNN).
\section{Data Collection and Annotation }
We collected our own Arabic-English CS data and annotated the tokens with their corresponding language tag. It is the first annotated AR-EN data set for intra-word code-switching Language Identification task. In this section, we illustrate how the data was collected from three different social media platforms: Twitter, Facebook, and WhatsApp, as their users frequently tend to code-switch. Before starting the data collection process, an Ethics proposal was submitted and approved by our faculty committee. We collected public data from Twitter and Facebook. Regarding WhatsApp data, we gathered
consent forms from users before acquiring their data from a particular group. The total number of gathered sentences was 2,507 containing 30,321 annotated tokens. The corpus included Arabic, English, Arabizi, and Engari tokens. After collecting the data, we manually segmented and annotated it with the corresponding language tags. 
Then, we analyzed the data to know the statistic of how the tags are distributed over the data. We also observed the common patterns users tend to use while code-switching in the same word.
\subsection{Data Collection}
We implemented three techniques to collect our data and build the corpus. The first technique was gathering data from Twitter using Tweepy API\footnote{\url{www.tweepy.org}}, and we collected 8,589 tokens. The second one was to collect data from Facebook\footnote{\url{www.facebook.com}}, and we got 8,692 tokens. The third one was to collect WhatsApp\footnote{\url{www.whatsapp.com}} data, and it was the most successful as we got 13,040 tokens.
 
\subsubsection{Twitter Data}

We first got a sample of random tweets and observed the words most used in a code-switching context, like \textit{Elcode} (The code) and \textit{Elgam3a} (The university). Then, these words were used as keywords to search for other tweets. In addition, we crawled tweets with a geographic location equal to \textit{Cairo}, and we extracted again a new set of words that was used again in the search queries. The total number of crawled tweets was 1,859 that was filtered by removing the duplicates and retweets to reach 545 tweets containing CS data. Another round of filtration was done on the remaining tweets to remove the hashtags, URLs, and usernames. Then the collected data was tokenized, and we got 8,589 tokens.

\subsubsection{Facebook Data}

To get more data, we also crawled public posts from Facebook pages. We collected 32 posts that belong to different topics like technology, food, travel, and jobs. We filtered the data to remove the hashtags, URLs, and usernames. These were composed of 337 sentences with a total of 8,692 tokens.
\subsubsection{WhatsApp Data}
To achieve our purpose and collect a significant amount of CS data, we investigated the usage of a third platform, WhatsApp. As it is one of the most used social media applications where users communicate through it daily, they tend to use informal languages. We selected a WhatsApp group of students and gave them a consent form to sign to get their permission to collect their data. We collected 1,625 sentences with 13,040 tokens, which is the most significant amount of the collected data. 
\subsection{Tag Description}
The second step after collecting the data was to annotate the tokens of the collected corpus manually. 
We had one expert annotator and one expert reviewer. They are both native
speakers of Arabic, and their second language is English. Besides, one
researcher helped to resolve conflicts arising between the annotator and the reviewer. 
We followed a similar annotation schema like \cite{Cetino1997}.
Each token/word in the collected data was annotated with one of 8 classes, which are EN, AR, LANG3, MIXED, AMBIG, NE.AR, NE.EN and OTHER. The tags \textit{EN} and \textit{AR} represented tokens written in English and Arabic languages, respectively. \textit{LANG3} tag corresponded to tokens in other languages. \textit{MIXED} tokens were for the tokens containing more than one language (intra-word CS). We also segmented the mixed tags and assigned a tag to each segment that corresponds to its language. \textit{AMBIG} refers to the tokens which could not be assigned to a language based on their context. Concerning the named entity tags, they were represented by \textit{NE} tag and the language ID of the token either \textit{AR} or \textit{EN}. The last tag \textit{OTHER} was used to tag tokens that do not represent actual words like punctuation marks, numbers, emoticons, and symbols. The following two examples represent two sentences from our data, their corresponding tags, and translations:

\begin{center}
\begin{exe}
\ex \gll I played {\textbf{el}game} elgedida ala {laptop\textbf{y}} \\
    EN EN {AR-EN} AR AR {EN-AR} \\
\trans’I played the new game on my laptop’
\end{exe}
\begin{exe}
\ex \gll Danke ana rayeh Germany \\
LANG3 AR AR NE.EN \\
\trans’Thank you I am going to Germany’
\end{exe}
\end{center}

\subsection{Data Statistics}
 The data contained Arabic, English, Arabizi, and Engari tokens. The Arabizi and Engari tokens are kept in the same format as collected. The following word is an example of the Arabizi token \textit{gedida} which means in English \textit{new}, and as an example of the Engari token, the word \textit{\<لابتوب>} which means \textit{laptop}. 

\begin{table}[ht!]

    \begin{center}
          \caption{Twitter Data}  
    \begin{tabular}{l|r|r|r|r}\hline  
\bf  Tag  & \bf Tokens  & \bf \% & \bf Unique & \bf  Unique \\
    & & & \bf Tokens & \bf \% \\\hline
    AR             &      5206& 60.61     &      2594&65.89     \\
    EN             &      2321& 27.02     &       902&22.91    \\
    OTHER          &       748& 08.71      &       236&05.99   \\
    NE.AR          &        80& 0.93      &        64&01.63      \\
    NE.EN          &        48& 0.56      &        39&0.99      \\
    AMBIG          &        18& 0.21      &        14&0.36      \\
    LANG3          &         1& 0.01      &         1&0.03      \\
    MIXED          &       167& 01.94      &        87&02.21       \\
   
    \quad AR,EN          &       158&94.61     &        78&89.66     \\
        \quad EN,AR          &         3&1.80      &         3&3.45      \\
    \quad AR,EN,AR       &         3&1.80      &         3&3.45      \\
    \quad AR,OTHER,EN    &         3&1.80      &         3&3.45      \\

     \hline
    \end{tabular}
        \label{table:twitter_data}
   \end{center}

\end{table}

Starting with the data of Twitter containing 545 sentences and 8,589 tokens, we noticed these belong to different tags as shown in Table \ref{table:twitter_data}. The total number of mixed tags was 167 tags consisting of 87 unique mixed words. The tag with the highest percentage was the AR having 60.61\% from all Twitter data tags. We can note that the most used mixed pattern on Twitter was AR,EN, which means adding an Arabic prefix to an English word. 
 Concerning the data collected from Facebook containing 337 sentences and 8,692 tokens, we noted that these belong to different tags as shown in Table \ref{table:facebook_data}. The total number of mixed tags was 173 tags consisting of 124 unique mixed words. The tag with the most number of tokens was the AR tag, it contains 6,726 tokens. The pattern with the highest number of words was the same as the Twitter data AR,EN.

\begin{table}[ht!]
 
  \begin{center}
 \caption{Facebook Data}
    \begin{tabular} {l|r|r|r|r}
    \hline  \bf Tag  & \bf Tokens  & \bf \% & \bf Unique &\ \bf Unique \\
    & & & \bf Tokens & \bf \% \\\hline
    AR             &      6726&77.38     &      2957&80.46     \\
    EN             &       386&04.44      &       275&07.48      \\
    OTHER          &      1127&12.97     &       127&03.46    \\
    NE.AR          &       128&01.47      &       100&02.72      \\
    NE.EN          &       146&01.68      &        90&02.45      \\
    AMBIG          &         5&0.06      &         1&0.03      \\
    LANG3          &         1&0.01      &         1&0.03      \\
    MIXED          &       173&01.99      &       124&03.37      \\
  
    \quad AR,EN          &       132&76.30     &       100&80.65     \\
    \quad EN,AR          &        22&12.72     &        11&08.87    \\
    \quad AR,EN,AR       &        19&10.98     &        13&10.48     \\
  
     \hline
    \end{tabular}
     \label{table:facebook_data}
      \end{center}
   
\end{table}

The last portion of the data from WhatsApp was the biggest containing 1,625 sentences and 13,040 tokens, and belonging to different tags as shown in Table \ref{table:whatsapp_data}. The total number of mixed words was 437 consisting of 377 unique words. They formed 56.24\% of the total mixed tags from the whole corpus, which could show that people code-switch more while chatting on WhatsApp. The pattern with the most words was the same as the previous ones AR,EN. 


\begin{table}[!ht]

 \begin{center}
    \caption{Whatsapp Data} 
    \begin{tabular}{l|r|r|r|r}
\hline \bf Tag  & \bf Tokens  & \bf \% &\bf Unique &\bf Unique \\
    & & & \bf Tokens &\bf \% \\\hline
    AR             &      7075&54.26     &      2258&51.80     \\
    EN             &      3403&26.10     &      1180&27.07     \\
    OTHER          &      1919&14.72     &       426&9.77    \\
    NE.AR          &       109&0.84      &        65&1.49      \\
    NE.EN          &        56&0.43      &        31&0.71      \\
    AMBIG          &        32&0.25      &        15&0.34      \\
    LANG3          &         9&0.07      &         7&0.16      \\
    MIXED          &       437&3.35      &       377&8.65      \\

    \quad AR,EN          &       425&97.25     &       366&97.08  \\
  
    \quad EN,AR          &         4&0.92      &         3&0.80   \\
    \quad AR,EN,AR       &         2&0.46      &         2&0.53 \\
    
    \quad AR,OTHER,EN,AR &         1&0.23      &         1&0.27 \\
     \hline
    \end{tabular}
        \label{table:whatsapp_data}
    \end{center}

\end{table}

\begin{table}[!ht]
  
  \begin{center}
  \caption{Total Data statistics: The number of tokens per each tag }
\begin{tabular}{l|r|r|r|r}
    \hline \bf Tag  & \bf Tokens  & \bf \% & \bf Unique &\bf Unique \\
    & & & \bf Tokens & \bf \% \\\hline
    AR             &     19007 &62.69     &      7013 &66.11     \\
    EN             &      6110 &20.15     &      1948 &18.36     \\
    OTHER          &      3794 &12.51     &        657 &6.19      \\
    NE.AR          &       317 &1.05      &       219 &2.06      \\
    NE.EN          &       250 &0.82      &       154 &1.45      \\
    AMBIG          &        55 &0.18      &        26 &0.25      \\
    LANG3          &        11 &0.04      &         9 &0.08      \\
    MIXED          &       777 &2.56      &       582 &5.49      \\

    \quad AR-EN          &       715 & 92.02     &       539 & 92.61  \\   
    \quad EN-AR          &        29 & 3.73      &        17 & 2.92   \\
    \quad AR-EN-AR       &        24 & 3.09      &        17 & 2.92     \\ 
    \quad AR-OTHER-EN    &         8 & 1.03      &         8 & 1.37      \\
    \hline
    
    \end{tabular}
     
    \label{table:cs_corpus}
    \end{center}

\end{table}
Table \ref{table:cs_corpus} shows the details of the distribution of the number of tokens over the tags along with their unique number. The total number of sentences in the final corpus was 2,507 containing 30,321 tokens, or 12.09 average token per sentence. The \textit{AR} tag contained the highest number of tokens, equal to 19,007 tokens. The total number of \textit{MIXED} tags was 777 (2.56\%) tokens, consisting of 582 (5.49\%) unique tokens. The \textit{LANG3} tag contained the minimum number of tokens, equal to 11 tokens.
Also, the Table illustrates the numbers of occurrences for each pattern of tags assigned for segmented mixed words. The majority of the patterns started with Arabic tokens. Moreover, the most repeated pattern was AR-EN amounting 715 of the total number of MIXED tokens.  

\subsection{Observations}
 We analyzed our CS corpus to extract some frequently used patterns to switch between the Arabic and English languages in the same word. As stated before in Chapter 2, we could find different lexical variations using different patterns in the Arabic language. The structure of the word could contain one or more prefixes, a stem, and one or more suffixes \cite{shaalan2014survey}. We noticed that most of the mixed words consisted mainly of English words with Arabic prefix or suffix or both. For instance, \textit{elgame} is composed of the word \textit{game} along with the prefix \textit{el} (the), \textit{matchat} consisting of the word \textit{match} with the suffix \textit{at} (to make it plural) and the last example is \textit{elmatchat} that contains the word \textit{match} with both a prefix \textit{el} and the suffix (at). 

Table \ref{table:arabic_examples} shows the most used prefixes and suffixes in the mixed words. The first set of prefixes all mean \textit{the} in English, and occurred 537 times out of the 777 mixed tokens. That means Arabic CS speakers tend to switch to English (or a different language) for content words and retain function words in native/base language. Native speakers of Arabic also used the definite articles with prepositions like \textit{fel} (in the) 48 times. In addition, they used it as a suffix \textit{\<ات>} which refers to the feminine plural in 18 tokens.

\begin{table}

\begin{center}
\caption{The most used prefixes/suffixes along with their English translation, the type of the token (AR: Arabic) and (ARB: Arabizi) and their count of occurrences in our data}
\begin{tabular}{l|l|l|r}
    \hline  \bf Token &  \bf Token Type &  \bf  Type   &  \bf Count\\ \hline
    \<ال> (The) & AR &prefix    &251 \\
    el & ARB & & 187\\
    l & ARB & & 89\\
    \<الـ> & AR & &10 \\
    \hline
       fl ( In the) & ARB &prefix    &30 \\
    fel & ARB & & 18\\
    \hline
    lel (To the)& ARB &prefix    &18     \\
    \<لل> & AR    & & 11\\
    \hline
    \<ات> (feminine plural)  & AR    &suffix    &18 \\
    \hline
\end{tabular}
\label{table:arabic_examples}
\end{center}

\end{table}
\section{Baseline Models}
We used several architectures to implement different models for solving the segmentation and language identification tasks for our CS data. The first baseline was implemented using Na\"{i}ve Bayes algorithm. The second one was created using the Character BiLSTM architecture. 
\subsection{Na\"{i}ve Bayes Baseline Model}
In order to implement our baseline model we used Na\"{i}ve Bayes algorithm. Our model started by taking a word/token as input and giving the tag with its highest probability as output. This process consists of two modules. The first one involved converting the token into Term Frequency–Inverse Document Frequency (TF-IDF) feature, which is a numerical statistic reflecting the importance of a token in our data based on analyzing the N-gram characters for N from 1 to 3. The second module was the multinomial Na\"{i}ve Bayes classifier, one of the two classic Na\"{i}ve Bayes variants used in text classification, with the data represented as word vector count. This module takes the TF-IDF features and computes a probability for each tag type. The output of the model is the tag with the highest probability.
\begin{figure}[ht!]
  \centerline{\includegraphics[scale=0.7]{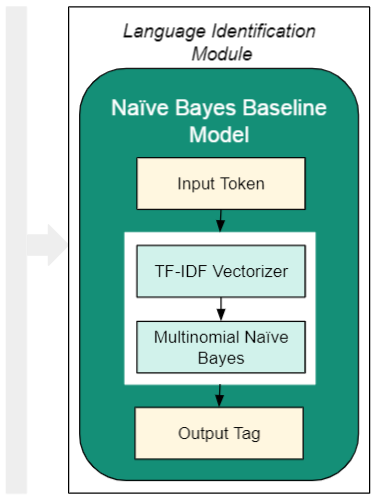}}
  \caption{Na\"{i}ve Bayes Model Architecture}
  \label{fig:NBArchitecture}
\end{figure}

\subsection{Character Bidirectional LSTM Baseline Models}
We used the same architecture proposed in \cite{Mager2019} of Character BiLSTM to create our second set of baseline models. The main model was composed of three layers: Character Embedding, BiLSTM, and time distributed layer. The input was a sequence of character ids passed to the embedding layer to get for each id a vector. The vectors were given to the BiLSTM layer accessing the preceding and succeeding contexts. Thus, the model will capture the long-distance relations in the sequence to predict the labels. Then the output was passed to the time distributed layer to have a tag for each character and wrap the output to one tag sequence. The time distributed layer applied a temporary BiLSTM layer to every character of the input.

We edited the architecture of this model and created one by adding the N-gram embedding feature to the character embedding. This feature allowed the model to take into consideration the information of each sequence of N characters. These captured more context around each character. For example, if the input token was \textit{elgym}, meaning in English \textit{the gym}, for N=1, which is the same as if we did not add the N-gram feature, the input will be each character alone \textit{e l g y m}. For N=2, the input will be \textit{el lg gy ym m}. For N=3, the input will be \textit{el elg lgy gym ym}. 
\begin{figure}[ht!]
    \centerline{\includegraphics[scale=0.7]{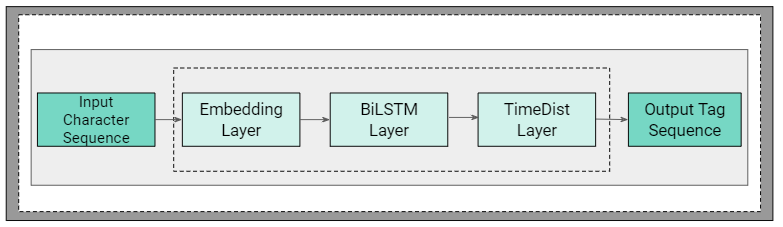}}
  \caption{Character Bidirectional LSTM Model Architecture}
  \label{fig:CharBiLSTM}
\end{figure}
\section{Segmental Recurrent Neural Networks Models}
The capabilities of classical automatic language identification techniques are still limited for detecting the language of a text segment in a code-switched content \cite{das2014identifying}. Thus, we based our main model on \cite{Mager2019} using the Segmental Recurrent Neural Networks \cite{Kong2016} and updated it to adapt to our needs for identifying the language of the word tokens in Arabic-English data. We investigated the usage of different types of embeddings along this main architecture. The model created for a joint probability distribution over its possible segmentation and labels for each segment for the input. 
\subsection{Data Pre-processing}
\begin{figure}[ht!]
  \centering
  \includegraphics[width=0.5\linewidth]{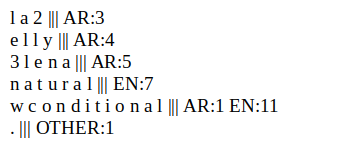}
  \caption{Data format for SegRNN model}
  \label{fig:DataFormat}
\end{figure}

The SegRNN model requires the data to be in a special format as shown in Figure \ref{fig:DataFormat}. This sentence is from our corpus \textit{la2 elly 3alena natural wconditional} (no what we have is natural and conditional) and its tags. For example, the first word \textit{la2} (no) has the tag \textit{AR} and its segmentation is 3 letters. However, \textit{wconditional} (and conditional) is composed of two sub-words and it is tagged with \textit{AR:1} to refer to the first letter as Arabic and \textit{EN:11} to refer to the other 11 letters as English. The words and their tags are separated with the symbols \textbf{$|||$}.

\subsection{Main SegRNN Model}

\begin{figure}[!ht]
  \centerline{\includegraphics[scale=0.7]{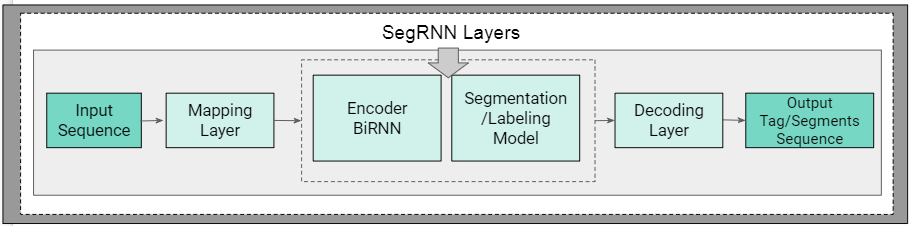}}
  \caption{Main Model Architecture}
  \label{fig:SegRNN}
\end{figure}

As shown in Figure \ref{fig:SegRNN}, the main model takes the sequence of characters as input. Then, it is passed to the mapping layer, which maps each character to a numeric value from a dictionary. Then it resizes the input dimension to 64. Afterwards, the output enters the SegRNN layers consisting of two sub-layers. The first layer is the BiRNN layer, containing a BiLSTM encoder to encode each embedding of the tokens in both directions to preserve the context. The second layer is the Segmentation/Labelling, responsible for the segmentation and tagging tasks of the input passed to it from the previous layer with a dimension equal to 16. Then, the tags with 32 dimensions and length four are given to the final layer of the decoding to decode it to the final output format. The Adam adaptive learning rate method was used for the training.

\subsection{SegRNN with Word Embeddings Models}
We added an extra layer of word embedding to our SegRNN model. We investigated the usage of the following types of embeddings. They generated different sizes of embedding vectors. Thus, we had to adjust the input size of the SegRNN model each time to fit it.\\
\textbf{FastText}: The first type of embedding we investigated was the FastText pre-trained embedding. We combined the Arabic and English FastText pre-trained models. The size of the vectors was 950.\\
\textbf{Pooled Flair}: To tackle the problem of having Arabizi and Engari in our data, which is not common in the available pre-trained embeddings, we implemented our language model and pre-trained embedding using FLAIR framework. Besides, by having our embeddings, we ensured that the intra-word CS phenomenon is included in the embedding. 
The size of our embedding vectors was 4096.\\
\textbf{MUSE}: The third type of embedding we tried was the multilingual universal sentence encoder (MUSE) \cite{yang2019multilingual} which maps text written in different languages having the same meaning, to nearby embedding space representations. It supports 16 languages, including Arabic and English. The embedding vector size was 512.
\section{Evaluation \& Results}

The data we used to evaluate our models was composed of 23,428 tokens for training and 6,893 tokens for testing. 
The train:test ratio of the number of mixed words in our data was approximately 3:1 (588:189). We used the same testing data in all models; it was based on the tokens. The performance of our models was evaluated using the F1-score (F1) performance measure, the accuracy of the tagging (Acc.), and character tagging accuracy (Char Acc.) calculated by assigning a language ID to each character and calculating the ratio of correct tags overall characters. 
The intuition behind using Char Acc. was to have another way to evaluate the model. While calculating the F1-score, if the tag is not entirely correct/typical to the expected, it will not be counted as correct. However, in some cases, parts of the tag could be correct and others wrong. For instance, if the word \textit{laptopy} is given the tag \textit{EN:5 AR:2} instead of \textit{EN:6 AR1}, it will be wrong. Thus, calculating the Char Acc. will show that the model miss-tagged only one letter.

We first stated the results of the evaluations applied to our main data-set. Besides, we created another version of the
named entities labeled coarse-grained by combining all the named Arabic and English entities falling under the NE tag without distinguishing the language. This was done to experiment with the effect on the results.

\subsection{Main Data-set}

Table \ref{tab:fscoreBasline} and Table \ref{tab:fscore} show all test results of tagging and segmentation (Seg.) for the entire dataset and the mixed words in our dataset. Concerning the baseline of the Na\"{i}ve Bayes model, no segmentation was applied. As shown in Table \ref{tab:fscoreBasline}, the Na\"{i}ve Bayes model failed to recognize the mixed tokens.
  \begin{table}[!ht]
\begin{center}
  \caption{Evaluation results of the baseline models}
    \begin{tabular}{l|r|rrrr}
    \hline
& &\multicolumn{4}{c}{\textbf{Character BiLSTM}} \\  \cline{3-6}
\bf    Evaluation    & \textbf{Na\"{i}ve } & \bf Main & \multicolumn{3}{c}{\bf + N-gram}\\
\bf Metric & \bf Bayes & \bf  Model&   \bf Bi & \bf Tri & \bf Bi \& Tri  \\
    \hline
      Tag F1 & 86.69 & 90.77 & 93.06 & 93.77 & 93.83   \\
        Acc. & 89.15 & 90.63 & 92.79 & 93.53 & 93.62   \\
        Seg. F1 & - & 96.73 & 97.88 & 98.06 & 98.31 \\
        Char Acc. & - & 92.88 & 93.77 & 94.17 & 94.22   \\
        Mixed Tag F1 & 0.0 & 24.94 & 41.29 & 44.37 & 48.59   \\
        Mixed Seg. F1 & - & 26.69 & 38.98 & 43.28 & 48.11  \\
        Mixed Acc. & - & 29.98 & 42.49 & 44.83 & 48.20   \\
        Mixed Char Acc. & - & 70.88 & 74.65 & 74.87 & 78.02  \\     \hline
\end{tabular}

\label{tab:fscoreBasline} 
\end{center}
\end{table}
Adding an N-gram feature to the main Character BiLSTM model improved the performance compared to the main model by around 3\% for the overall tagging and by the double for the mixed tokens. This improvement is a result of considering the context of the tokens more by the N-gram feature. However, even the highest model of Character BiLSTM with bi-gram and tri-gram combined under-perform some of the SegRNN models.

Several configurations of the SegRNN model were evaluated. It was trained using three different techniques. The first technique was training the model using single tokens as training data, represented in the main model. The second one was the phrase model; it was trained using sentences. The third one was training the model on N-gram words/tokens to consider the context of the word. The N-gram was used to create a sub-sequence of N adjacent words from a given sequence. We trained the model on tri-gram and bi-gram tokens. The tuned SegRNN model represents the model resulted after applying the hyper-parameters tuning for the main model. 

\begin{table}[!ht]

\begin{center}
\caption{Evaluation results of the SegRNN models}
\resizebox{\textwidth}{!}{

    \begin{tabular}{l|r|r|r|rrr|rrr}
    \hline
        & \multicolumn{9}{c}{\textbf{SegRNN}} \\ 
    \cline{2-10}
        \textbf{Evaluation} & \textbf{Main} & \textbf{Tuned} & \textbf{Phrase} & \multicolumn{3}{c|}{\underline{\textbf{N-gram Training Data}}} & \multicolumn{3}{c}{\underline{\textbf{Models With Embedding}}}  \\
    \textbf{Metric}  & \textbf{Model} & \textbf{Model} & \textbf{Model} & \bf Tri & \bf Bi & \bf Bi \& Uni & \bf Flair & \bf FastText & \bf Muse \\   \hline
    
   Tag F1 & 94.57 & \textbf{94.84} & 87.65 & 54.01 & 56.36 & 92.80 & 94.33 & 94.53 & 94.70  \\
        Acc.  & 94.95 & \textbf{95.21} & 85.85 & 60.35 & 42.51 & 92.06 & 94.94 & 95.05 & 95.02  \\
        Seg. F1  & \textbf{99.17} & \textbf{99.17} & 94.90 & 89.65 & 53.78 & 97.41 & 99.11 & 99.16 & 99.12 \\
        Char Acc.  & 94.80 & \textbf{95.05} & 89.66 & 66.35 & 85.80 & 93.85 & 94.63 & 94.73 & 94.80  \\
        Mixed Tag F1   & 76.85 & 81.15 & 15.16 & 0.0 & 2.48 & 38.59 & 74.73 & \textbf{81.45} & 79.12  \\
        Mixed Seg. F1  & 75.27 & \textbf{78.40} & 12.15 & 2.08 & 1.74 & 32.83 & 72.52 & 78.29 & 75.66 \\
        Mixed Acc.  & 74.40 & 74.75 & 16.25 & 0.0 & 3.90 & 36.83 & 72.51 & \textbf{75.0} & 73.87  \\
        Mixed Char Acc. & \textbf{88.77} & 87.42 & 71.17 & 49.66 & 83.09 & 82.83 & 85.60 & 87.14 & 88.46  \\ 
        \hline
        
\end{tabular}}

\label{tab:fscore} 
\end{center}
\end{table}

\begin{figure}[!ht]
  \centerline{\includegraphics[width=2.5in]{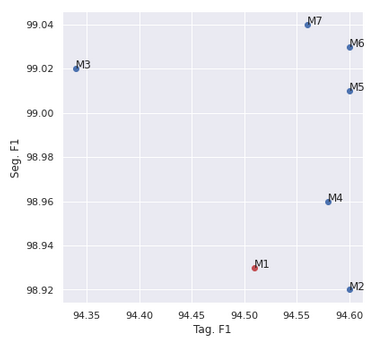}}
  \caption{hyper models result}
  \label{fig:hyperparam}
\end{figure}

As training and testing on tokens gave the best results, we applied the 5-fold cross-validation to tune the hyper-parameters of this model. We trained 41 models and, 6 of them performed better than the initial model based on the F1-score for the segmentation or for the tagging. Figure \ref{fig:hyperparam} shows the tagging and segmentation F1-scores of the 6 models illustrated in blue and the initial model illustrated in red. M6 has the best average F1-score, and M1 is considered as the base model. To achieve the best model, the learning rate changed from 0.0005 to 0.0003 and the maximum token length changed from 1000 to 20 characters. 

As shown in Table \ref{tab:fscore} training on single tokens resulted in higher results compared to the one using sentences. The lower results were achieved by the SegRNN model trained using the tri-gram and bi-gram data, and they almost failed to recognize the mixed tokens. As both types of data did not achieve good results, we combined the bi-gram with the uni-gram tokens. This type of training data performed better; however, still the performance was lower or poorer than the main model. 

The addition of the embedding layer did not enhance the model, due to the fact of not covering the Arabizi and Engari in the available pre-trained embedding models. Moreover, the Pooled Flair embedding model we created had a small size and did not have a significant effect. Adding the embeddings to the main model did not enhance the performance much except tagging the mixed words. The performance differences between the tuned SegRNN and the one with the embeddings were small. The best performance for tagging the entire data is obtained by the tuned SegRNN model equal to 94.84\% and then by the one with MUSE embedding equal to 94.70\%. The SegRNN with FastText model achieved the best performance for tagging the mixed words equal to 81.45\% F1-score, while the tuned SegRNN model and its main model got very high results for segmentation equal to 99.17\%. 

We can conclude that the SegRNN models trained on single tokens tag and segment well intra-word CS data compared to other approaches presented here. Table \ref{table:seg} shows a classification matrix for best model of SegRNN. The model predicted well the monolingual tags (AR, OTHER, EN) and the tokens from different languages. However, it did not perform well in detecting the AMBIG tokens and named entities (NE.AR and NE.EN). For bilingual tags, it achieved better results assigning (AR-EN and AR-EN-AR) compared to (EN-AR) tokens. 
\begin{table}[h]
\begin{center}
  \caption{Classification matrix for best model of SegRNN }
    \begin{tabular}{l|r|r|r}
    \hline
   \textbf{Tag} & \textbf{Precision}  &  \textbf{Recall} & \textbf{F1-Score} \\
        \hline
              AR &  96.30 & 98.60 & 97.44    \\
              EN &  91.69 & 94.74 & 93.19    \\
           OTHER &  98.58 & 95.76 & 97.15    \\
           LANG3 &  100.00 & 50.00 & 66.67    \\
           NE.EN &  69.44 & 37.88 & 49.02    \\
           NE.AR &  70.45 & 31.00 & 43.06    \\
           AMBIG &  33.33 & 20.00 & 25.00    \\
           EN-AR &  50.00 & 25.00 & 33.33    \\
           AR-EN &  93.38 & 81.03 & 86.77    \\
        AR-EN-AR &  100.00 & 75.00 & 85.71    \\
     AR-OTHER-EN &  100.00 & 33.33 & 50.00    \\
        \hline
   \end{tabular}
   \label{table:seg}
    \end{center}
\end{table}
The following are some examples of tokens that the SegRNN model failed to tag: \textit{benefit}, which should be tagged as EN, but the model segmented it into \textit{be} and \textit{nefit} with the corresponding tags AR and EN. In the case where segmentation errors occur, the resulting tag is also wrong/false. Another example, is the word \textit{mlinsertions} (from insertions) which should be tagged and segmented to be \textit{ml} and \textit{insertions} as AR and EN but the model tagged all its characters as EN. In addition, as examples of words that were successfully segmented and tagged by the SegRNN model and not by BiLSTM model are \textit{\<لكويزات>} (for the quizzes) which should be segmented to be \textit{\<لل>} (for), \textit{\<كويز>} (quiz), \textit{\<ات>} (plural) and tagged with AR-EN-AR. The SegRNN succeeded in this task. However, BiLSTM segmented it as \textit{\<لل>}, \textit{\<كو>}, \textit{\<يزات>} with the tags AR-EN-AR. Besides, the SegRNN tagged right the word \textit{messages}, which should be tagged as EN, the SegRNN tagged it right. Nevertheless, BiLSTM tagged it to be \textit{mess} as AR, \textit{ag} as EN, and \textit{es} as AR.

\subsection{Coarse-grained NE}

One way of annotating the named entities that we could have used is based on the script.
However, this would have led to more problems, like not considering the Arabic
words written in Lattin letters (Arabizi). For example, if the annotation were
based on the script, the word \textit{Masr} (Egypt) would be \textit{NE.EN}, but there is no such English word,
and it would not be distinguished from \textit{Egypt}, which is \textit{NE.EN}. Then any valuable language information is lost. Having language IDs on NEs also has a
practical aspect. When the NE of the language is known, it can be found in a
dictionary or an embedding. Otherwise, these will be unknown words. Thus, our
annotation decision of the entities was based on the context. Whenever most of
the words in a sentence are Arabic, common entities are considered Arabic, like
Arabic names written in English letters.

\begin{table}[h]
    
    \begin{center}
    \caption{Coarse-grained NE training models results}
    \small
    \begin{tabular}{l|r|r|r}
    \hline
   \textbf{Evaluation} & \textbf{Na\"{i}ve}  &  \textbf{Character} & \textbf{SegRNN} \\
   \textbf{Metric} & \textbf{Bayes}  &  \textbf{BiLSTM} & \\
        \hline
Tag F1          &   90.81   & 86.69  & 94.82   \\ 
Acc             &   90.40   & 89.15  & 95.07   \\
Seg F1          &   96.61   & -      & 99.30   \\
Char Acc        &   92.91   & -      & 94.89   \\
Mixed Tag F1    &   25.04   & 0.0    & 80.76   \\
Mixed Seg F1    &   26.15   & -      & 79.28   \\
Mixed Acc       &   28.14   & -      & 77.34   \\
Mixed Char Acc  &   71.48   & -      & 90.12   \\
        \hline
   \end{tabular}
   \label{table:Coarse}
    \end{center}

\end{table}

In order to compare the fine-grained NE, we tested our models with the coarse-grained named entity by combining the NE.AR and NE.EN tags to NE tag. As shown in Table \ref{table:Coarse}, no change exists for the Na\"{i}ve Bayes model as initially it failed to recognize the NE tags in the main model. However, for the Character BiLSTM, there is some improvement in the tagging F1-score and Char. Acc. parameters for all data and mixed data, but it does not perform well in the segmentation task. For the SegRNN model, training the model with coarse-grained NE achieves better results for the segmentation task for all data, but its performance was lower than for the main model in the tagging and accuracy tasks for all data. However, it achieved better results in the mixed data for all parameters except the tagging task. 
 \section{Summary}
 
Arabic speakers also use code-switching within the same word by adding an Arabic prefix or suffix to an English stem. We created the first annotated AR-EN corpus with the focus of intra-word CS. We collected and annotated 2,507 sentences 30,321 tokens from various social media platforms. We implemented LID models using SegRNN and investigated the usage of different types of embeddings. We compared the models with two baseline models. The results showed that the SegRNN model outperformed all baseline models, and it achieved an F1-score of 94.84\% for LID intra-word tagging and 99.17\% for segmentation.

%% file: conclusion.tex
\chapter{Conclusion \& Future Work}\label{chap:concl}

Code-switching has become a typical linguistic behavior being studied, and a massive amount of CS data is generated through different social media platforms. This data needs to be investigated and analyzed for several linguistic tasks. In this work, we focused on applying the NER task on CS data by proposing a pipeline approach composed of deep learning NER models and other complementing NLP tasks. This pipeline approach could be used to investigate other code-switched language pair after applying similar techniques and following guidelines to collect CS data for the different proposed tasks.

We proposed two techniques that could be used to complement the NER taggers on CS data and enhance their performance. The first technique was training existing contextual embedding models using our created corpora for this task. In addition, we proposed a new contextual model called KERMIT. As word embeddings are essential in any deep learning model, having contextual embeddings trained on the same language pair is an important step, as has been proved in this work.

The second technique was proposed to overcome the problem of limited CS data-set by applying data augmentation techniques. The proposed techniques that we presented were word embedding substitution, back-translation and modified approach of EDA. These techniques could be used as well on monolingual NER data. We proved that a convenient augmentation approach could positively impact the NER task, especially for CS data, as it will reduce the need to collect and annotate new data while enhancing performance at the same time.

Besides, we implemented an intra-word language identification task on CS data that could be used as a preprocessing step to identify the language of different tokens and apply the NER task on the intra-words. We created the first annotated AR-EN corpus with the focus of intra-word CS.

Recommendations for future work include the fact that, there are many new directions that could be investigated and applied on CS data and our currently available ones could be enhanced. We can apply all our NER and LID models on other available language pairs and evaluate their performance.
Starting with the available NER taggers, we can enhance them for both MSA and CS data by adding different layers, such as, for example, attention layer in the different RNN models and compare the performance. Also, we can add affix embeddings with the word and character embeddings in the RNN models. The affix embeddings consider all N-gram prefixes and suffixes of words in the training data, which could be useful for the Arabic language. In addition, the Transformers architectures could be explored more and used in the implementation of NER taggers for MSA and CS data.
We intend to publish our Arabic-English CS corpus that we created along with the different taggers. We may use some transfer learning techniques to import models trained on other CS languages. Besides, collecting more CS data can improve performance; especially if CS data is combined with the data pool, introduced in this work. New data could have the focus as well as Arabizi, Engary and intra-word CS. We can evaluate all our available taggers on the new data to recognize a wider range of Arabic variants. We can also explore the explainable AI \cite{arrieta2020explainable} field, to check the reasons behind the models performance and explore the problems. 

Regarding the first enhancement technique of CS contextual embeddings, more comparisons with available models could be made. We can pre-train variants of the models by keeping the monolingual data without CS behavior and comparing performance in different tasks. The models can be evaluated in different NLP tasks using various annotated data-sets. An evaluation technique should be applied to the generated CS data to check how it simulates the CS behavior. As training word embeddings require having a huge corpora, several approaches could be investigated to overcome this problem or collect more data such as, for example, by applying data augmentation techniques to available corpora or using generative adversarial networks \cite{gao2019code}. In addition, since translating data was the main approach to generate CS corpus for training the embedding models, another approach should be computed to have real data, rather than simulated ones. Another direction that can be explored is enhancing the efficiency of the Transformer models as they are huge models that require millions of parameters to compute attention. Some of the enhancement techniques that can reduce the parameters are pruning \cite{gordon2020compressing}, weight quantization \cite{dong2019hawq} and parameter sharing \cite{lan2019albert}. Another important direction could be exploring how the powerful character-level language models such as, for example, FLAIR could be integrated with Transformer architectures.


Concerning data augmentation, we can enhance the existing methods and use new ones. Different back-translation paths can be evaluated by trying different intermediate languages and using more robust machine translation systems between the language pairs. More combinations between the different data augmentation techniques should be applied, like combining the back-translation approach with the contextual embedding substitution. Also, we can apply the three augmentation techniques simultaneously on the same data-set and evaluate the performance for the model. 
About the selection of words to be substituted in the EDA and the Word Embedding Substitution method, we can compare the results of these two techniques substituting the same proportion of words. A validation technique should be applied to the assigned labels for the augmented data, like using multiple taggers for the same word to reduce the percentage of possibility of having a wrong label assigned. Besides, we can implement an automatic technique to check the quality of the augmented data and make the data available for other researchers. We can evaluate each language alone by applying the DA technique on it and check its impact on NER performances. Moreover, we can extend the use of augmentation techniques to other CS data-sets of different NLP tasks to support applying these techniques in a multilingual context.


About the intra-word LID task, we can collect more data containing a bigger portion of mixed words. Other new techniques could be investigated to apply intra-word LID task. Another version of the
data containing the labels more fine-grained by having labels reflecting the writing script whether it is MSA, Engary, Arabizi or English can be developed. In addition, we can apply
these labels to the NE tokens as well. 

As for the general directions and new tasks related to the NER that we can investigate in the future for the Arabic language and especially Arabic-English CS. These are Entity resolution/linking, Named entity disambiguation, Co-reference resolution, and Relation Extraction. Entity resolution/linking could be considered a second step task after NER. It is the process of matching the entities/mentions to the knowledge base, such as, for example, Wikipedia records based on their context. This task would be interesting and challenging when it is applied to CS text. In order to find the correct link, another task should be applied, which is Named entity disambiguation, to solve the ambiguity of the entities. Co-reference resolution is another task that we can apply along with the Entity resolution task. It refers to identifying which NE (mentions) refer to the same entity and creating mention clusters. Besides, we can apply the Relation extraction task, which is discovering relationships between entities by extracting pairs of entities related to each other and identifying the type of relation. In addition, we can apply the sentiment analysis task, analogous to what we proposed in \cite{sabtygamifiedFB} but for the language pair AR-EN. They are all challenging tasks for Arabic and new ones for CS Arabic-English data that we can explore.


%% file: appendix.tex
\appendix
\renewcommand{\appendixtocname}{Appendix}
\renewcommand{\appendixpagename}{\appendixtocname}
\addappheadtotoc
\setboolean{@twoside}{false}

\listoffigures
\addcontentsline{toc}{section}{List of Figures}